\documentclass[runningheads]{llncs}

\usepackage{eccv}

\usepackage{eccvabbrv}

\usepackage{graphicx}
\usepackage{booktabs}
\usepackage{subcaption}
\usepackage{multirow}
\usepackage{float}

\usepackage{wrapfig}   % preamble
\usepackage[accsupp]{axessibility}  % Improves PDF readability for those with disabilities.

\usepackage{hyperref}
\usepackage{multirow}
\usepackage{multicol}
\usepackage{soul}
\usepackage{booktabs}
\usepackage{bbm}

\usepackage{orcidlink}
\usepackage{bm}
\newcommand{\PaperTitle}{AdaptiveSplat}

\begin{document}

% ---------------------------------------------------------------
\title{AdaptiveSplat: Texture Aware Controllable 3D Gaussian Allocation for Feed-Forward Reconstruction}

\titlerunning{\PaperTitle}

\author{Badrinath Singhal\inst{1} \and
Srihari K G\inst{1} \and
Sreehari Iyer\inst{1} \and
Ankit Dhiman\inst{1,2} \and
Venkatesh Babu Radhakrishnan\inst{1}
}

\authorrunning{Singhal et al.}

\institute{Vision and AI Lab, Indian Institute of Science, Bangalore \and Samsung R \& D Institute India - Bangalore \\
\email{\{badrinaths, sriharig, sreeharid, ankitd, venky\}@iisc.ac.in}\\
\textcolor{blue}{\url{https://badrinaths.github.io/projects/adaptive-splat/} 
}}

\maketitle

\begin{figure}
    \centering
    \includegraphics[width=\textwidth]{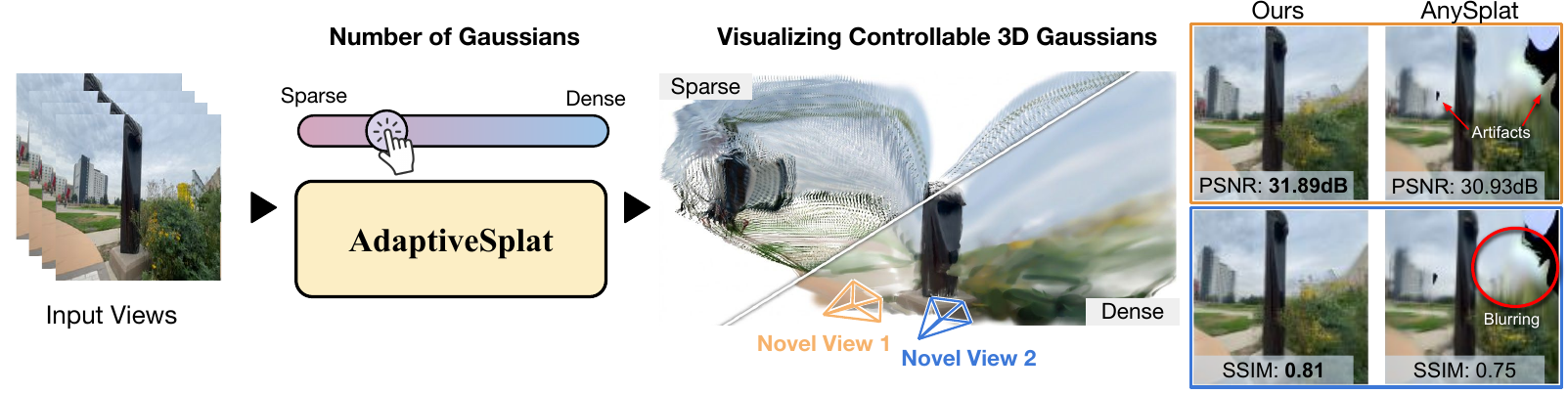}
    \captionof{figure}{\textbf{Overview:} We demonstrate \textbf{\PaperTitle}'s ability to maintain scene fidelity while controlling the allocation of primitives. Given input images and a target Gaussian budget, our feed-forward approach produces sparse yet high-quality reconstructions. In contrast, prior methods exhibit artifacts and over-smoothing at high sparsity.}
    \label{fig:teaser}
\end{figure}

\vspace{-10mm}
\begin{abstract}
Current feed-forward 3D reconstruction methods predict pixel aligned Gaussian primitives, resulting in highly redundant representations.
A natural solution is to prune the redundant Gaussians, but naive pruning introduces severe artifacts and often requires inference time fine-tuning, breaking the feed-forward paradigm.
Based on previous works, high frequency regions require more Gaussian primitives, while low frequency regions can be represented with significantly fewer primitives.
Motivated by this, we propose a novel approach to explicitly control the number of Gaussians by leveraging local texture information. 
Our approach achieves this through three key components: (1) texture estimation to capture spatial variation in scene detail, (2) texture-aware pruning that removes redundant Gaussians from low frequency regions, and (3) an adaptive Gaussian head that predicts the modified attributes of the retained primitives without breaking the feed-forward paradigm.
Experiments on RE10K, ACID, DL3DV, Tanks and Temples, and DTU demonstrate the effectiveness of our approach, while ablation studies validate the contributions of its key components.
\end{abstract}

\section{Introduction}
\label{sec:intro}
Feed-forward models for 3D reconstructions \cite{splatt3r, noposplat, jiang2025anysplat, pixelsplat, mvsplat, latentsplat, gaussiangraphnetwork, splatterimage, longlrm, hisplat, li2025flashworld} generate 3D Gaussian representations \cite{3dgs_paper} directly from input views.
These approaches predict pixel-aligned Gaussians, where each input pixel produces the parameters of a corresponding 3D Gaussian in the scene.
While this design enables efficient feed-forward inference, it allocates Gaussians uniformly across the image without considering the underlying scene complexity.
A straightforward way to reduce this redundancy is to prune the Gaussians.
However, naive pruning strategies introduce severe artifacts in rendered novel views, often producing empty regions that disrupt scene continuity and degrade rendering quality.

Several works have explored compression techniques for 3D Gaussian representations \cite{eagles_compression, SOG3DGS, minisplatting, trimminggaussian, navaneet2023compact3d, compressed_3dgaussian}.
However, these approaches primarily focus on post-hoc quantization or encoding and typically require dense multi-view inputs along with scene-specific optimization.
Such reliance on iterative optimization limits their applicability to the challenging feed-forward setting.
Other works attempt to prune pixel-aligned Gaussians directly within feed-forward pipelines using geometry-driven heuristics such as overlap removal \cite{gaussiangraphnetwork}, opacity thresholding \cite{longlrm}, or enforcing a single Gaussian per voxel \cite{jiang2025anysplat}.
While computationally simple, these strategies are largely agnostic to scene complexity and offer little control over the pruned Gaussians.
Consequently, they may remove primitives that are necessary to represent high-frequency structures \cite{GenerativeDensification, hisplat}.
These limitations raise an important question: \textbf{Can we enable controllable 3D Gaussian density in feed-forward models without sacrificing their fidelity?}

\begin{figure}[!t]
    \centering
    \includegraphics[width=\linewidth]{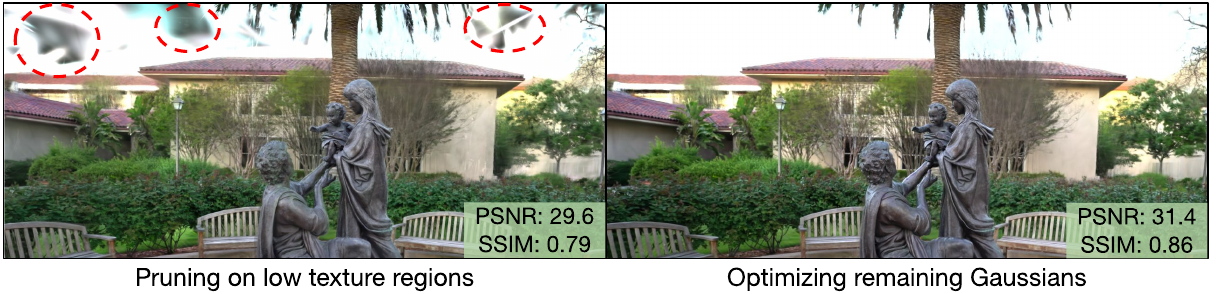}
    \caption{Toy experiment demonstrating core insight for \textbf{\PaperTitle}. (Left) A rendered view shows visible artifacts, such as "black patches" that are indicative of naive pruning in low-texture regions. (Right) The same scene is rendered after refining the remaining Gaussians. The rendering quality is significantly restored as the surviving Gaussians in the low-texture areas have adaptively expanded to fill the gaps, thereby eliminating the observed artifacts.}

    \label{fig:toy-experiment}
    \vspace{-5mm}

\end{figure}

We draw inspiration from well-established statistical properties of natural images, whose power spectral density (PSD) follows an approximate power-law decay ($PSD \propto \frac{1}{f^\alpha}$, where $f$ denotes spatial frequency and $\alpha \approx 2$)~\cite{naturalscenestatistics}. 
This observation implies that most visual energy is concentrated in low-frequency components, while high-frequency structures are sparse and localized. 
Consequently, different regions of a scene require different levels of representation density: smooth areas with low-frequency content can be reconstructed with relatively few Gaussians, whereas highly textured regions containing fine structural details require denser Gaussian allocation. 
Prior work~\cite{GenerativeDensification, hisplat} empirically supports this hypothesis, demonstrating that high-frequency regions benefit from increased Gaussian density for accurate reconstruction. 
However, existing feed-forward architectures largely ignore this requirement due to their fixed pixel-aligned prediction strategy.
To further illustrate this observation, we conduct a simple experiment shown in Fig.~\ref{fig:toy-experiment}. 

Motivated by this key insight, we propose \texttt{\PaperTitle}, a feed-forward framework that enables controllable 3D Gaussian allocation while preserving reconstruction fidelity. 
Our method focuses on capturing the variance in visual complexity. This is achieved by estimating the texture of the scene.
Based on this signal, we perform texture-aware pruning that removes redundant Gaussians more aggressively in smooth regions while preserving those necessary for detailed structures. 
To compensate for the removed primitives, we introduce a novel adaptive Gaussian head that re-predicts the attributes of the remaining Gaussians within the same feed-forward pipeline. 
This design enables flexible control over the Gaussian budget without requiring test-time optimization or post-pruning fine-tuning.
Importantly, our approach can be seamlessly integrated into existing feed-forward 3D Gaussian pipelines without requiring architectural modifications. 
In summary, our contributions are as follows:
\begin{itemize}
\item \textit{Texture-Based Scene Energy Estimation}: We introduce a texture-driven scene energy estimation mechanism that quantifies the amount of high-frequency content present in different image regions, enabling identification of areas that require higher representational capacity.

\item \textit{Texture-Aware Selective Pruning:}
Building on the estimated scene energy, we propose a principled pruning strategy that selectively removes Gaussians from low-texture regions while preserving those in high-detail areas.

\item \textit{Adaptive Gaussian Head:}
To mitigate artifacts introduced by pruning, we design an adaptive Gaussian head that re-predicts the attributes of the retained Gaussians, allowing the representation to adjust to the modified Gaussians.

\item We perform extensive experiments for various feed-forward models on datasets like RE10K, ACID, DL3DV, Tanks and Temples, and DTU. 
\end{itemize}

\section{Related Work}
\label{sec:relatedwork}

\subsection{Radiance Fields and Multi View Reconstruction}
Reconstructing 3D scenes from multi-view images is a core computer vision problem. Modern radiance field methods like NeRF \cite{nerf_paper,Dhiman_2023_ICCV} and 3D Gaussian Splatting (3DGS) \cite{3dgs_paper,dhiman2026turbo} achieve high-quality novel view synthesis, with 3DGS offering notable advantages in rendering speed through rasterization.\cite{scaffold} and \cite{hac} use structural anchors and contextual compression to reduce redundancy. These approaches have found success in applications such as 3D asset generation \cite{zero1to3, diffsplat, zero1tog, cat3d, DiffGS, dreamscene360, SyncTweedies, gaussianobject}, scene editing \cite{instructnerf2nerf, gaussctrl2024, GaussianEditor, nerfediting2021, editfreenerf, localeditnerf, dgsdrag, editing_dge, refGS}, and dynamic scene reconstruction \cite{3dgs_paper, dynamicgauss, cat4d, dynamicgsmonocularvideo, Dynnerf}. However, they often require dense multi-view images which limit adaptability and efficiency in dynamic, real-time environments. Sparse-view reconstruction methods \cite{pixelnerf, pixelsplat, latentsplat, zero1to3, reconfusion, poole2022dreamfusion, Koo:2024PDS} leverage learned priors from datasets or foundational models \cite{latentdiffusion}. Recently, Feed-forward methods \cite{DUST3R, MAST3R, wang2025vggt, keetha2026mapanything, wang2025pi3} offer a compelling alternative, directly inferring 3D representations with superior generalization across diverse scenes, bypassing the computationally expensive steps of Structure-from-Motion (SfM) \cite{colmap} and radiance field optimization by directly predicting dense 3D scene representations from limited views \cite{xu2024grm, wang2025f3dgaus, GenerativeDensification, keetha2024splatam} but are rigid in number of 3D Gaussian predictions. These methods have been extended to dynamic scenes \cite{monst3r}, multi-view consistency \cite{MET3R}, and estimating 3DGS representations \cite{zhou2024feature, hisplat, Yu2024PolyOculusNVS, noposplat, splatt3r, spars3r}, even from just a few views \cite{jiang2025anysplat, longlrm}.

\subsection{Compression in 3D Gaussian Splatting}
Due to its explicit nature, 3DGS often requires a large number of Gaussians to represent a scene, leading to higher storage, processing time, and slow rendering. To address these challenges, several works focus on compressing 3DGS representations. EAGLES \cite{eagles_compression} prunes based on scene-wide influence, targeting low-contribution Gaussians (e.g., small scale, low opacity, occlusion) and LightGaussian \cite{lightgaussian} prunes based on estimated importance (ray hits, volume), PUP-3DGS \cite{HansonTuPUP3DGS} uses uncertainty estimation to estimate Gaussians' importance by approximating second order gradient of loss. EfficientGS\cite{efficientgs} estimates importance of Gaussians per pixel by selecting Top-K Gaussians to be retained, and SOG-3DGS \cite{SOG3DGS} applies image compression techniques. Other methods like \cite{trimminggaussian} remove Gaussians with low gradients followed by fine-tuning, while \cite{minisplatting} proposes pruning and densification based on alpha blending importance. However, a common limitation across these compression techniques is their reliance on quantization or their inability to dynamically modify the attributes of remaining Gaussians post-pruning, which can lead to visible artifacts. Moreover, all existing methods require dense multi-view images of the scene to compress which fail in the sparse-view feed-forward setting. Interestingly, previous work like \cite{xu2023wavenerf} has explored using wavelet based frequency decomposition to achieve high quality synthesis for multiview images but no such method exists for 3D Gaussian Splatting representation.

\begin{figure*}[t]
    \centering
    \includegraphics[width=\linewidth]{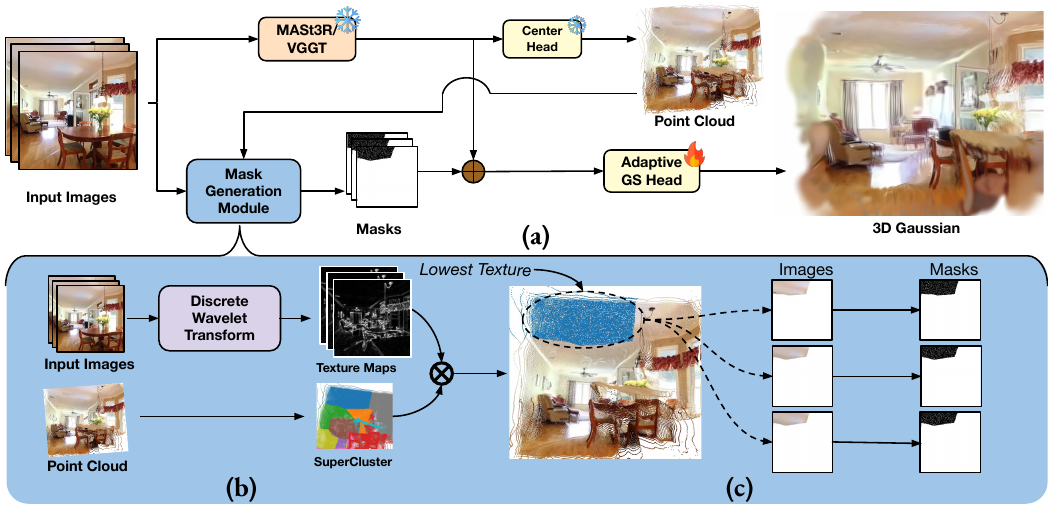}
    \caption{ Our pipeline allocates the desired number of 3D Gaussians for real-time, novel-view synthesis. (a) \textbf{Initial Representation:} We infer a 3D point cloud from context images using the MASt3R/VGGT backbone. (b) \textbf{SuperCluster Formation and Texture Analysis:} The point cloud is partitioned into SuperClusters, and texture information is simultaneously analyzed using DWT to identify low-texture regions. (c) \textbf{Gaussian prediction:} Based on the texture, masks are generated that are used in Adaptive Gaussian Head to predict optimized attributes of remaining 3D Gaussian primitives.}
    \label{fig:pipeline}
    \vspace{-4mm}
\end{figure*}

\section{Method}
We introduce a controllable feed forward 3D Gaussian generation pipeline.  \cref{fig:pipeline} shows the overall pipeline of our method.

\paragraph{\textbf{Problem Formulation}}

Existing feed-forward 3D Gaussian prediction models generate a pixel-aligned set of 3D Gaussians $\mathcal{G}_k$ for each input view. Given $m$ views $\{(I_k, v_k)\}_{k=1}^{m}$, where $I_k \in \mathbb{R}^{H \times W \times 3}$ denotes the RGB image and $v_k$ its associated camera pose, such models produce a set of $N = mHW$ primitives without explicit control over representation size. In contrast, we introduce explicit control over the number of retained Gaussians. Under a user-defined pruning budget $\beta \in [0,1)$, our objective is to predict a compact subset $\mathcal{G} = \{g_i\}_{i=1}^{B}$ where $B=\lfloor(1-\beta)N\rfloor$. Each Gaussian is parameterized as $g_i = \{\bm{\mu}_i, \bm{s}_i, \bm{q}_i, \bm{\alpha_i}, \bm{\rho}_i\}$, where $\bm{\mu}_i \in \mathbb{R}^3$ denotes the 3D mean, $\bm{s}_i \in \mathbb{R}^3$ anisotropic scale, $\bm{q}_i \in \mathbb{R}^4$ a unit quaternion encoding rotation, $\alpha_i \in [0,1]$ opacity, and $\bm{\rho}_i \in \mathbb{R}^{h}$ spherical harmonics coefficients modeling view-dependent appearance. 

\subsection{Texture-Based Scene Energy Estimation}

We estimate spatially localized scene complexity using the Discrete Wavelet Transform (DWT) applied to the input images. Given images $\{I_k\}_{k=1}^m$, its single-level 2D DWT is computed as $W_{ij}^{d} = \sum_{u}\sum_{v} I_k(u,v) \cdot \psi^{d}(i-u, j-v)$, where $\psi^{d}$ denotes the directional wavelet basis. This decomposes the image into three high-frequency detail components corresponding to horizontal, vertical, and diagonal orientations. 
We define the local texture energy at pixel $(i,j)$ in view $k$ as $E_{i,j}^{(k)} = \sum_{d \in {h,v,diag}} |W_{ij}^{d}|$, which aggregates directional high-frequency responses.

\subsection{Texture-Aware Selective Pruning}
\label{sec:method_region_selection}

Using a pretrained multi-view backbone, we extract a dense pointmap $\mathcal{P}_k = \{ p_i\}_{i=1}^{HW}$ from each input view $I_k$ where each point $p \in \mathbb{R}^6$ encodes its 3D location and RGB color. We aggregate all views to form a global point cloud $\mathcal{P} = \bigcup_{k=1}^{m} \mathcal{P}_k$, where $\mathcal{P} = \{ p_i\}_{i=1}^{N}$.
To obtain homogeneous regions, we partition $\mathcal{P}$ into $K$ SuperClusters using $K$-means clustering. Clustering in $(x,y,z,r,g,b)$ space encourages points that are spatially proximate and photometrically consistent to be grouped together. Such regions correspond to locally smooth surfaces that can be represented with fewer but spatially broader Gaussians. Formally, each SuperCluster $SC_i$ is defined as $SC_i = \{ p \in \mathcal{P} \mid |p - c_i|^2 \leq |p - c_j|^2, \forall j \neq i \}$ where the centroid is $c_i = \frac{1}{|SC_i|} \sum_{p \in SC_i} p$. The resulting clusters $\{SC_i\}_{i=1}^{K}$ form a disjoint partition of the point cloud such that $\bigcup_{i=1}^{K} SC_i = \mathcal{P}$ and $SC_i \cap SC_j = \emptyset$ for $i \neq j$.  

We assign texture energy to each SuperCluster by averaging the wavelet coefficients associated with its constituent points $p$. Let $\pi_k(p)$ denote the projection of point $p$ onto image $I_k$ via pointmap $\mathcal{P}_k$. The SuperCluster energy is computed as $E_{SC_i} = \frac{1}{|SC_i|} \sum_{p \in SC_i} \bar{E}(\pi(p))$ where $\bar{E}(p)$ denotes the wavelet coefficient of the pixel corresponding to the point $p$.
These SuperCluster-level energies provide a structured measure of regional complexity and guide subsequent Gaussian allocation under the pruning budget.

\paragraph{\textbf{Binary Mask Construction:}}\mbox{}\\
\begin{wrapfigure}{r}{0.45\linewidth}
\centering
\vspace{-\baselineskip}
\includegraphics[width=\linewidth]{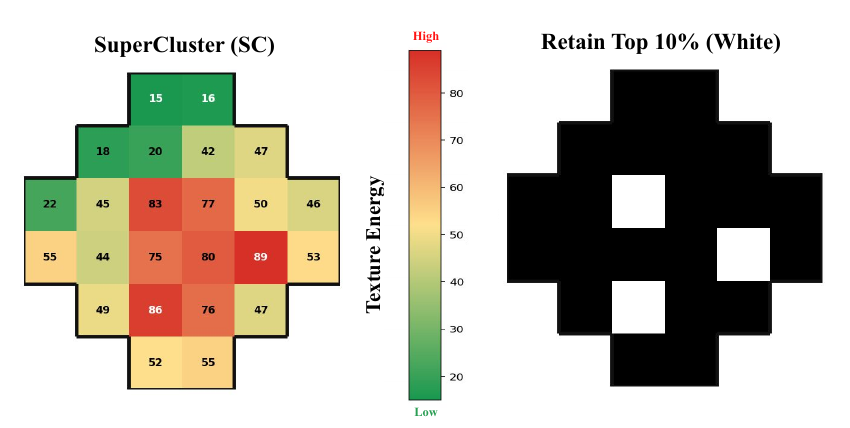}
\caption{Within each SuperCluster, Gaussians are ranked by texture energy, and the top 10\% ($\gamma = 0.1$) are kept while the rest are masked out. This rule is applied only to low-energy clusters; high-energy clusters are fully retained.}
\label{Rep_Gauss}
\end{wrapfigure}
$\{SC_1,\dots,SC_K\}$ denote the regions ordered by increasing energy $E_{SC_j}$.
Each region $SC_j$ has a nominal retention of $\gamma |SC_j|$ Gaussians, where $\gamma=0.1$ (empirically validated, ablation provided). We choose the largest index $\ell$ such that $\sum_{j=1}^{\ell} \gamma |SC_j| \;\le\; B$. This ensures that we can effectively cover the entire low-texture region with 3D Gaussians with fewer retained Gaussians while pruning the rest.

For all regions $j<\ell$, we retain the $\gamma |SC_j|$ Gaussians. For the final region $SC_\ell$, we retain only
$
|R_\ell| = B - \sum_{j=1}^{\ell-1} \gamma |SC_j|.
$
Let $\mathcal{G}_{\mathrm{ret}}$ denote the retained set of Gaussians.
The corresponding binary mask entries for each context view $i \in \{1,\dots, m\}$ are given by
$M_{i,j} = \mathbbm{1}\!\left[\, G_j \in \mathcal{G}_{\mathrm{ret}} \cap \mathcal{G}_i \,\right], \quad j \in \{1,\dots,HW\}$ (see Fig.~\ref{Rep_Gauss}).

\subsection{Adaptive Gaussian Head}

Given the features $F$ and binary masks $M_1, \dots M_m$, the Adaptive Gaussian Head $f_\theta$ predicts refined Gaussian attributes for the retained indices: 

$f_\theta(F,M) \rightarrow \{\bm{s}_i, \bm{q}_i, \bm{\alpha_i}, \bm{\rho}_i\}_{i=1}^{B}$. 
The Adaptive Gaussian Head is built on the DPT architecture~\cite{dpt_paper} with additional convolutional layers that fuse ViT~\cite{visiontransformervit} features extracted from the multiview feature map $F$ with binary masks $M_1, \dots, M_m$.

\subsection{Training and Inference}
\label{sec:training_pipeline}

\paragraph{\textbf{Training:}} To ensure that the model's performance is stable across diverse budget rates $\beta$ at test time, it is important that it generalizes well to a broad spectrum of pruning percentages during training. 
To achieve this, we train the Gaussian Head with $\beta \sim \mathcal{U}[0, 1)$. This improves robustness of the method to different budgets $\beta$, enabling adaptation of the retained 3D Gaussians.

The network is trained using photometric loss on the rendered image from the retained 3D Gaussians at the target camera pose and the corresponding ground truth image using the following loss formulation $\mathcal{L}$ as follows:
\begin{equation}
    \mathcal{L} = \frac{1}{|v_t|}
\sum_{v \in {v_t}} \left( \|I_{v} - \hat{I}_{v}\|_2^2
+ \lambda \cdot 
\mathrm{LPIPS} \left(I_{v}, \hat{I}_{v}\right) \right),
\end{equation}
Here $v_t$ represents the set of target camera views, $I_v$ denotes the target ground truth image and $\hat{I}_v$ denotes the rendered image from the target view $v$. 
3D Gaussians are obtained from Gaussian Head $f_\theta$ using context view decoder features along with its corresponding masks $M_1, M_2, \dots M_m$. The weighting coefficient $\lambda$ controls the perceptual regularization strength. 

\paragraph{\textbf{Inference:}} At inference, the user selects a desired sparsity level $\beta \in [0 ,0.8]$. The system computes DWT from images, forms masks, and directly predicts adapted Gaussian attributes via the Gaussian Head.

\section{Experiments}
\label{sec:exp}

\subsection{Experimental Setup}

\paragraph{\textbf{Implementation Details}} Our approach builds upon the architecture of MASt3R \cite{MAST3R} for sparse-view inputs and VGGT \cite{wang2025vggt} for dense-views with a DPT \cite{dpt_paper} Gaussian Head to predict 3D Gaussian primitives similar to \cite{noposplat, jiang2025anysplat}, which integrates a Gaussian prediction head into the MASt3R/VGGT pipeline. This allows for extraction of 3D Gaussian attributes for each point in the point cloud, providing an initial 3D representation. These methods don't require camera poses as input; thus, we also implement our approach on MVSplat\cite{mvsplat} backbone which also takes camera pose as input. We choose $K=300$ and $\gamma=10\%$ for our pipeline (ablations provided on these hyperparameters) and $\lambda$ is set to $0.001$. We trained our model on 1 NVIDIA A100 40GB GPU for 48 GPU hours on each dataset. We use available pretrained checkpoints of the baseline feed-forward models for the evaluation.

\paragraph{\textbf{Datasets}} We train and evaluate our method on diverse real-world datasets following the experimental setup of pixelSplat \cite{pixelsplat} and AnySplat \cite{jiang2025anysplat}. Our primary training and evaluation data is derived from RealEstate10K \cite{re10kdataset} (indoor, multi-view videos), ACID \cite{acid_dataset} (large-scale, dynamic outdoor drone scenes) and DL3DV \cite{dl3dv_dataset} (large scale scene) datasets. These datasets provide multi-view images with COLMAP-computed \cite{colmap} camera poses, and we adhere to their official train-test splits. To assess zero-shot generalization, we also evaluate on the DTU dataset \cite{dtu_dataset, DTU_dataset_cvpr}, comprising high-quality, object-centric scans. We further report results on the Tanks and Temples dataset~\cite{tanksandtemples} in the supplementary material.

\paragraph{\textbf{Evaluation Criteria}} We benchmark our method on novel view synthesis using widely adopted image quality metrics: Peak Signal-to-Noise Ratio (PSNR), Structural Similarity Index Measure (SSIM) \cite{ssim}, and Learned Perceptual Image Patch Similarity (LPIPS) \cite{lpips}. Beyond these measures, we explicitly compare against $\beta$ used for 3D scene representation. This metric directly demonstrates our approach's efficiency in reducing redundancy while maintaining superior visual fidelity.

\subsection{Comparisons with Baselines}
Comparisons with existing feed-forward methods are challenging because they lack controllable Gaussian allocation and are designed for dense, per-pixel prediction, unlike our approach. To enable fair comparison, we construct baselines by applying established pruning techniques on top of state-of-the-art feed-forward methods. We consider four representative sparse-view feed-forward methods: pixelSplat \cite{pixelsplat}, MVSplat \cite{mvsplat}, NoPoSplat \cite{noposplat}, and HiSplat \cite{hisplat}, as well as Gaussian Graph Network (GGN) \cite{gaussiangraphnetwork}, AnySplat \cite{jiang2025anysplat}, a recent method designed for dense-view settings. On top of the Gaussian primitives predicted by these models, we apply four representative pruning strategies: EAGLES \cite{eagles_compression}, LightGaussian \cite{lightgaussian}, EfficientGS \cite{efficientgs}, and PUP3DGS \cite{HansonTuPUP3DGS}.

To compare with existing methods fairly, baselines must also operate under the feed-forward setting. However, directly applying pruning to the predicted Gaussians is suboptimal, as most pruning strategies assume per-scene optimization. While this optimization can partially recover reconstruction quality, it breaks the feed-forward assumption. To account for this trade-off, we evaluate baselines under two settings: (1) pruning followed by per-scene fine-tuning, which we refer to as with fine-tuning, and (2) direct pruning, which preserves the feed-forward property and is referred to as without fine-tuning. We evaluate our approach in both sparse-view and dense-view feed-forward settings. For sparse-view evaluation, we use two input views, following the protocol of pixelSplat~\cite{pixelsplat}. For dense-view settings we evaluate using 6, 9, 16, and 32 view inputs following the protocol of AnySplat \cite{jiang2025anysplat}. We report results for the best-performing backbone, while detailed comparisons across all backbones pixelSplat, MVSplat, HiSplat, NoPoSplat, GGN, InstantSplat \cite{fan2024instantsplat} and AnySplat are provided in the supplementary.

\subsection{Inference Time Analysis}
To highlight the practical benefits of controllable pruning, we report rendering throughput under different pruning strengths. We measure rendering speed by synthesizing $1000$ novel views for each pruning strength and reporting the Frames Per Second (FPS).

\subsection{Ablation Study}
% \label{sec:ablation-main}
We analyze the impact of the key design choices in our method.
Each ablation isolates a specific component to evaluate its contribution to the overall performance.

$\bullet$ \textit{Texture Ablation:} We analyse the importance of texture for ranking SuperClusters. In our method, texture energy determines the order in which SuperClusters are selected for pruning. In this ablation, we remove the texture-based ranking and train the model end-to-end while selecting SuperClusters randomly for pruning to evaluate the importance of texture guidance.

$\bullet$ \textit{Importance of Hyperparameter $K$ and $\gamma$: } We provide ablations on various choices of $K$ and $\gamma$, which control the number of SuperClusters and the retention ratio within each SuperCluster.

$\bullet$ \textit{Adaptive vs Fixed Gaussian Head:} Here, we replace the Adaptive Gaussian Head with a fixed, pretrained Gaussian head and directly select the retained Gaussians from its predictions without further adaptation. This allows us to evaluate the importance of learning adaptive Gaussian attributes after pruning.

$\bullet$ \textit{Binary Mask:} To validate the importance of the binary mask, we train the Adaptive Gaussian Head with the mask as input and compare it with another Adaptive Gaussian head trained end-to-end without the mask.

Collectively, these ablations validate the importance of the key components of our method in achieving superior performance.

\subsection{Qualitative Texture Analysis}

The experiments so far focus on novel view synthesis to evaluate reconstruction quality. However, our method explicitly leverages texture signals to guide Gaussian allocation. To analyze this component more directly, we perform qualitative analysis with baselines focusing on the trends in the texture structure of the reconstructed scenes.

\begin{figure}[t]
    \centering
    \includegraphics[width=\linewidth]{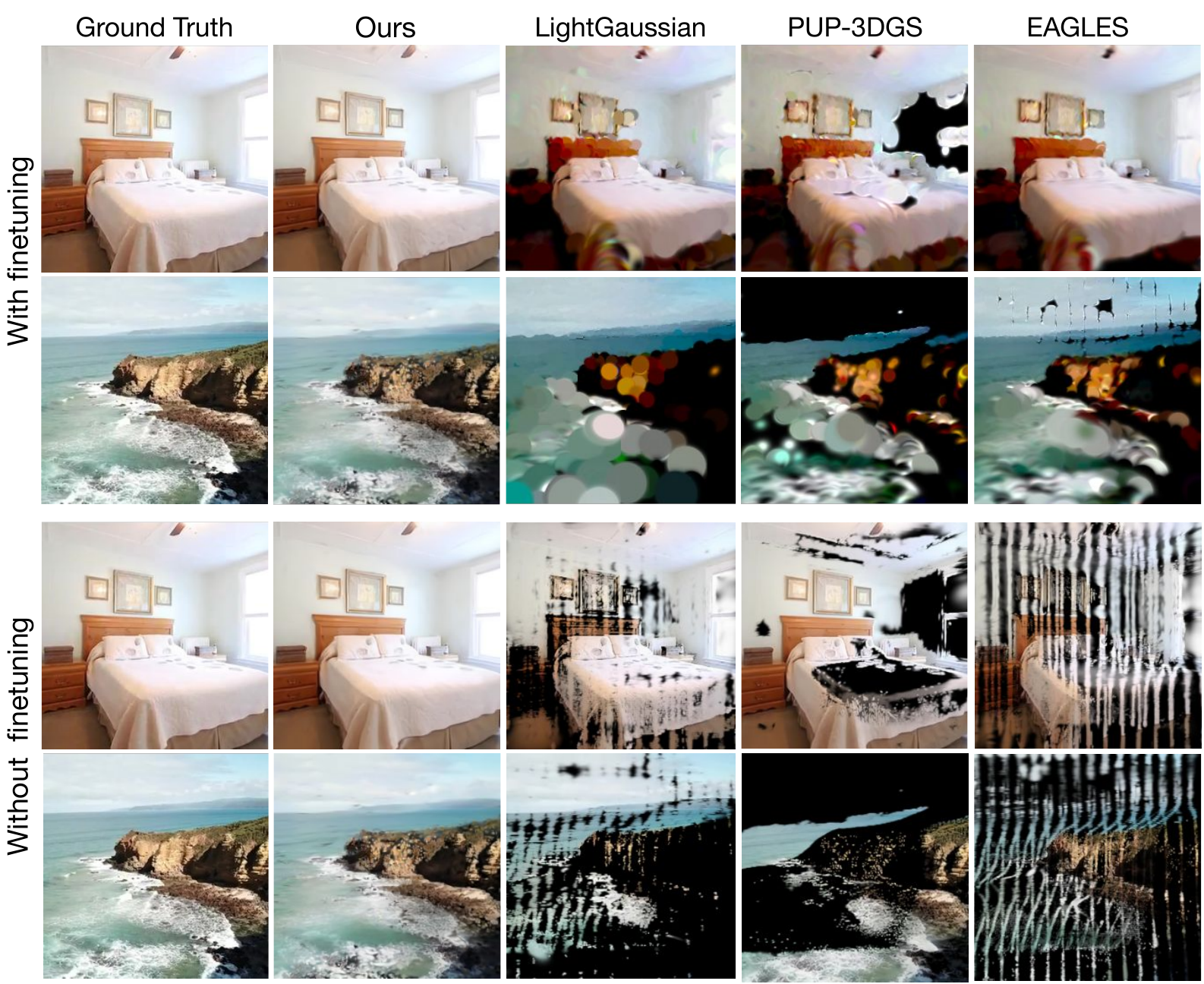}
    \caption{Comparison of results on RE10K and ACID scenes using $\beta=0.4$ and $\beta=0.8$ of target budget. We allocate fewer but larger 3D Gaussians to low texture SuperClusters by the feed-forward backbone.}
    \label{fig:pruning_comparison_combined}
\end{figure}

\begin{figure}[t]
    \centering
    \includegraphics[width=\linewidth]{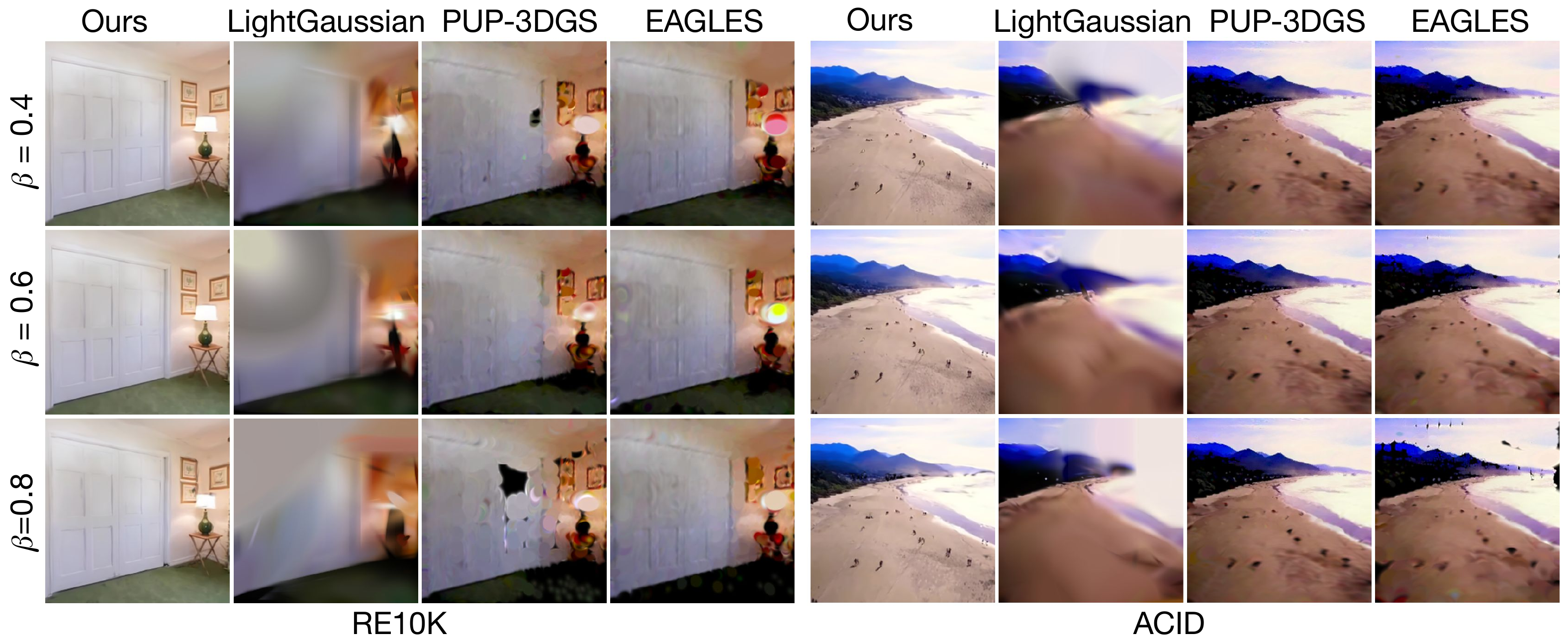}
    \caption{\textbf{Qualitative comparison} of our method with baseline methods (with finetuning) on a single scene from RE10K and ACID datasets. Our method faithfully represents a scene even with $\beta=0.8$ ($80\%$ pruning)}
    \label{fig:comparison_all_baseline_all_pruning_acid}
\end{figure}

\section{Results}

\begin{table*}[!t]
\centering
\caption{\textbf{Quantitative Results on ACID and RE10K Datasets.} Comparison of pruning methods with NoPoSplat and HiSplat as backbones across different pruning ratios $\beta$.}
\label{tab:results_acid_re10k}
\footnotesize
\setlength{\tabcolsep}{3pt}
\resizebox{\linewidth}{!}{
\begin{tabular}{@{}lllccccccccc@{}}
\toprule
 & & & \multicolumn{3}{c}{$\beta=0.4$} & \multicolumn{3}{c}{$\beta=0.6$} & \multicolumn{3}{c}{$\beta=0.8$} \\
\cmidrule(lr){4-6} \cmidrule(lr){7-9} \cmidrule(lr){10-12}
 & & Method & \textbf{PSNR$\uparrow$} & \textbf{LPIPS$\downarrow$} & \textbf{SSIM$\uparrow$} & \textbf{PSNR$\uparrow$} & \textbf{LPIPS$\downarrow$} & \textbf{SSIM$\uparrow$} & \textbf{PSNR$\uparrow$} & \textbf{LPIPS$\downarrow$} & \textbf{SSIM$\uparrow$} \\
\midrule
\multirow{9}{*}{\rotatebox[origin=c]{90}{ACID}}
 & \multirow{4}{*}{\rotatebox[origin=c]{90}{NoPoSplat}} & EAGLES      & 11.115 & 0.592 & 0.344 & 10.834 & 0.617 & 0.316 & 10.316 & 0.656 & 0.265 \\
 & & LightGauss. & 10.835 & 0.632 & 0.350 & 10.706 & 0.636 & 0.346 & 10.611 & 0.638 & 0.343 \\
 & & PUP-3DGS    &  9.672 & 0.640 & 0.266 &  8.616 & 0.672 & 0.203 &  7.722 & 0.705 & 0.137 \\
 & & EfficientGS &  9.171 & 0.665 & 0.193 &  9.290 & 0.661 & 0.196 &  9.303 & 0.663 & 0.195 \\
\cmidrule(l){2-12}
 & \multirow{4}{*}{\rotatebox[origin=c]{90}{HiSplat}} & EAGLES & 19.433 & 0.620 & 0.555 & 19.769 & 0.605 & 0.565 & 19.464 & 0.618 & 0.556 \\
 & & LightGauss. & 19.785 & 0.604 & 0.566 & 19.594 & 0.612 & 0.560 & 18.660 & 0.628 & 0.541 \\
 & & PUP-3DGS    & 19.382 & 0.617 & 0.558 & 18.529 & 0.643 & 0.538 & 18.003 & 0.659 & 0.527 \\
 & & EfficientGS & 19.762 & 0.607 & 0.565 & 19.677 & 0.608 & 0.564 & 19.310 & 0.619 & 0.556 \\
\cmidrule(l){2-12}
 & & \textbf{Ours} & \textbf{22.549} & \textbf{0.299} & \textbf{0.640} & \textbf{22.310} & \textbf{0.310} & \textbf{0.627} & \textbf{21.055} & \textbf{0.342} & \textbf{0.593} \\
\cmidrule(lr){1-12}
\multirow{9}{*}{\rotatebox[origin=c]{90}{RE10K}}
 & \multirow{4}{*}{\rotatebox[origin=c]{90}{NoPoSplat}} & EAGLES & 11.017 & 0.573 & 0.415 & 10.846 & 0.588 & 0.399 & 10.530 & 0.614 & 0.367 \\
 & & LightGauss. & 10.185 & 0.625 & 0.393 & 10.282 & 0.625 & 0.397 & 10.373 & 0.623 & 0.397 \\
 & & PUP-3DGS    & 10.298 & 0.599 & 0.371 &  9.737 & 0.623 & 0.332 &  8.772 & 0.656 & 0.263 \\
 & & EfficientGS &  8.742 & 0.658 & 0.262 &  8.841 & 0.655 & 0.266 &  8.801 & 0.658 & 0.263 \\
\cmidrule(l){2-12}
 & \multirow{4}{*}{\rotatebox[origin=c]{90}{HiSplat}} & EAGLES & 16.526 & 0.598 & 0.556 & 16.404 & 0.606 & 0.552 & 16.243 & 0.614 & 0.547 \\
 & & LightGauss. & 16.598 & 0.596 & 0.558 & 16.421 & 0.604 & 0.552 & 16.056 & 0.618 & 0.542 \\
 & & PUP-3DGS    & 16.472 & 0.601 & 0.558 & 16.083 & 0.618 & 0.543 & 15.588 & 0.638 & 0.529 \\
 & & EfficientGS & 16.525 & 0.598 & 0.556 & 16.404 & 0.606 & 0.551 & 16.242 & 0.613 & 0.546 \\
\cmidrule(l){2-12}
 & & \textbf{Ours} & \textbf{22.294} & \textbf{0.235} & \textbf{0.735} & \textbf{22.110} & \textbf{0.243} & \textbf{0.726} & \textbf{20.740} & \textbf{0.272} & \textbf{0.692} \\
\bottomrule
\end{tabular}
}
\vspace{-5pt}
\end{table*}

\begin{table*}[!t]
\centering
\caption{\textbf{Multiview Results on DL3DV Dataset.} Comparison of pruning methods using the AnySplat base model across varying view counts and pruning ratios $\beta$.}
\label{tab:results_dl3dv}
\footnotesize
\setlength{\tabcolsep}{3pt}
\resizebox{\linewidth}{!}{
\begin{tabular}{@{}llccccccccc@{}}
\toprule
 & & \multicolumn{3}{c}{$\beta=0.4$} & \multicolumn{3}{c}{$\beta=0.6$} & \multicolumn{3}{c}{$\beta=0.8$} \\ \cmidrule(l){3-5} \cmidrule(l){6-8} \cmidrule(l){9-11} 
$\quad$ & Method & \textbf{PSNR$\uparrow$} & \textbf{LPIPS$\downarrow$} & \textbf{SSIM$\uparrow$} & \textbf{PSNR$\uparrow$} & \textbf{LPIPS$\downarrow$} & \textbf{SSIM$\uparrow$} & \textbf{PSNR$\uparrow$} & \textbf{LPIPS$\downarrow$} & \textbf{SSIM$\uparrow$} \\ \midrule
\multirow{4}{*}{\rotatebox[origin=c]{90}{6 Views}} 
 & EAGLES & 10.464 & 0.675 & 0.241 & 9.206 & 0.688 & 0.193 & 7.693 & 0.703 & 0.129 \\
 & LightGauss. & 11.361 & 0.671 & 0.262 & 10.295 & 0.683 & 0.223 & 9.169 & 0.697 & 0.177 \\
 & PUP-3DGS & 10.746 & 0.672 & 0.249 & 9.695 & 0.683 & 0.206 & 8.424 & 0.695 & 0.157 \\
 & Ours & \textbf{20.448} & \textbf{0.334} & \textbf{0.601} & \textbf{20.307} & \textbf{0.366} & \textbf{0.582} & \textbf{20.049} & \textbf{0.408} & \textbf{0.556} \\ \cmidrule(lr){1-11}
\multirow{4}{*}{\rotatebox[origin=c]{90}{9 Views}} 
 & EAGLES & 11.384 & 0.662 & 0.289 & 9.813 & 0.678 & 0.234 & 8.147 & 0.695 & 0.159 \\
 & LightGauss. & 13.252 & 0.652 & 0.325 & 12.272 & 0.665 & 0.294 & 10.874 & 0.682 & 0.244 \\
 & PUP-3DGS & 12.236 & 0.654 & 0.305 & 10.753 & 0.672 & 0.260 & 9.178 & 0.686 & 0.198 \\
 & Ours & \textbf{19.678} & \textbf{0.354} & \textbf{0.569} & \textbf{19.582} & \textbf{0.381} & \textbf{0.555} & \textbf{19.437} & \textbf{0.413} & \textbf{0.536} \\ \cmidrule(lr){1-11}
\multirow{4}{*}{\rotatebox[origin=c]{90}{16 Views}} 
 & EAGLES & 6.460 & 0.683 & 0.074 & 5.773 & 0.711 & 0.031 & 9.002 & 0.692 & 0.206 \\
 & LightGauss. & 14.580 & 0.646 & 0.366 & 13.806 & 0.661 & 0.343 & 12.375 & 0.678 & 0.295 \\
 & PUP-3DGS & 13.847 & 0.655 & 0.353 & 12.399 & 0.664 & 0.314 & 10.493 & 0.682 & 0.252 \\
 & Ours & \textbf{19.125} & \textbf{0.392} & \textbf{0.524} & \textbf{19.082} & \textbf{0.410} & \textbf{0.512} & \textbf{18.990} & \textbf{0.435} & \textbf{0.498} \\ \cmidrule(lr){1-11}
\multirow{4}{*}{\rotatebox[origin=c]{90}{32 Views}} 
 & EAGLES & 13.944 & 0.665 & 0.373 & 12.828 & 0.672 & 0.346 & 10.817 & 0.691 & 0.276 \\
 & LightGauss. & 14.362 & 0.663 & 0.375 & 13.835 & 0.669 & 0.361 & 12.531 & 0.683 & 0.322 \\
 & PUP-3DGS & 14.335 & 0.661 & 0.376 & 13.630 & 0.668 & 0.355 & 11.938 & 0.685 & 0.312 \\
 & Ours & \textbf{17.677} & \textbf{0.446} & \textbf{0.476} & \textbf{17.653} & \textbf{0.461} & \textbf{0.467} & \textbf{17.607} & \textbf{0.481} & \textbf{0.456} \\ \bottomrule
\end{tabular}
}
\end{table*}

\subsection{Comparison with Baselines}
\paragraph{\textbf{Quantitative Analysis}}
Tab. \ref{tab:results_acid_re10k} presents a detailed comparison of \PaperTitle{}  against established baselines on the ACID and RE10K datasets with finetuning for sparse view input (additional evaluations without finetuning included in supplementary) and Tab. \ref{tab:results_dl3dv} presents results on the DL3DV dataset with finetuning on dense view (additional evaluations without finetuning included in supplementary). Performance is reported across $\beta=\{ 0.4, 0.6, 0.8\}$. 
Our method consistently outperforms existing pruning strategies across all datasets and $\beta$ values. On ACID, \PaperTitle{}  improves PSNR from $19.43dB$ (EAGLES) to $22.55dB$ at $\beta=0.4$, while reducing LPIPS from $0.620$ to $0.299$. On RE10K, it achieves $22.29$ PSNR compared to $16.60$ from LightGaussian and $16.53$ from EAGLES. Improvements are even larger on DL3DV, for example with $6$ views and $\beta=0.4$, \PaperTitle{} reaches $20.45dB$ PSNR versus $11.36dB$ from LightGaussian. Notably, while baseline methods degrade significantly at high pruning ($\beta=0.8$), our method maintains strong performance across all view counts.

Overall, the trend remains the same, highlighting the superiority of the proposed approach. We report only the best-performing baselines here, while the remaining baseline comparisons, wider pruning ranges, ablation studies (with and without fine-tuning across various backbones), and cross-domain evaluation on the DTU dataset are deferred to the supplementary material.

\paragraph{\textbf{Qualitative Analysis}} We present a qualitative comparison of our method in Fig. \ref{fig:pruning_comparison_combined} and Fig. \ref{fig:comparison_all_baseline_all_pruning_acid}. In all these results, we highlight two important issues with the baseline methods: 
(a) For non-finetuned cases, ``black patch" artifacts can be consistently seen, which is indicative of empty spaces within the 3D scene where Gaussians were removed without any readjustment. Because of ``black patches" artifacts, we see lower metrics across all baselines compared, whereas our results show consistent results without any such artifacts and simultaneously provide high fidelity.
(b) For finetuned cases, results struggle with ``blobby Gaussian'' artifact. This happens as existing pruning methods completely remove all Gaussians from a local region without replacing them. To compensate for the empty space, Gaussians from neighboring areas must expand, which distorts their shape and negatively impacts the quality of the overall representation. This effect persists even after further finetuning, as the fundamental problem of sub-optimal Gaussian allocation remains.

We also show effects of increasing $\beta$ in Fig. \ref{fig:comparison_all_baseline_all_pruning_acid}, where we pick three broad pruning rates ($\beta= \{0.4, 0.6, 0.8 \}$) and compare \PaperTitle{} with the baselines. As can be seen, baselines degrade the scene quality and completely fail to adapt existing Gaussian attributes at higher $\beta$ value. In contrast, \PaperTitle{}~optimally re-adjusts 3D Gaussians to adapt to higher pruning rates, consistently maintaining the high scene quality. We show more qualitative results in the supplementary.

\subsection{Ablation Study}

\cref{tab:withoutmask-ablation_results} evaluates the contribution of key components of our method at a fixed pruning budget $\beta=0.4$. Our analysis reveals that removing the Adaptive GS Head training significantly degrades performance, reducing PSNR from $22.55$ dB to $17.44$ dB on ACID and from $22.29$ dB to $19.81$ dB on RE10K, which indicates that learning to adapt Gaussian attributes post-pruning is critical for recovering reconstruction quality. Similarly, excluding the DWT-based texture signal leads to noticeable performance decreases specifically from $22.55$ dB to $21.31$ dB on ACID confirming the importance of texture guidance in ranking SuperClusters and allocating Gaussians effectively. Finally, removing the contextual mask further degrades RE10K performance to $20.06$ dB, highlighting its vital role in providing spatial context for the Adaptive Gaussian Head. The impact of hyperparameters $K$ and $\gamma$ is evaluated in \cref{tab:ablation_k_and_gamma_horizontal}, where increasing $K$ generally improves reconstruction quality by producing finer SuperClusters that better identify low-texture regions and enable more precise allocation of larger Gaussians. However, these quality gains become marginal beyond $K=300$ while training costs continue to rise. Regarding the retention ratio, we observe that retaining $\gamma=0.10$ of Gaussians achieves the best performance, particularly under high pruning scenarios such as $\beta=0.9$. Larger $\gamma$ values tend to introduce redundant Gaussians in the same regions, which ultimately degrades reconstruction quality, consistent with observations in prior works~\cite{lightgaussian, minisplatting, eagles_compression}. Overall, these results demonstrate that each individual component contributes significantly to the final performance, with their combination yielding the highest reconstruction quality across both the ACID and RE10K datasets.

\begin{table}[t]
\centering
\small
\setlength{\tabcolsep}{3pt}
\caption{\textbf{Ablation study.} We evaluate the impact of training the GS Head, texture features, and the contextual mask at a fixed budget of $\beta=0.4$. }
\resizebox{0.9\linewidth}{!}{
\begin{tabular}{l ccc ccc}
\toprule
& \multicolumn{3}{c}{ACID} & \multicolumn{3}{c}{RE10K} \\
\cmidrule(lr){2-4} \cmidrule(lr){5-7}
Ablation Variant & \textbf{PSNR$\uparrow$} & \textbf{LPIPS$\downarrow$} & \textbf{SSIM$\uparrow$} & \textbf{PSNR$\uparrow$} & \textbf{LPIPS$\downarrow$} & \textbf{SSIM$\uparrow$} \\
\midrule
w/o Texture            & 21.31 & 0.326 & 0.623 & 20.73 & 0.260 & 0.713 \\ 
w/o Adaptive GS Head   & 17.44 & 0.445 & 0.490 & 19.81 & 0.399 & 0.617 \\ 
w/o Binary Mask    & 22.52 & 0.318 & \textbf{0.642} & 20.06 & 0.339 & 0.656 \\
\midrule
\textbf{Ours} & \textbf{22.55} & \textbf{0.299} & 0.639 & \textbf{22.29} & \textbf{0.235} & \textbf{0.735} \\
\bottomrule
\end{tabular}
}

\label{tab:withoutmask-ablation_results}
\end{table}

\begin{table}[h]
\centering
\caption{Effect of the number of clusters $k$ (left) and retention percentage $\gamma$ (right) on reconstruction quality across scenes.}
\setlength{\tabcolsep}{4pt}
\resizebox{\linewidth}{!}{
\begin{tabular}{@{}lccc|cccc@{}}
\toprule
\multicolumn{4}{c|}{\textbf{A. Varying number of clusters $k$}} & \multicolumn{4}{c}{\textbf{B. Varying retention percentage $\gamma$}} \\ \cmidrule(r){1-4} \cmidrule(l){5-8}
\textbf{Config} & \textbf{PSNR$\uparrow$} & \textbf{LPIPS$\downarrow$} & \textbf{SSIM$\uparrow$} & \textbf{Config} & \textbf{PSNR$\uparrow$} & \textbf{LPIPS$\downarrow$} & \textbf{SSIM$\uparrow$} \\ \midrule
$k=50$  & 22.416 & 0.303 & 0.636 & $\gamma = 0.05$ & 21.796 & 0.347 & 0.608 \\
$k=100$ & 22.488 & 0.301 & 0.638 & $\gamma = 0.10$ & \textbf{22.549} & \textbf{0.299} & \textbf{0.640} \\
$k=300$ & 22.549 & 0.299 & 0.640 & $\gamma = 0.15$ & 21.587 & 0.352 & 0.589 \\
$k=400$ & \textbf{22.559} & \textbf{0.298} & \textbf{0.641} & $\gamma = 0.20$ & 22.262 & 0.304 & 0.629 \\ \bottomrule
\end{tabular}
}
\label{tab:ablation_k_and_gamma_horizontal}
\end{table}

\subsection{Qualitative Texture Analysis}
We provide additional analysis focusing on scene texture preservation. 
Our method explicitly targets low-texture regions for pruning, allowing the model to allocate fewer but larger Gaussians in such areas while preserving fine details in highly textured regions. ~\cref{fig:anysplat_9view_results} presents a qualitative comparison of texture preservation on a representative scene across different pruning ratios. We observe that baseline methods tend to oversmooth textured regions under pruning, leading to noticeable loss of high-frequency details.
In comparison, our approach better maintains scene fidelity by leveraging texture-aware clustering and natural scene statistics. 
Furthermore, Fig.~\ref{fig:anysplat_texture_results} provides additional qualitative comparisons across multiple scenes at $\beta = 0.4$, demonstrating that while baselines suffer from blurred edges, over-smoothing, and structural distortion, our method maintains sharp high-frequency details and structural integrity.
\begin{table}[H]

\hfill
\begin{minipage}{0.48\linewidth}
\centering
    \includegraphics[width=\linewidth]{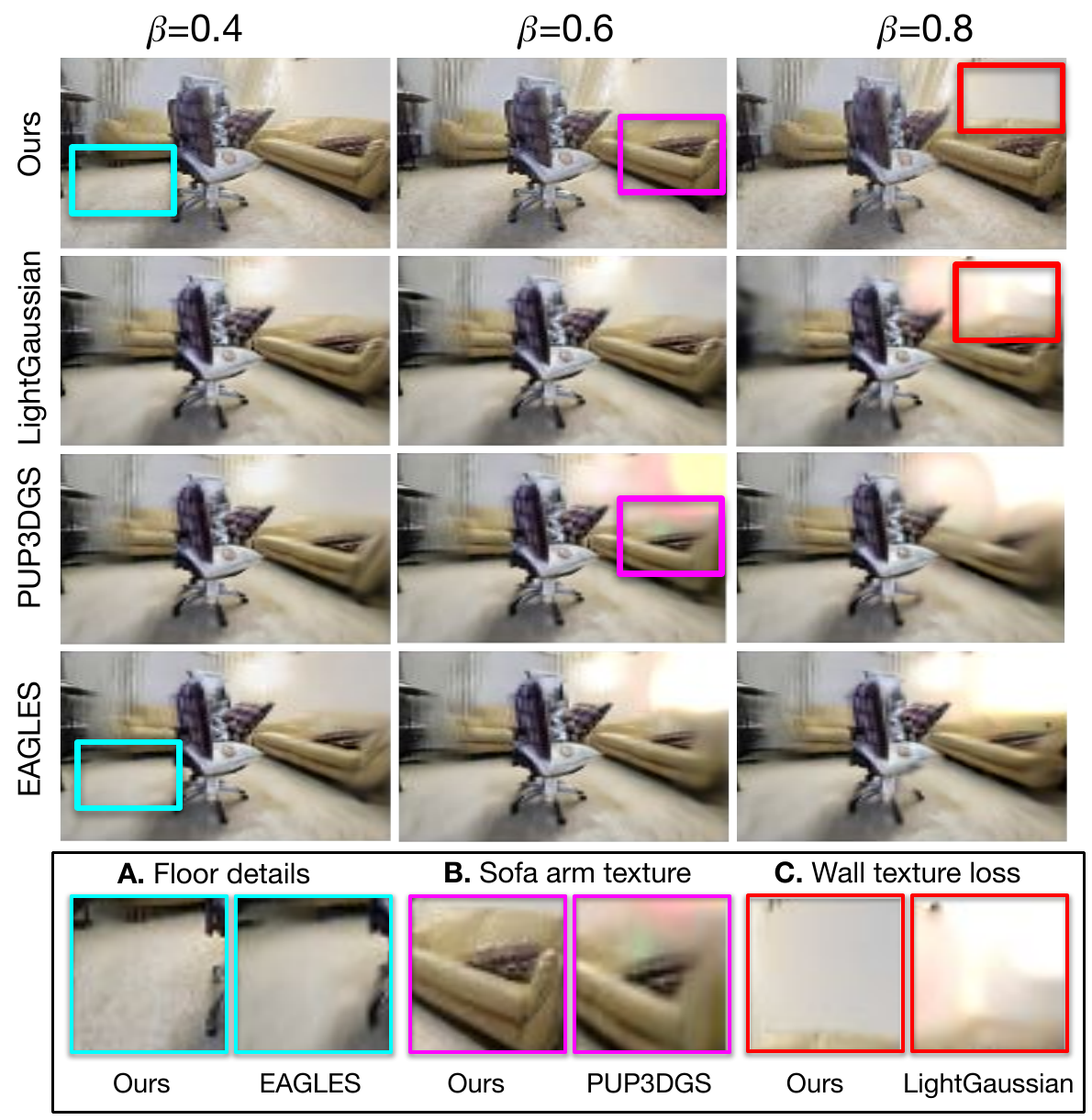}
    \captionof{figure}{\textbf{Texture Analysis (fixed scene)}: Baselines smooth out the textures whereas we retain texture of the scene across the regions.}
    \label{fig:anysplat_9view_results}
\end{minipage}
\hfill
\begin{minipage}{0.48\linewidth}
\centering
    \includegraphics[width=\linewidth]{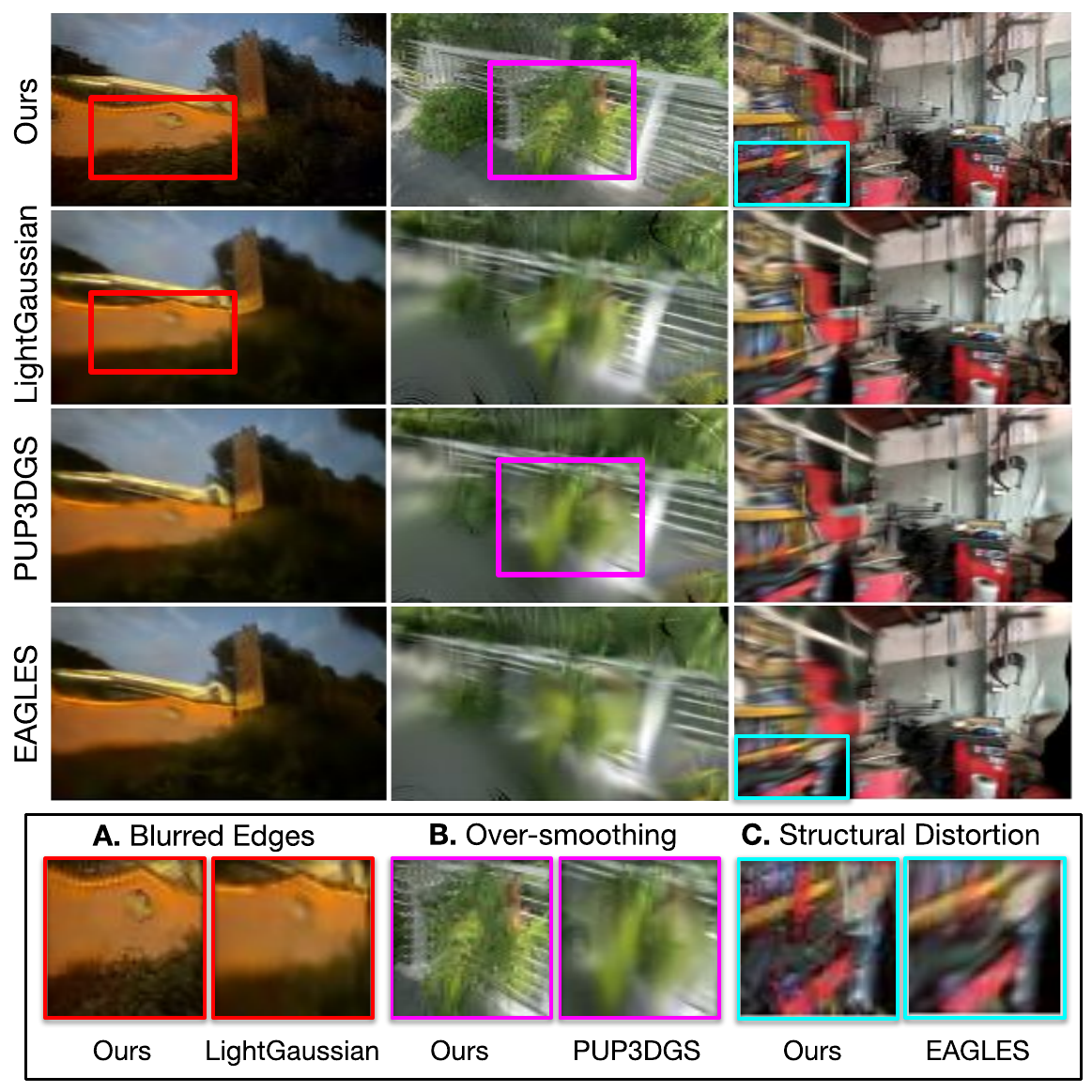}
    \captionof{figure}{\textbf{Texture Analysis (varying scene)}: Across the scenes we observe oversmoothed regions whereas we retain the texture of the scene.}
    \label{fig:anysplat_texture_results}
\end{minipage}
\vspace{-10pt}
\end{table}
These results consistently demonstrate that competing methods introduce excessive smoothing in high-texture areas, whereas our method preserves structural details more effectively under increasing pruning strengths.
\cref{fig:scale_statistics} shows the histogram of mean scale values demonstrating our insight empirically that mean scale values of 3D Gaussians corresponding to low texture regions are predicted larger in our method to retain the quality of scene; thus, redundant 3D Gaussians can be pruned.

\begin{table}[H]
\hfill
\begin{minipage}{0.48\linewidth}
\centering
    \includegraphics[width=\linewidth]{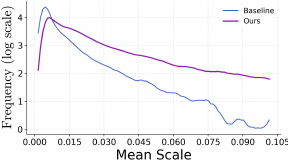}
    \captionof{figure}{\textbf{Plot of mean scale values with $log_{10}$ frequency:} Predicted 3D Gaussians from our method are larger than traditional 3D Gaussian representation.}
    \label{fig:scale_statistics}
\end{minipage}
\hfill
\begin{minipage}{0.48\linewidth}
\centering
    \includegraphics[width=\linewidth]{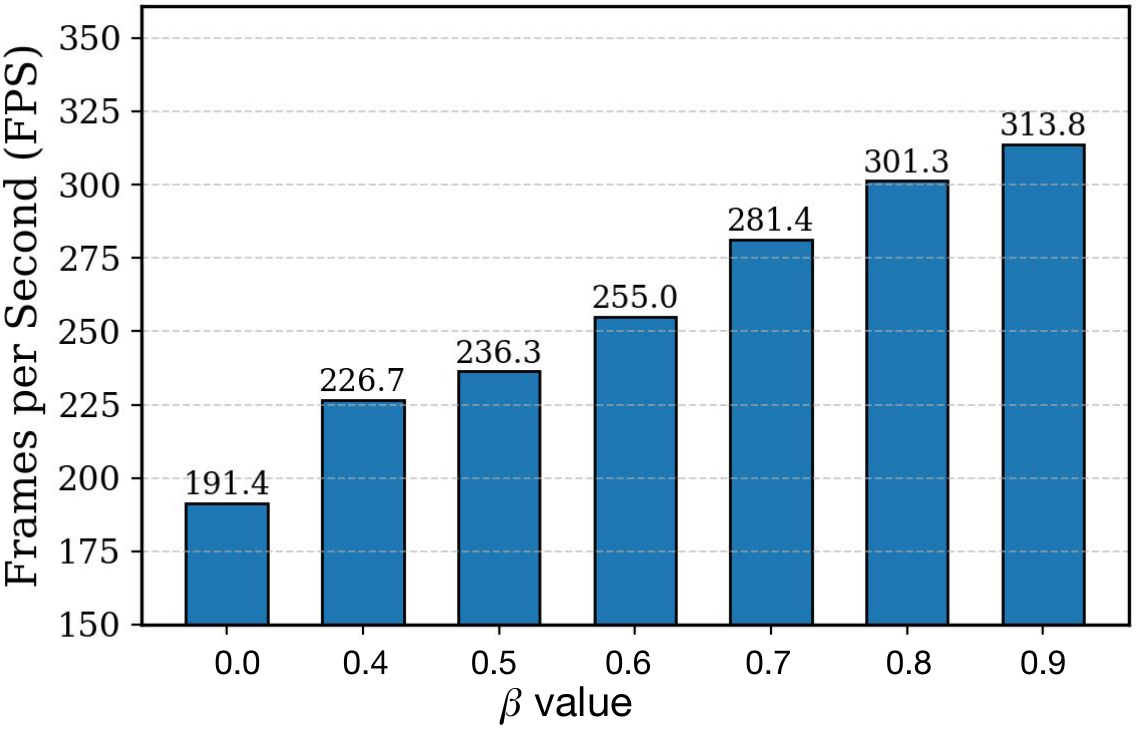}
    \captionof{figure}{Plot of rendering speed in Frame Per Second (FPS) with increasing pruning strength.}
    \label{fig:FPS_plot_mainpaper}
\end{minipage}
\vspace{-5mm}
\end{table}

\subsection{Inference Time Analysis}
\setlength{\intextsep}{0pt}
\begin{figure}[H]
\begin{minipage}[c]{0.48\linewidth}
Fig.~\ref{fig:FPS_plot_mainpaper} summarizes the FPS achieved across pruning strengths. As pruning strength increases, the reduced number of active Gaussians leads to faster rendering speeds and lower computational overhead, demonstrating the effectiveness of pruning for balancing reconstruction quality and runtime efficiency. We use the same GPU memory load as the base model~\cite{noposplat, jiang2025anysplat}. Fig.~\ref{fig:pareto_mainpaper} presents a Pareto plot comparing inference time and quality of ours with other methods. Our method occupies the top-left region, achieving the highest PSNR at substantially lower latency than the finetuned baselines. Further, we include additional discussions and comparisons on inference latency analysis of ours with baselines in the supplementary.
\end{minipage}
\hfill
\begin{minipage}[c]{0.48\linewidth}
\centering
\includegraphics[width=\linewidth]{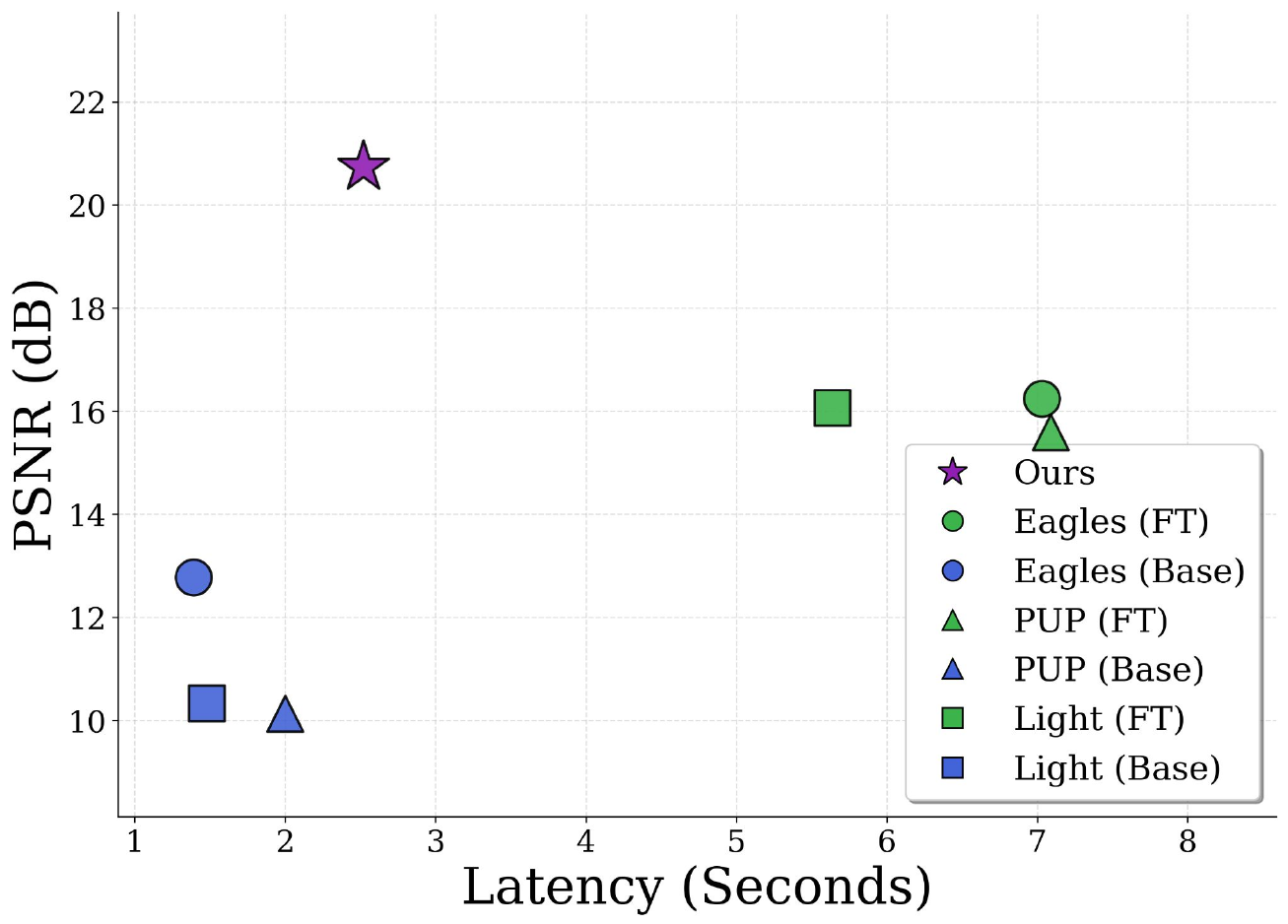}
\captionof{figure}{Pareto Plot for inference time and quality comparison for RE10K at $\beta=0.8$}
\label{fig:pareto_mainpaper}
\end{minipage}
\end{figure}

\section{Conclusions and Future Work}
We introduce \textbf{\PaperTitle}, a novel, feed-forward framework for controllable 3D Gaussian pruning. Our method's core novelty is texture-aware Gaussian control that significantly reduces primitives by targeting low-texture regions. By eliminating the need for expensive per-scene optimization, \PaperTitle{} provides a practical solution for controllable 3D reconstruction, achieving a superior quality-efficiency trade-off even at extreme pruning rates. On smooth surfaces with high-frequency texture (such as intricate wall patterns) or specular highlights, our method suffers from the same limitation as any pixel-aligned feed-forward model. We briefly discuss these limitations in the supplementary.
Future work will focus on integrating a more nuanced understanding of both texture and geometry to enable more robust pruning decisions, paving the way for truly efficient 3D scene representation.

\section*{Acknowledgement}
We sincerely thank Anudeep for careful review of our manuscript and for providing valuable feedback and suggestions that significantly improved the quality of the article. This work is supported by Google and Kotak IISc AI-ML Centre (KIAC). Badrinath Singhal is supported by TCS Research Scholarship and Qualcomm Innovation Fellowship (QIF).

\clearpage

\bibliographystyle{splncs04}
\bibliography{main}
\clearpage

\clearpage
\appendix
% \cleardoublepage
\pagenumbering{Alph} % Sets page anchors to A, B, C for hyperref
\renewcommand{\thesection}{\Alph{section}}
\renewcommand{\thefigure}{\Alph{figure}}
\renewcommand{\thetable}{\Alph{table}}

% CRUCIAL FIX FOR HYPERREF LINKS:
\renewcommand{\theHsection}{Appendix.\arabic{section}} 
\renewcommand{\theHfigure}{Appendix.\arabic{figure}}
\renewcommand{\theHtable}{Appendix.\arabic{table}}

\setcounter{section}{0}
\setcounter{figure}{0}
\setcounter{table}{0}
\setcounter{page}{1}
% \maketitle

% --- Appendix layout: avoid weird stretched gaps and pack floats tightly ---
% Collect leftover vertical space at the bottom of a page instead of
% distributing it as stretched glue between paragraphs/headings/floats.
\raggedbottom
% Allow more / larger floats per page and let them occupy more of the page
% so figures and tables pack instead of leaving half-empty pages.
\setcounter{topnumber}{4}
\setcounter{bottomnumber}{4}
\setcounter{totalnumber}{8}
\renewcommand{\topfraction}{0.92}
\renewcommand{\bottomfraction}{0.92}
\renewcommand{\textfraction}{0.06}
\renewcommand{\floatpagefraction}{0.70}
\renewcommand{\dbltopfraction}{0.92}
\renewcommand{\dblfloatpagefraction}{0.70}
% On float-only pages, pack floats to the top (leftover space collects at the
% bottom) instead of LaTeX's default behaviour of spreading them out with
% stretchable glue, which produced even but distracting mid-page gaps.
\makeatletter
\setlength{\@fptop}{0pt}
\setlength{\@fpsep}{10pt plus 0fil}
\setlength{\@fpbot}{0pt plus 1fil}
\makeatother
% --------------------------------------------------------------------------

\section*{\centering Supplementary Material: AdaptiveSplat: Texture Aware Controllable 3D Gaussian Allocation for Feed-Forward Reconstruction}

\section*{Table of Contents}
\noindent
\begin{tabular}{@{}p{0.38\linewidth} p{0.58\linewidth}@{}}
\toprule
\textbf{Section} & \textbf{Description} \\
\midrule
\hyperref[sec:additional_baselines_supplementary]{A. Additional Baselines} & 
We created additional baselines for comparisons \\
\addlinespace[0.3em]
\hyperref[sec:dense_view_supplementary]{B. Extension to Dense View Input} & 
Extension of our method for dense view and comparison with baselines \\
\addlinespace[0.3em]
\hyperref[sec:inference_latency_supplementary]{C. Inference Latency} & 
Inference time analysis of our method with baselines \\
\addlinespace[0.3em]
\hyperref[sec:naive_pruning_results]{D. Results with Naive Pruning Baselines} & 
Results with opacity-based and low-resolution naive pruning \\
\addlinespace[0.3em]
\hyperref[sec:results-on-different-backbone]{E. Results on Different Backbones} & 
Detailed RE10K and ACID results across pruning stages on NoPoSplat, HiSplat, pixelSplat, and MVSplat \\
\addlinespace[0.3em]
\hyperref[sec:cross-doman-eval]{F. Cross Domain Evaluation} & 
Train on ACID/RE10K, evaluate on DTU for out-of-distribution generalization \\
\addlinespace[0.3em]
\hyperref[sec:failure_case]{G. Failure Case} & 
DWT limitation under specular highlights and light-induced high frequencies \\
\addlinespace[0.3em]
\hyperref[sec:additional_ablation_supplementary]{H. Additional Ablation} & 
Detailed ablation results on RE10K, ACID for evaluating the importance of each component \\
\bottomrule
\end{tabular}

\section{Additional Baselines}
\label{sec:additional_baselines_supplementary}
We implemented additional baselines where pruning masks are first generated using existing pruning pipelines and then concatenated with the Gaussian Head input to predict the attributes of the retained Gaussians. The Gaussian Head for these baselines is trained following the same procedure and loss functions as in our method to ensure fairness.
\begin{table}[!tb]
\centering
\setlength{\tabcolsep}{8pt}
\caption{\textbf{Results on Additional Baselines on ACID dataset:} We created additional baselines where the mask is created using the pruning methods and used in Gaussian Head which predicts the attributes of retained Gaussians. We consistently outperform the baselines across all pruning strengths by a significant margin.}
\resizebox{\linewidth}{!}{
\begin{tabular}{@{}l|lll|lll|lll@{}}
\toprule
                & \multicolumn{3}{c|}{\textbf{$\beta=0.4$}}                 & \multicolumn{3}{c|}{\textbf{$\beta=0.6$}}                 & \multicolumn{3}{c}{\textbf{$\beta=0.9$}}                  \\ \midrule
\textbf{Method} & \textbf{PSNR$\uparrow$}   & \textbf{LPIPS$\downarrow$} & \textbf{SSIM$\uparrow$}  & \textbf{PSNR$\uparrow$}   & \textbf{LPIPS$\downarrow$} & \textbf{SSIM$\uparrow$}  & \textbf{PSNR$\uparrow$}   & \textbf{LPIPS$\downarrow$} & \textbf{SSIM$\uparrow$}  \\ \midrule
EAGLES          & 18.747          & 0.480          & 0.512          & 18.716          & 0.488          & 0.514          & 16.514          & 0.547          & 0.463          \\
LightGaussian   & 19.439          & 0.453          & 0.519          & 19.176          & 0.476          & 0.507          & 14.333          & 0.572          & 0.356          \\
PUP-3DGS        & 17.912          & 0.555          & 0.503          & 17.193          & 0.578          & 0.492          & 13.177          & 0.662          & 0.384          \\
Ours            & \textbf{22.549} & \textbf{0.299} & \textbf{0.640} & \textbf{22.310} & \textbf{0.310} & \textbf{0.627} & \textbf{18.476} & \textbf{0.379} & \textbf{0.553} \\ \bottomrule
\end{tabular}
}

\label{tab:additional-baseline-acid}
\end{table}

We evaluate these baselines on both the ACID and RE10K datasets, and the results are provided in Tables \ref{tab:additional-baseline-acid} and \ref{tab:additional-baseline-re10k}. These enhanced baselines achieve stronger results than the post-hoc pruning baselines reported in the main paper, particularly at higher pruning ratios. However, our method still consistently outperforms them across all pruning levels and datasets, primarily due to our texture-aware region selection strategy, which better preserves scene structure under aggressive pruning.

\begin{table}[!tb]
\centering
\setlength{\tabcolsep}{8pt}
\caption{\textbf{Results on Additional Baselines on RE10K dataset:} We've implemented additional baseline where we create mask using the existing pruning methods and append to Gaussian Head which predicts attributes of retained Gaussians. We consistently outperform the baselines by a significant margin across all pruning strengths.}
\resizebox{\linewidth}{!}{
\begin{tabular}{@{}l|lll|lll|lll@{}}
\toprule
                & \multicolumn{3}{c|}{\textbf{$\beta=0.4$}}                 & \multicolumn{3}{c|}{\textbf{$\beta=0.6$}}                 & \multicolumn{3}{c}{\textbf{$\beta=0.9$}}                  \\ \midrule
\textbf{Method} & \textbf{PSNR$\uparrow$}   & \textbf{LPIPS$\downarrow$} & \textbf{SSIM$\uparrow$}  & \textbf{PSNR$\uparrow$}   & \textbf{LPIPS$\downarrow$} & \textbf{SSIM$\uparrow$}  & \textbf{PSNR$\uparrow$}   & \textbf{LPIPS$\downarrow$} & \textbf{SSIM$\uparrow$}  \\ \midrule
EAGLES          & 17.828          & 0.412          & 0.555          & 17.726          & 0.421          & 0.551          & 14.437          & 0.514          & 0.435          \\
LightGaussian   & 17.964          & 0.395          & 0.545          & 17.599          & 0.416          & 0.531          & 13.004          & 0.534          & 0.394          \\
PUP-3DGS        & 15.519          & 0.585          & 0.501          & 15.276          & 0.602          & 0.490          & 11.794          & 0.666          & 0.387          \\
Ours            & \textbf{22.294} & \textbf{0.235} & \textbf{0.735} & \textbf{22.110} & \textbf{0.243} & \textbf{0.726} & \textbf{17.848} & \textbf{0.314} & \textbf{0.645} \\ \bottomrule
\end{tabular}
}

\label{tab:additional-baseline-re10k}
\end{table}

It is also important to note that these additional baselines are not feed-forward: they require (1) predicting the full set of 3D Gaussians, (2) generating a pruning mask, and only then (3) predicting the attributes of the retained Gaussians in a second pass. This makes them inherently two-stage. In contrast, our approach performs pruning within the feed-forward generation pipeline, offering controllable pruning during inference.

\begin{table}[!tb]
\centering
\small
\setlength{\tabcolsep}{2pt}
\caption{\textbf{Extension to a posed feed-forward backbone (MVSplat) on ACID.} All entries use the MVSplat backbone on the identical ACID test split; \textbf{Ours} adds the Adaptive Gaussian Head trained with mask conditioning. Vanilla MVSplat produces a fixed Gaussian count ($\beta=0$); we additionally report Ours at $\beta\in\{0,0.4,0.6,0.8\}$.}
\label{tab:results_mvsplat_acid}
\resizebox{\linewidth}{!}{
\begin{tabular}{@{}c|ccc|ccc|ccc|ccc|ccc@{}}
\toprule
 & \multicolumn{3}{c|}{\textbf{MVSplat}} & \multicolumn{12}{c}{\textbf{Ours}} \\
\cmidrule(lr){2-4}\cmidrule(lr){5-16}
 & \multicolumn{3}{c|}{$\beta=0$} & \multicolumn{3}{c|}{$\beta=0$} & \multicolumn{3}{c|}{$\beta=0.4$} & \multicolumn{3}{c|}{$\beta=0.6$} & \multicolumn{3}{c}{$\beta=0.8$} \\
\cmidrule(lr){2-4}\cmidrule(lr){5-7}\cmidrule(lr){8-10}\cmidrule(lr){11-13}\cmidrule(lr){14-16}
Views & \textbf{PSNR$\uparrow$} & \textbf{LPIPS$\downarrow$} & \textbf{SSIM$\uparrow$} & \textbf{PSNR$\uparrow$} & \textbf{LPIPS$\downarrow$} & \textbf{SSIM$\uparrow$} & \textbf{PSNR$\uparrow$} & \textbf{LPIPS$\downarrow$} & \textbf{SSIM$\uparrow$} & \textbf{PSNR$\uparrow$} & \textbf{LPIPS$\downarrow$} & \textbf{SSIM$\uparrow$} & \textbf{PSNR$\uparrow$} & \textbf{LPIPS$\downarrow$} & \textbf{SSIM$\uparrow$} \\
\midrule
2  & 25.31 & 0.23 & 0.76 & \textbf{25.75} & \textbf{0.23} & \textbf{0.77} & 25.37 & 0.25 & 0.76 & 24.91 & 0.26 & 0.75 & 24.18 & 0.29 & 0.73 \\
4  & 22.21 & 0.29 & 0.69 & \textbf{22.58} & \textbf{0.28} & \textbf{0.71} & 22.61 & 0.29 & 0.70 & 22.43 & 0.31 & 0.69 & 22.03 & 0.33 & 0.68 \\
8  & 21.64 & 0.29 & 0.68 & \textbf{22.08} & \textbf{0.29} & \textbf{0.69} & 22.08 & 0.31 & 0.68 & 21.89 & 0.32 & 0.67 & 21.51 & 0.34 & 0.66 \\
16 & 21.91 & 0.29 & 0.69 & \textbf{22.31} & \textbf{0.29} & \textbf{0.70} & 22.32 & 0.30 & 0.69 & 22.14 & 0.315 & 0.68 & 21.77 & 0.33 & 0.68 \\
\bottomrule
\end{tabular}}
\end{table}

MASt3R and VGGT are foundation models that generalize well, so we built on AnySplat and NoPoSplat, which use them as backbones. Our method is pose-agnostic; to demonstrate this, we further extend it to MVSplat, a posed feed-forward method, on ACID dataset. Vanilla MVSplat returns a fixed Gaussian count ($\beta=0$). We evaluate our approach over $\beta\in\{0,0.4,0.6,0.8\}$ (\cref{tab:results_mvsplat_acid}), confirming that it works across both posed and unposed feed-forward methods.

\section{Extension to Dense View input}
\label{sec:dense_view_supplementary}

\begin{table}[!tb]
\centering
\setlength{\tabcolsep}{4pt}
\caption{\textbf{Evaluation of our texture-aware pruning method on InstantSplat across 3, 4, and 10 views inputs.}
For each pruning strength, we apply our SuperCluster-based pruning before InstantSplat optimization and disable densification to keep the Gaussian count fixed. Across all settings, our method consistently preserves significantly higher reconstruction fidelity (PSNR, SSIM) and lower perceptual error (LPIPS) compared to InstantSplat’s native pruning-by-optimization baseline.
}
\resizebox{\linewidth}{!}{
\begin{tabular}{@{}cl|ccc|ccc|ccc@{}}
\toprule
\multicolumn{2}{l|}{Pruning Strength}                           & \multicolumn{3}{c|}{\textbf{$\beta=0.4$}}             & \multicolumn{3}{c|}{\textbf{$\beta=0.6$}}             & \multicolumn{3}{c}{\textbf{$\beta=0.8$}}              \\ \midrule
\multicolumn{1}{l|}{\textbf{Input View}}      & \textbf{Method} & \textbf{PSNR} & \textbf{LPIPS} & \textbf{SSIM} & \textbf{PSNR} & \textbf{LPIPS} & \textbf{SSIM} & \textbf{PSNR} & \textbf{LPIPS} & \textbf{SSIM} \\ \midrule
\multicolumn{1}{c|}{\multirow{2}{*}{3 Views}}  & Baseline        & 21.565        & 0.217          & 0.745         & 19.320        & 0.321          & 0.662         & 15.482        & 0.479          & 0.508         \\
\multicolumn{1}{c|}{}                         & Ours             & 21.536        & 0.215          & 0.762         & 21.464        & 0.261          & 0.746         & 21.484        & 0.276          & 0.747         \\ \midrule
\multicolumn{1}{c|}{\multirow{2}{*}{4 Views}}  & Baseline        & 21.815        & 0.201          & 0.735         & 19.673        & 0.310          & 0.681         & 15.564        & 0.469          & 0.510         \\
\multicolumn{1}{c|}{}                         & Ours             & 23.932        & 0.176          & 0.820         & 23.692        & 0.215          & 0.804         & 23.530        & 0.236          & 0.799         \\ \midrule
\multicolumn{1}{c|}{\multirow{2}{*}{10 Views}} & Baseline        & 24.754        & 0.141          & 0.852         & 20.655        & 0.249          & 0.746         & 14.827        & 0.416          & 0.558         \\
\multicolumn{1}{c|}{}                         & Ours             & 27.649        & 0.113          & 0.896         & 27.310        & 0.138          & 0.886         & 27.178        & 0.154          & 0.882         \\ \bottomrule
\end{tabular}
}

\label{tab:multiview-instantsplat-results}
\end{table}

\begin{table}[!tb]
\centering
\setlength{\tabcolsep}{4pt}
\caption{\textbf{Results with AnySplat on DL3DV dataset: } Here we do not finetune the remaining Gaussians.}
\resizebox{\linewidth}{!}{
\begin{tabular}{@{}llccccccccc@{}}
\toprule
                          & \multicolumn{1}{c}{}                 & \multicolumn{3}{c}{\textbf{$\beta=0.4$}}                                                            & \multicolumn{3}{c}{\textbf{$\beta=0.6$}}                                                            & \multicolumn{3}{c}{\textbf{$\beta=0.8$}}                                       \\ \midrule
      & \multicolumn{1}{l|}{\textbf{Method}} & \textbf{PSNR$\uparrow$} & \textbf{LPIPS$\downarrow$} & \multicolumn{1}{c|}{\textbf{SSIM$\uparrow$}} & \textbf{PSNR$\uparrow$} & \textbf{LPIPS$\downarrow$} & \multicolumn{1}{c|}{\textbf{SSIM$\uparrow$}} & \textbf{PSNR$\uparrow$} & \textbf{LPIPS$\downarrow$} & \textbf{SSIM$\uparrow$} \\ \midrule
\multirow{4}{*}{\rotatebox[origin=c]{90}{6 Views}}   & \multicolumn{1}{l|}{EAGLES}          & 5.892                   & 0.709                      & \multicolumn{1}{c|}{0.033}                   & 5.577                   & 0.723                      & \multicolumn{1}{c|}{0.013}                   & 5.413                   & 0.725                      & 0.006                   \\
                          & \multicolumn{1}{l|}{LightGaussian}   & 6.377                   & 0.673                      & \multicolumn{1}{c|}{0.076}                   & 5.978                   & 0.702                      & \multicolumn{1}{c|}{0.044}                   & 5.613                   & 0.722                      & 0.017                   \\
                          & \multicolumn{1}{l|}{PUP3DGS}         & 6.151                   & 0.690                      & \multicolumn{1}{c|}{0.054}                   & 5.771                   & 0.709                      & \multicolumn{1}{c|}{0.029}                   & 5.514                   & 0.720                      & 0.012                   \\
                          & \multicolumn{1}{l|}{Ours}            & \textbf{20.448}         & \textbf{0.334}             & \multicolumn{1}{c|}{\textbf{0.601}}          & \textbf{20.307}         & \textbf{0.366}             & \multicolumn{1}{c|}{\textbf{0.582}}          & \textbf{20.049}         & \textbf{0.408}             & \textbf{0.556}          \\ \midrule
\multirow{4}{*}{\rotatebox[origin=c]{90}{9 Views}}   & \multicolumn{1}{l|}{EAGLES}          & 6.053                   & 0.703                      & \multicolumn{1}{c|}{0.044}                   & 5.644                   & 0.720                      & \multicolumn{1}{c|}{0.018}                   & 5.441                   & 0.726                      & 0.008                   \\
                          & \multicolumn{1}{l|}{LightGaussian}   & 6.689                   & 0.656                      & \multicolumn{1}{c|}{0.099}                   & 6.180                   & 0.688                      & \multicolumn{1}{c|}{0.06}                    & 5.733                   & 0.717                      & 0.026                   \\
                          & \multicolumn{1}{l|}{PUP3DGS}         & 6.438                   & 0.675                      & \multicolumn{1}{c|}{0.078}                   & 5.926                   & 0.702                      & \multicolumn{1}{c|}{0.041}                   & 5.587                   & 0.718                      & 0.018                   \\
                          & \multicolumn{1}{l|}{Ours}            & \textbf{19.678}         & \textbf{0.354}             & \multicolumn{1}{c|}{\textbf{0.569}}          & \textbf{19.582}         & \textbf{0.381}             & \multicolumn{1}{c|}{\textbf{0.555}}          & \textbf{19.437}         & \textbf{0.413}             & \textbf{0.536}          \\ \midrule
\multirow{4}{*}{\rotatebox[origin=c]{90}{16 Views}}  & \multicolumn{1}{l|}{EAGLES}          & 6.139                   & 0.699                      & \multicolumn{1}{c|}{0.047}                   & 5.689                   & 0.719                      & \multicolumn{1}{c|}{0.019}                   & 5.481                   & 0.726                      & 0.007                   \\
                          & \multicolumn{1}{l|}{LightGaussian}   & 7.133                   & 0.639                      & \multicolumn{1}{c|}{0.124}                   & 6.499                   & 0.673                      & \multicolumn{1}{c|}{0.081}                   & 5.899                   & 0.709                      & 0.036                   \\
                          & \multicolumn{1}{l|}{PUP3DGS}         & 6.796                   & 0.661                      & \multicolumn{1}{c|}{0.098}                   & 6.126                   & 0.694                      & \multicolumn{1}{c|}{0.051}                   & 5.651                   & 0.717                      & 0.018                   \\
                          & \multicolumn{1}{l|}{Ours}            & \textbf{19.125}         & \textbf{0.392}             & \multicolumn{1}{c|}{\textbf{0.524}}          & \textbf{19.082}         & \textbf{0.410}             & \multicolumn{1}{c|}{\textbf{0.512}}          & \textbf{18.990}         & \textbf{0.435}             & \textbf{0.498}          \\ \midrule
\multirow{4}{*}{\rotatebox[origin=c]{90}{32 Views}}  & \multicolumn{1}{l|}{EAGLES}          & 6.460                   & 0.683                      & \multicolumn{1}{c|}{0.074}                   & 5.773                   & 0.711                      & \multicolumn{1}{c|}{0.031}                   & 5.426                   & 0.723                      & 0.011                   \\
                          & \multicolumn{1}{l|}{LightGaussian}   & 7.482                   & 0.629                      & \multicolumn{1}{c|}{0.147}                   & 6.826                   & 0.657                      & \multicolumn{1}{c|}{0.103}                   & 6.038                   & 0.697                      & 0.040                   \\
                          & \multicolumn{1}{l|}{PUP3DGS}         & 7.239                   & 0.642                      & \multicolumn{1}{c|}{0.129}                   & 6.388                   & 0.682                      & \multicolumn{1}{c|}{0.073}                   & 5.665                   & 0.714                      & 0.025                   \\
                          & \multicolumn{1}{l|}{Ours}            & \textbf{17.677}         & \textbf{0.446}             & \multicolumn{1}{c|}{\textbf{0.476}}          & \textbf{17.653}         & \textbf{0.461}             & \multicolumn{1}{c|}{\textbf{0.467}}          & \textbf{17.607}         & \textbf{0.481}             & \textbf{0.456}          \\ \bottomrule
\end{tabular}
}

\label{tab:results_withoutfinetuning_anysplatt_comparison_dl3dv}
\end{table}

We additionally evaluate our texture-aware pruning strategy on the InstantSplat framework. We used Tanks and Temples dataset for our experiment, specifically we used the `Horse' scene from Tanks and Temples. We first use MASt3R to obtain the pixel-aligned point cloud from sparse-view images, construct texture-based SuperClusters, and rank them using our wavelet textureness metric. We prune points according to the desired pruning strength and then train the 3DGS stage for 1000 iterations. Importantly, we disable densification and pruning within InstantSplat to ensure that the Gaussian count remains fixed throughout the optimization pipeline. 

We report results for 3, 4, and 10-view input settings and evaluate on novel views (Tab. ~\ref{tab:multiview-instantsplat-results}). Our method consistently achieves higher PSNR, SSIM, and lower LPIPS across all pruning strengths and view configurations. This demonstrates that the proposed texture-aware pruning not only preserves reconstruction quality but also enhances efficiency in multiview sparse-input pipelines without requiring any architectural changes to InstantSplat.

Additionally we provide results for dense view feed forward  in \cref{tab:results_withoutfinetuning_anysplatt_comparison_dl3dv} on DL3DV where we do not finetune the remaining Gaussians. The results are provided for 6, 9, 16 and 32 input views using VGGT as multiview feature extractor. The corresponding results for fine tune are included in main paper. \cref{tab:dense_dl3dv} reports the pretrained AnySplat baseline against Ours at $\beta=0$ on DL3DV across 9, 16, and 32 input views.

\begin{table}[!tb]
\centering
\caption{\textbf{Dense-view comparison on DL3DV.} We report pretrained AnySplat vs.\ Ours at $\beta=0$ across 9, 16, and 32 input views.}
\label{tab:dense_dl3dv}
\setlength{\tabcolsep}{10pt}
\begin{tabular}{@{}ll|ccc@{}}
\toprule
Views & Method & \textbf{PSNR$\uparrow$} & \textbf{LPIPS$\downarrow$} & \textbf{SSIM$\uparrow$} \\
\midrule
\multirow{2}{*}{9}  & AnySplat & \textbf{19.83} & \textbf{0.301} & \textbf{0.601} \\
                    & Ours     & 19.80          & 0.316          & 0.587 \\
\midrule
\multirow{2}{*}{16} & AnySplat & 18.87          & \textbf{0.362} & 0.535 \\
                    & Ours     & \textbf{19.17} & 0.364          & \textbf{0.537} \\
\midrule
\multirow{2}{*}{32} & AnySplat & 17.24          & \textbf{0.420} & 0.486 \\
                    & Ours     & \textbf{17.71} & 0.424          & \textbf{0.489} \\
\bottomrule
\end{tabular}
\end{table}

\section{Inference Latency}
\label{sec:inference_latency_supplementary}
To highlight the practical benefits of controllable pruning, we report rendering throughput under different pruning strengths. We adopt the InstantSplat setting for sparse-view reconstruction, where MASt3R provides the initial pixel-aligned points and the 3D Gaussian attributes are subsequently optimized. To ensure a fair comparison, we disable densification so that the number of Gaussians remains fixed to the MASt3R initialization. We used the Tanks and Temples dataset specifically the Horse scene for this analysis.

\begin{table}[!tb]
\centering
\setlength{\tabcolsep}{14pt}
\caption{\textbf{Rendering speed vs. pruning strength}. FPS increases steadily as more Gaussians are pruned while maintaining scene fidelity, demonstrating the utility of pruning as a mechanism for real-time, controllable efficiency during inference. The PSNR was obtained for each pruning strength on Tanks and Temples Horse scene.}
\resizebox{\linewidth}{!}{
\begin{tabular}{@{}l|ccccccc@{}}
\toprule
\textbf{Pruning Strength ($\beta$)} & \textbf{0.0} & \textbf{0.4} & \textbf{0.5} & \textbf{0.6} & \textbf{0.7} & \textbf{0.8} & \textbf{0.9} \\ \midrule
\textbf{FPS }             & 191.4 & 226.7 & 236.3 & 255.0 & 281.4 & 301.3 & 313.8 \\ 
\textbf{PSNR}             & 25.744 & 23.932 & 23.859 & 23.692 & 23.573 & 23.530 & 23.974 \\
\bottomrule
\end{tabular}
}

\label{tab:fps-with-pruningstrength}
\end{table}

\begin{table}[!tb]
\centering
\setlength{\tabcolsep}{30pt}
\caption{\textbf{Inference latency comparison across methods ($\beta=0.9$).} Our approach achieves latency similar to the non finetuned baseline while being significantly faster than baselines that require per-scene finetuning of retained Gaussians. This demonstrates that \PaperTitle{} preserves fast feed-forward inference while supporting controllable pruning.}
\resizebox{\linewidth}{!}{
\begin{tabular}{@{}l|l|l@{}}
\toprule
\textbf{Method}                & \textbf{Time (s)} & \textbf{PSNR} \\ \midrule
EAGLES (With Finetuning)    & 7.03              & 16.243 \\
PUP3DGS (With Finetuning)   & 7.09              & 15.588  \\
LightGaussian (With Finetuning) & 5.64          & 16.056\\
EAGLES (Without Finetuning) & 1.39              & 12.774 \\
PUP3DGS (Without Finetuning) & 2.00             & 10.131 \\ 
LightGaussian (Without Finetuning) & 1.48       & 10.327 \\
\textbf{Ours}                           & \textbf{2.53}              & \textbf{20.740}              \\ \bottomrule
\end{tabular}
}

\label{tab:latency-withbaselines}
\end{table}

\begin{table*}
\begin{minipage}{0.49\linewidth}
\centering
    \includegraphics[width=\linewidth]{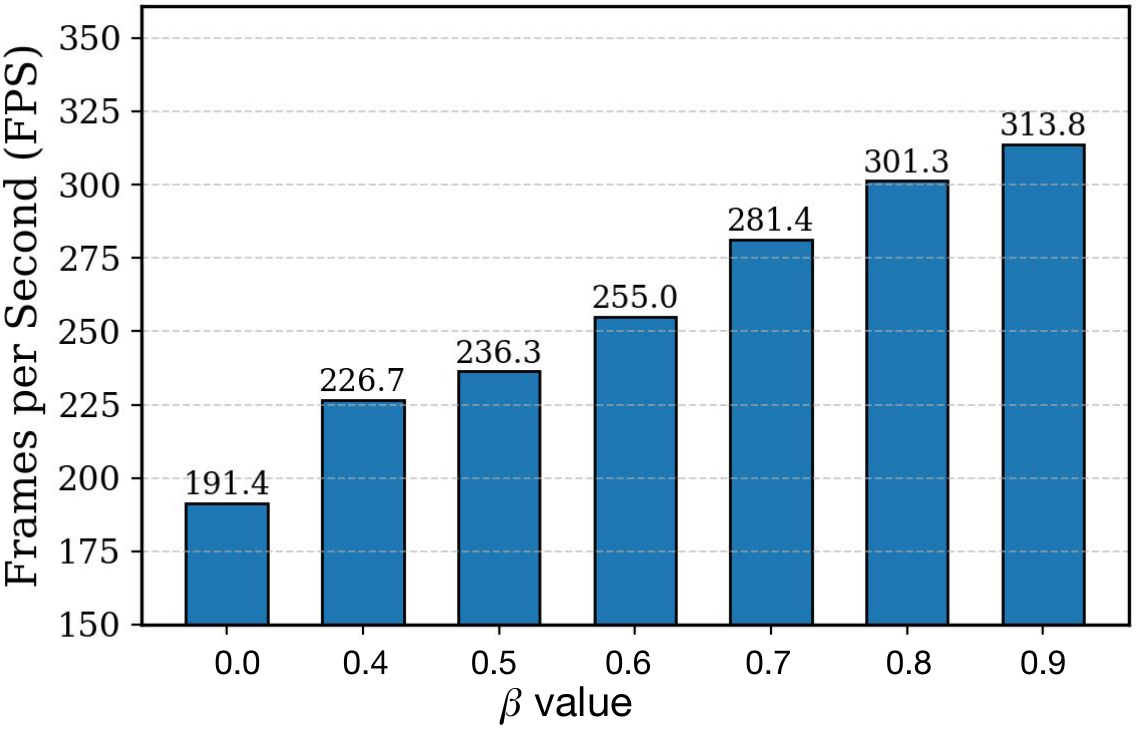}
    \captionof{figure}{Plot for FPS with increasing $\beta$ Strength}
    \label{fig:fps-prune-plot}
\end{minipage}
\hfill
\begin{minipage}{0.49\linewidth}
\centering
    \includegraphics[width=0.9\linewidth]{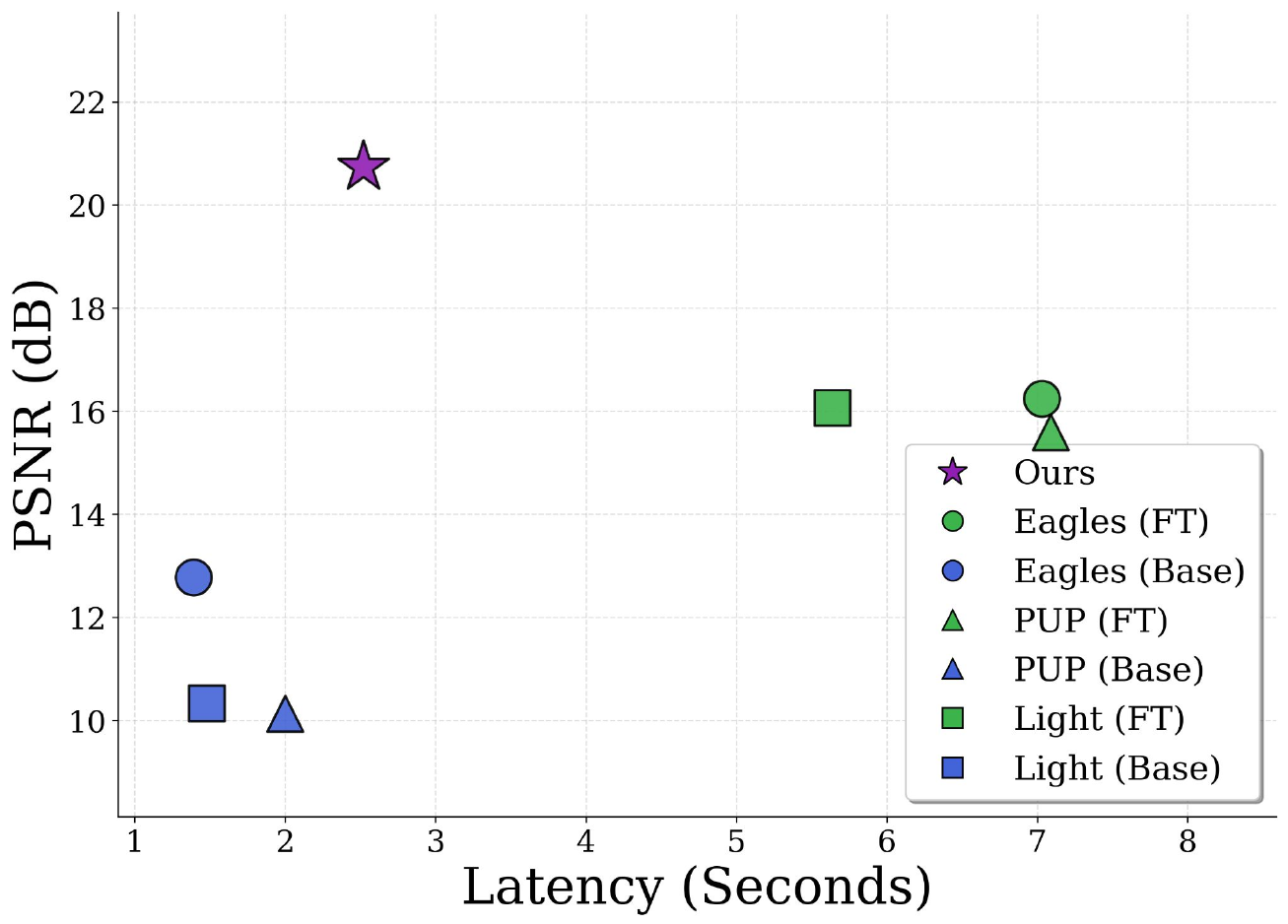}
    \captionof{figure}{Pareto Plot for inference time and quality comparison for RE10K at $\beta=0.8$}
    \label{fig:pareto}
\end{minipage}
\end{table*}

We measure rendering speed by synthesizing 1000 novel views for each pruning stage and reporting the Frames Per Second (FPS). We utilize
our pruning technique to prune the Gaussians. Tab. \ref{tab:fps-with-pruningstrength} summarizes the FPS achieved across pruning strengths, and Fig. \ref{fig:fps-prune-plot} visualizes this trend. As pruning strength increases, the reduced number of active Gaussians consistently improves rendering speed, demonstrating the effectiveness of pruning as a controllable knob for balancing reconstruction quality and runtime efficiency. \cref{fig:pareto} presents pareto plot to compare inference time and quality analysis of ours with other methods.

We report the end-to-end latency of predicting 3D Gaussians from sparse input views. Tab. ~\ref{tab:latency-withbaselines} summarizes the runtime across different baselines. Our method exhibits latency comparable to the non-finetuned baseline and is substantially faster than baselines that require per-scene finetuning of the remaining Gaussians. This highlights that \PaperTitle{} retains the feed-forward efficiency of MASt3R while enabling controllable pruning without the overhead of optimization. Since rendering is efficient, changing $\beta$ does not significantly affect the result in our method.

\section{Results with naive pruning baselines}
\label{sec:naive_pruning_results}

We include results with naive pruning baselines for completeness. We include results with pruning based on opacity as well as low resolution input to decrease number of 3D Gaussian primitives. Specifically we remove lowest opacity Gaussians till pruning budget $\beta$ is met; similarly, we downscale the image till the total number of pixels in multi view input images meets the $\beta$ budget. Tab. \ref{tab:naivepruningbaselines} presents these results, we observe that this naive pruning degrades the scene as the remaining Gaussians are not refined to adjust for the removed primitives.

\begin{table}[!tb]
\centering
\small
\setlength{\tabcolsep}{4pt}
\caption{Results with naive pruning baseline on AnySplat backbone.}
\resizebox{\linewidth}{!}{%
\begin{tabular}{@{}ccccccccccc@{}}
\toprule
\multicolumn{2}{c}{}                                                                                         & \multicolumn{3}{c}{$\beta=0.4$}                                      & \multicolumn{3}{c}{$\beta=0.6$}                                      & \multicolumn{3}{c}{$\beta=0.8$}                 \\ \midrule
\multicolumn{1}{c|}{}                                                  & \multicolumn{1}{c|}{\textbf{Views}} & \textbf{PSNR$\uparrow$}  & \textbf{LPIPS$\downarrow$} & \multicolumn{1}{c|}{\textbf{SSIM$\uparrow$}} & \textbf{PSNR$\uparrow$}  & \textbf{LPIPS$\downarrow$} & \multicolumn{1}{c|}{\textbf{SSIM$\uparrow$}} & \textbf{PSNR$\uparrow$}  & \textbf{LPIPS$\downarrow$} & \textbf{SSIM$\uparrow$} \\ \midrule
\multicolumn{1}{c|}{\multirow{4}{*}{\rotatebox[origin=c]{90}{Opacity}}} & \multicolumn{1}{c|}{6}              & 17.49          & \textbf{0.30}  & \multicolumn{1}{c|}{0.60}          & 13.73          & 0.38           & \multicolumn{1}{c|}{0.51}          & 9.67           & 0.51           & 0.40          \\
\multicolumn{1}{c|}{}                                                  & \multicolumn{1}{c|}{9}              & 17.77          & \textbf{0.32}  & \multicolumn{1}{c|}{0.56}          & 14.27          & 0.39           & \multicolumn{1}{c|}{0.49}          & 10.08          & 0.50           & 0.38          \\
\multicolumn{1}{c|}{}                                                  & \multicolumn{1}{c|}{16}             & 17.59          & \textbf{0.37}  & \multicolumn{1}{c|}{0.51}          & 14.72          & 0.41           & \multicolumn{1}{c|}{0.46}          & 10.41          & 0.51           & 0.35          \\
\multicolumn{1}{c|}{}                                                  & \multicolumn{1}{c|}{32}             & 16.17          & \textbf{0.44}  & \multicolumn{1}{c|}{0.44}          & 14.43          & 0.46           & \multicolumn{1}{c|}{0.41}          & 10.85          & 0.53           & 0.34          \\ \midrule
\multicolumn{1}{c|}{\multirow{4}{*}{\rotatebox[origin=c]{90}{Low Res}}} & \multicolumn{1}{c|}{6}              & 16.45          & 0.49           & \multicolumn{1}{c|}{0.49}          & 10.60          & 0.66           & \multicolumn{1}{c|}{0.24}          & 6.65           & 0.69           & 0.19          \\
\multicolumn{1}{c|}{}                                                  & \multicolumn{1}{c|}{9}              & 16.61          & 0.50           & \multicolumn{1}{c|}{0.47}          & 11.25          & 0.65           & \multicolumn{1}{c|}{0.25}          & 7.02           & 0.69           & 0.19          \\
\multicolumn{1}{c|}{}                                                  & \multicolumn{1}{c|}{16}             & 16.37          & 0.52           & \multicolumn{1}{c|}{0.43}          & 12.14          & 0.64           & \multicolumn{1}{c|}{0.27}          & 7.50           & 0.69           & 0.17          \\
\multicolumn{1}{c|}{}                                                  & \multicolumn{1}{c|}{32}             & 15.69          & 0.55           & \multicolumn{1}{c|}{0.39}          & 12.83          & 0.64           & \multicolumn{1}{c|}{0.29}          & 8.69           & 0.68           & 0.18          \\ \midrule
\multicolumn{1}{c|}{\multirow{4}{*}{\rotatebox[origin=c]{90}{Ours}}}    & \multicolumn{1}{c|}{6}              & \textbf{20.45} & 0.33           & \multicolumn{1}{c|}{\textbf{0.60}} & \textbf{20.31} & \textbf{0.37}  & \multicolumn{1}{c|}{\textbf{0.58}} & \textbf{20.05} & \textbf{0.41}  & \textbf{0.56} \\
\multicolumn{1}{c|}{}                                                  & \multicolumn{1}{c|}{9}              & \textbf{19.68} & 0.35           & \multicolumn{1}{c|}{\textbf{0.57}} & \textbf{19.58} & \textbf{0.38}  & \multicolumn{1}{c|}{\textbf{0.55}} & \textbf{19.44} & \textbf{0.41}  & \textbf{0.54} \\
\multicolumn{1}{c|}{}                                                  & \multicolumn{1}{c|}{16}             & \textbf{19.13} & 0.39           & \multicolumn{1}{c|}{\textbf{0.52}} & \textbf{19.08} & \textbf{0.41}  & \multicolumn{1}{c|}{\textbf{0.51}} & \textbf{18.99} & \textbf{0.44}  & \textbf{0.50} \\
\multicolumn{1}{c|}{}                                                  & \multicolumn{1}{c|}{32}             & \textbf{17.68} & 0.45           & \multicolumn{1}{c|}{\textbf{0.48}} & \textbf{17.65} & \textbf{0.46}  & \multicolumn{1}{c|}{\textbf{0.47}} & \textbf{17.61} & \textbf{0.48}  & \textbf{0.46} \\ \bottomrule
\end{tabular}
}
\label{tab:naivepruningbaselines}
\end{table}

\begin{table}[!tb]
\centering
\setlength{\tabcolsep}{4pt}
\caption{Results on ACID dataset with various existing pruning methods without finetuning the remaining Gaussians using NoPoSplat and HiSplat as backbones.}
\resizebox{\linewidth}{!}{
\begin{tabular}{@{}clccccccccc@{}}
\toprule
\multicolumn{1}{l}{} & \multicolumn{10}{c}{\textbf{ACID Dataset}} \\ \midrule
\multicolumn{2}{c|}{\multirow{2}{*}{\begin{tabular}[c]{@{}c@{}}\textbf{Pruning Method}\\ \textbf{(Without Finetuning)}\end{tabular}}} &
\multicolumn{3}{c|}{\textbf{$\beta=0.4$}} &
\multicolumn{3}{c|}{\textbf{$\beta=0.6$}} &
\multicolumn{3}{c}{\textbf{$\beta=0.8$}} \\ \cmidrule(l){3-11}
\multicolumn{2}{c|}{} &
\multicolumn{1}{c}{\textbf{PSNR$\uparrow$}} &
\multicolumn{1}{c}{\textbf{LPIPS$\downarrow$}} &
\multicolumn{1}{c|}{\textbf{SSIM$\uparrow$}} &
\multicolumn{1}{c}{\textbf{PSNR$\uparrow$}} &
\multicolumn{1}{c}{\textbf{LPIPS$\downarrow$}} &
\multicolumn{1}{c|}{\textbf{SSIM$\uparrow$}} &
\multicolumn{1}{c}{\textbf{PSNR$\uparrow$}} &
\multicolumn{1}{c}{\textbf{LPIPS$\downarrow$}} &
\multicolumn{1}{c}{\textbf{SSIM$\uparrow$} }\\ \midrule
\multirow{4}{*}{\rotatebox[origin=c]{90}{NoPoSplat}}                        & \multicolumn{1}{l|}{EAGLES}                                  & 15.028                                      & 0.511                                          & \multicolumn{1}{c|}{0.415}          & 11.883                                      & 0.605                                          & \multicolumn{1}{c|}{0.264}          & 8.977                                       & 0.683                                          & 0.131                              \\
                                                     & \multicolumn{1}{l|}{LightGaussian}                           & 19.670                                      & 0.363                                          & \multicolumn{1}{c|}{0.578}          & 15.502                                      & 0.484                                          & \multicolumn{1}{c|}{0.433}          & 11.228                                      & 0.622                                          & 0.233                              \\
                                                     & \multicolumn{1}{l|}{PUP-3DGS}                                & 11.845                                      & 0.424                                          & \multicolumn{1}{c|}{0.478}          & 9.257                                       & 0.514                                          & \multicolumn{1}{c|}{0.332}          & 7.516                                       & 0.617                                          & 0.171                              \\
                                                     & \multicolumn{1}{l|}{EfficientGS}                             & 15.189                                      & 0.446                                          & \multicolumn{1}{c|}{0.468}          & 12.316                                      & 0.524                                          & \multicolumn{1}{c|}{0.376}          & 8.886                                       & 0.653                                          & 0.196                              \\ \midrule
\multirow{4}{*}{\rotatebox[origin=c]{90}{HiSplat}}                             & \multicolumn{1}{l|}{EAGLES}                                  & 14.921                                      & 0.538                                          & \multicolumn{1}{c|}{0.452}          & 11.001                                      & 0.636                                          & \multicolumn{1}{c|}{0.272}          & 10.999                                      & 0.636                                          & 0.272                              \\
                                                     & \multicolumn{1}{l|}{LightGaussian}                           & 21.986                                      & 0.32                                           & \multicolumn{1}{c|}{0.705}          & 21.986                                      & 0.32                                           & \multicolumn{1}{c|}{0.705}          & 17.095                                      & 0.455                                          & 0.549                              \\
                                                     & \multicolumn{1}{l|}{PUP-3DGS}                                & 10.221                                      & 0.492                                          & \multicolumn{1}{c|}{0.438}          & 8.133                                       & 0.614                                          & \multicolumn{1}{c|}{0.245}          & 8.132                                       & 0.614                                          & 0.245                              \\
                                                     & \multicolumn{1}{l|}{EfficientGS}                             & 14.333                                      & 0.416                                          & \multicolumn{1}{c|}{0.528}          & 10.866                                      & 0.571                                          & \multicolumn{1}{c|}{0.32}           & 8.394                                       & 0.682                                          & 0.158                              \\ \midrule
\multicolumn{1}{l}{}                                 & \multicolumn{1}{l|}{\textbf{Ours}}                                    & \textbf{22.294}                             & \textbf{0.2991}                                & \multicolumn{1}{c|}{\textbf{0.640}} & \textbf{22.110}                             & \textbf{0.310}                                 & \multicolumn{1}{c|}{\textbf{0.627}} & \textbf{20.740}                             & \textbf{0.342}                                 & \textbf{0.593}                     \\ \bottomrule
\end{tabular}
}
\label{tab:results_acid_comparison}
\end{table}
\begin{table}[!tb]
\centering
\setlength{\tabcolsep}{6pt}
\caption{\textbf{Results on RE10K dataset} \textbf{without finetuning the remaining Gaussians.} We see baselines performance degrade with increase in pruning strength. In extreme pruning scenario, our method outperforms the baselines by a large margin.}
\resizebox{\linewidth}{!}{
\begin{tabular}{@{}cl|ccc|ccc|ccc@{}}
\toprule
\multicolumn{2}{c|}{} & \multicolumn{3}{c|}{\textbf{$\beta=0.4$}} & \multicolumn{3}{c|}{\textbf{$\beta=0.6$}} & \multicolumn{3}{c}{\textbf{$\beta=0.8$}} \\ \cmidrule(l){3-11}
\multicolumn{2}{c|}{\multirow{-2}{*}{\textbf{Pruning Method}}} & \textbf{PSNR$\uparrow$} & \textbf{LPIPS$\downarrow$} & \textbf{SSIM$\uparrow$} & \textbf{PSNR$\uparrow$} & \textbf{LPIPS$\downarrow$} & \textbf{SSIM$\uparrow$} & \textbf{PSNR$\uparrow$} & \textbf{LPIPS$\downarrow$} & \textbf{SSIM$\uparrow$} \\ \midrule
\multicolumn{1}{c|}{}                            & EAGLES        & 13.769          & 0.483             & 0.461          & 10.908          & 0.588             & 0.289          & 8.117                         & 0.669                        & 0.139                        \\
\multicolumn{1}{c|}{}                            & LightGaussian & 18.218          & 0.357             & 0.621          & 13.642          & 0.486             & 0.444          & 8.435                         & 0.621                        & 0.206                        \\
\multicolumn{1}{c|}{}                            & PUP-3DGS      & 11.649          & 0.396             & 0.510          & 8.901           & 0.490             & 0.352          & 6.743                         & 0.589                        & 0.174                        \\
\multicolumn{1}{c|}{\multirow{-4}{*}{\rotatebox[origin=c]{90}{NoPoSplat}}} & EfficientGS   & 13.755          & 0.407             & 0.520          & 10.686          & 0.513             & 0.384          & 7.842                         & 0.631                        & 0.209                        \\ \midrule
\multicolumn{1}{c|}{}                            & EAGLES        & 16.258          & 0.415             & 0.580          & 16.265          & 0.415             & 0.580          & {12.774} & {0.516} & {0.410} \\
\multicolumn{1}{c|}{}                            & LightGaussian & 14.425          & 0.461             & 0.504          & 14.427          & 0.461             & 0.504          & {10.327} & {0.589} & {0.297} \\
\multicolumn{1}{c|}{}                            & PUP-3DGS      & 13.577          & 0.355             & 0.629          & 13.576          & 0.355             & 0.629          & {10.131} & {0.465} & {0.460} \\
\multicolumn{1}{c|}{\multirow{-4}{*}{\rotatebox[origin=c]{90}{HiSplat}}}   & EfficientGS   & 13.635          & 0.359             & 0.576          & 10.292          & 0.485             & 0.397          & 7.890                         & 0.606                        & 0.226                        \\ \midrule
\multicolumn{1}{l}{}                             & Ours          & \textbf{22.294} & \textbf{0.235}    & \textbf{0.735} & \textbf{22.110} & \textbf{0.243}    & \textbf{0.726} & \textbf{20.740}               & \textbf{0.272}               & \textbf{0.692}               \\ \bottomrule
\end{tabular}
}

\label{tab:results_re10k_comparison_nopo_hisplat}
\end{table}

\begin{table}[!tb]
\centering
\setlength{\tabcolsep}{4pt}
\caption{Quantitative comparison on the RE10K dataset (No Finetune). We evaluate metrics across different pruning targets.}
\resizebox{\linewidth}{!}{
\begin{tabular}{cl|ccc|ccc|ccc|ccc|ccc|ccc}
\toprule
& & \multicolumn{3}{c|}{\textbf{$\beta = 0.4$}} & \multicolumn{3}{c|}{\textbf{$\beta = 0.5$}} & \multicolumn{3}{c|}{\textbf{$\beta = 0.6$}} & \multicolumn{3}{c|}{\textbf{$\beta = 0.7$}} & \multicolumn{3}{c|}{\textbf{$\beta = 0.8$}} & \multicolumn{3}{c}{\textbf{$\beta = 0.9$}} \\
\textbf{Base} & \textbf{Method} & \textbf{PSNR$\uparrow$} & \textbf{LPIPS$\downarrow$} & \textbf{SSIM$\uparrow$} & \textbf{PSNR$\uparrow$} & \textbf{LPIPS$\downarrow$} & \textbf{SSIM$\uparrow$} & \textbf{PSNR$\uparrow$} & \textbf{LPIPS$\downarrow$} & \textbf{SSIM$\uparrow$} & \textbf{PSNR$\uparrow$} & \textbf{LPIPS$\downarrow$} & \textbf{SSIM$\uparrow$} & \textbf{PSNR$\uparrow$} & \textbf{LPIPS$\downarrow$} & \textbf{SSIM$\uparrow$} & \textbf{PSNR$\uparrow$} & \textbf{LPIPS$\downarrow$} & \textbf{SSIM$\uparrow$} \\
\midrule
\multirow{4}{*}{\rotatebox[origin=c]{90}{pixelSplat}} & Random & 14.67 & 0.480 & 0.530 & 13.22 & 0.520 & 0.470 & 11.70 & 0.550 & 0.400 & 10.10 & 0.590 & 0.320 & 8.43 & 0.640 & 0.240 & 6.70 & 0.690 & 0.140 \\
& LightGaus. & 21.32 & 0.350 & 0.720 & 20.58 & \underline{0.380} & 0.690 & 19.83 & 0.410 & 0.670 & \underline{18.68} & 0.450 & \underline{0.640} & \underline{16.23} & 0.520 & \underline{0.570} & \underline{12.79} & 0.600 & \underline{0.460} \\
& EAGLES & \textbf{24.75} & \textbf{0.190} & \textbf{0.840} & \textbf{23.14} & \textbf{0.240} & \textbf{0.810} & \underline{20.52} & \underline{0.310} & \textbf{0.750} & 17.72 & \underline{0.400} & 0.620 & 13.58 & \underline{0.490} & 0.470 & 9.79 & \underline{0.580} & 0.290 \\
& PUP-3DGS & 16.21 & 0.380 & 0.620 & 15.41 & 0.390 & 0.610 & 14.16 & 0.420 & 0.590 & 12.68 & 0.450 & 0.540 & 11.01 & 0.500 & 0.470 & 9.03 & \underline{0.580} & 0.340 \\ 
\midrule
\multirow{4}{*}{\rotatebox[origin=c]{90}{MVSplat}} & Random & 19.27 & 0.440 & 0.570 & 16.99 & 0.500 & 0.470 & 14.65 & 0.560 & 0.380 & 12.28 & 0.610 & 0.290 & 9.88 & 0.660 & 0.200 & 7.43 & 0.710 & 0.100 \\
& LightGaus. & 7.71 & 0.690 & 0.130 & 7.33 & 0.700 & 0.110 & 7.02 & 0.710 & 0.100 & 6.69 & 0.720 & 0.090 & 6.19 & 0.730 & 0.070 & 5.77 & 0.740 & 0.050 \\
& EAGLES & 7.52 & 0.720 & 0.070 & 11.94 & 0.550 & 0.370 & 10.51 & 0.600 & 0.280 & 9.14 & 0.640 & 0.200 & 7.84 & 0.690 & 0.130 & 6.57 & 0.720 & 0.070 \\
& PUP-3DGS & 11.51 & 0.370 & 0.590 & 10.23 & 0.420 & 0.510 & 9.13 & 0.470 & 0.430 & 8.11 & 0.520 & 0.340 & 7.14 & 0.570 & 0.250 & 6.17 & 0.620 & 0.140 \\ 
\midrule
\multicolumn{2}{l|}{\textbf{Ours}} & \underline{22.29} & \underline{0.230} & \underline{0.740} & \underline{22.24} & \textbf{0.240} & \underline{0.730} & \textbf{22.11} & \textbf{0.240} & \underline{0.730} & \textbf{21.77} & \textbf{0.250} & \textbf{0.720} & \textbf{20.74} & \textbf{0.270} & \textbf{0.690} & \textbf{17.85} & \textbf{0.310} & \textbf{0.650} \\
\bottomrule
\end{tabular}%
}

\label{tab:results_re10k_comparison_pixel_mv_nopo_random_wofinetune}
\end{table}

\begin{table}[!tb]
\centering
\setlength{\tabcolsep}{4pt}
\caption{Detailed  Results on the ACID dataset (No Finetune) comparing different backbone methods across varying numbers of 3D Gaussians.}
\resizebox{\linewidth}{!}{
\begin{tabular}{ll|ccc|ccc|ccc|ccc|ccc|ccc}
\toprule
& & \multicolumn{3}{c|}{$\beta = 0.4$} & \multicolumn{3}{c|}{$\beta = 0.5$} & \multicolumn{3}{c|}{$\beta = 0.6$} & \multicolumn{3}{c|}{$\beta = 0.7$} & \multicolumn{3}{c|}{$\beta = 0.8$} & \multicolumn{3}{c}{$\beta = 0.9$} \\
\textbf{Base} & \textbf{Method} & \textbf{PSNR$\uparrow$} & \textbf{LPIPS$\downarrow$} & \textbf{SSIM$\uparrow$} & \textbf{PSNR$\uparrow$} & \textbf{LPIPS$\downarrow$} & \textbf{SSIM$\uparrow$} & \textbf{PSNR$\uparrow$} & \textbf{LPIPS$\downarrow$} & \textbf{SSIM$\uparrow$} & \textbf{PSNR$\uparrow$} & \textbf{LPIPS$\downarrow$} & \textbf{SSIM$\uparrow$} & \textbf{PSNR$\uparrow$} & \textbf{LPIPS$\downarrow$} & \textbf{SSIM$\uparrow$} & \textbf{PSNR$\uparrow$} & \textbf{LPIPS$\downarrow$} & \textbf{SSIM$\uparrow$} \\
\midrule
\multirow{4}{*}{\rotatebox[origin=c]{90}{pixelSplat}} & Random & 15.47 & 0.501 & 0.492 & 14.04 & 0.533 & 0.431 & 12.54 & 0.569 & 0.364 & 10.99 & 0.612 & 0.293 & 9.37 & 0.662 & 0.211 & 7.69 & 0.711 & 0.122 \\
& LightGaus. & 23.27 & 0.382 & 0.691 & 22.76 & 0.405 & 0.663 & 22.34 & 0.428 & 0.635 & 21.76 & 0.453 & 0.618 & 20.21 & 0.499 & 0.582 & 15.95 & 0.576 & 0.494 \\
& EAGLES & \textbf{25.14} & \textbf{0.252} & \textbf{0.794} & 22.58 & 0.305 & \textbf{0.742} & 19.63 & 0.381 & 0.669 & 16.66 & 0.462 & 0.561 & 13.58 & 0.551 & 0.434 & 10.35 & 0.642 & 0.271 \\
& PUP-3DGS & 19.65 & 0.341 & 0.658 & 17.96 & 0.372 & 0.639 & 13.78 & 0.451 & 0.528 & 11.65 & 0.514 & 0.449 & 11.65 & 0.512 & 0.449 & 9.55 & 0.589 & 0.322 \\ 
\midrule
\multirow{4}{*}{\rotatebox[origin=c]{90}{MVSplat}} & Random & 18.72 & 0.483 & 0.489 & 16.46 & 0.531 & 0.412 & 14.28 & 0.579 & 0.334 & 12.15 & 0.624 & 0.251 & 10.07 & 0.672 & 0.169 & 8.02 & 0.714 & 0.091 \\
& LightGaus. & 8.09 & 0.721 & 0.089 & 7.78 & 0.728 & 0.076 & 7.54 & 0.735 & 0.068 & 7.28 & 0.742 & 0.061 & 6.88 & 0.756 & 0.044 & 6.53 & 0.762 & 0.032 \\
& EAGLES & 14.86 & 0.509 & 0.462 & 13.12 & 0.556 & 0.378 & 11.56 & 0.598 & 0.294 & 10.13 & 0.641 & 0.212 & 8.79 & 0.682 & 0.141 & 7.52 & 0.721 & 0.074 \\
& PUP-3DGS & 11.43 & 0.394 & 0.561 & 10.26 & 0.442 & 0.489 & 9.32 & 0.491 & 0.412 & 8.53 & 0.542 & 0.334 & 7.80 & 0.591 & 0.241 & 7.05 & 0.652 & 0.141 \\ 
\midrule
\multicolumn{2}{l|}{\textbf{Ours}} & 22.55 & 0.301 & 0.642 & \textbf{22.47} & \textbf{0.304} & 0.639 & \textbf{22.31} & \textbf{0.312} & \textbf{0.631} & \textbf{21.96} & \textbf{0.321} & \textbf{0.619} & \textbf{21.06} & \textbf{0.342} & \textbf{0.592} & \textbf{18.48} & \textbf{0.381} & \textbf{0.552} \\
\bottomrule
\end{tabular}%
}

\label{tab:results_acid_comparison_pixel_mv_nopo_random_wofinetune}
\end{table}
\section{Results on different backbones}
\label{sec:results-on-different-backbone}
\begin{figure}[htbp]
    \centering
    \includegraphics[width=0.6\linewidth]{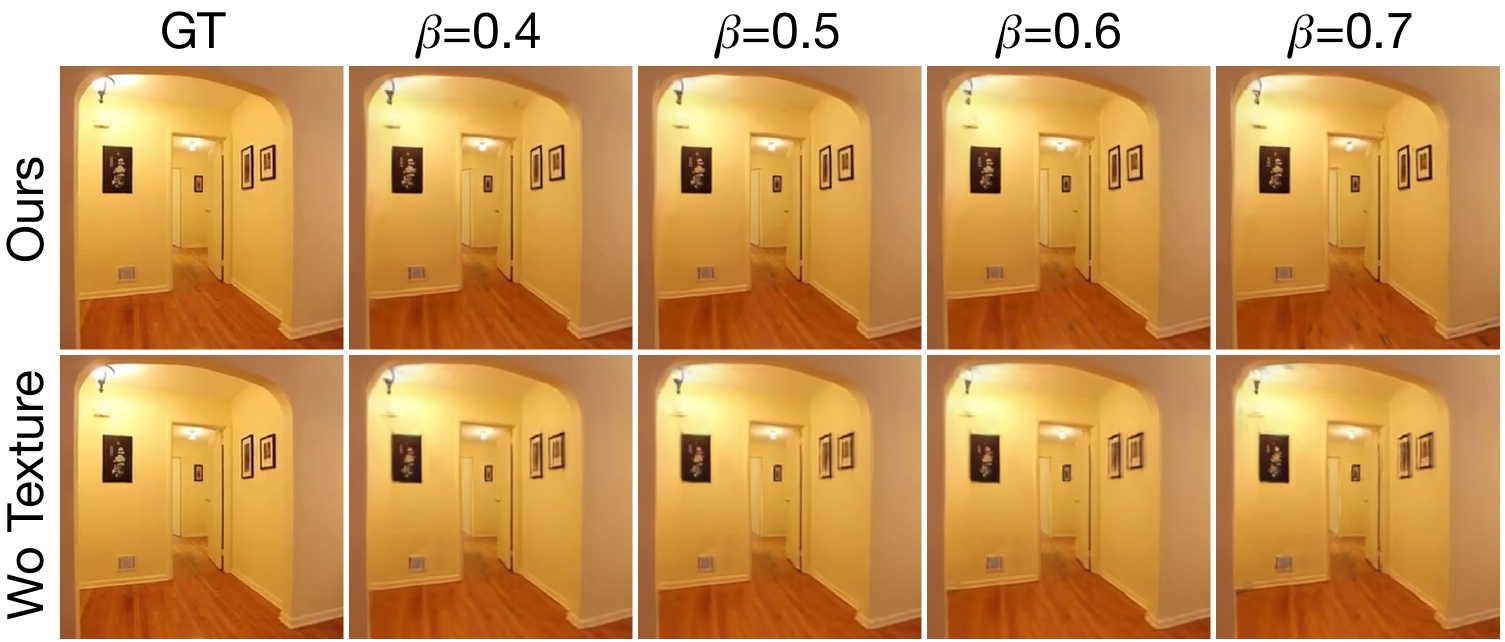}
    \caption{Results for RE10K and ACID dataset with $\beta=0.4$ and $\beta=0.8$ respectively.}
    \label{fig:scene_fig4_combined_both}
\end{figure}

Here we present detailed results of our experiments on RE10K and ACID dataset for different pruning stages on NoPoSplat and HiSplat backbone. \cref{tab:results_acid_comparison}, \cref{tab:results_re10k_comparison_nopo_hisplat} presents results without finetuning on both ACID and RE10K, respectively, and \cref{fig:scene_fig4_combined_both} demonstrates results for ACID and RE10K for both with and without finetuning. We also include results with different feed-forward backbones with various pruning strategies. Specifically we use pixelSplat and MVSplat backbones on RE10K and ACID dataset with all pruning strategies discussed before (we've also added random 3D Gaussian pruning) along with before and after finetuning results. \cref{tab:results_re10k_comparison_pixel_mv_nopo_random_wofinetune} on RE10K dataset and \cref{tab:results_acid_comparison_pixel_mv_nopo_random_wofinetune} on ACID dataset presents results of all pruning stages for 2 different feed-forward backbone on 4 different pruning strategies where we don't fine tune the remaining 3D Gaussians after pruning. The table demonstrates that our method achieves better results across most of the pruning stages consistently than any other baselines. Similarly \cref{tab:results_acid_comparison_pixel_mv_nopo_random_withfinetuning} for ACID dataset and \cref{tab:results_re10k_comparison_pixel_mv_nopo_random_withfinetuning} for RE10K dataset contains results for different feed-forward backbones where we finetune the remaining 3D Gaussians.

\begin{table}[!tb]
\centering
\setlength{\tabcolsep}{2pt}
\caption{Detailed Results on the ACID dataset with finetuning after pruning comparing different backbone methods across varying pruning techniques.}
\resizebox{\linewidth}{!}{
\begin{tabular}{l|ccc|ccc|ccc|ccc|ccc|ccc}
\toprule
\textbf{Method} & \multicolumn{3}{c|}{$\beta = 0.4$} & \multicolumn{3}{c|}{$\beta = 0.5$} & \multicolumn{3}{c|}{$\beta = 0.6$} & \multicolumn{3}{c|}{$\beta = 0.7$} & \multicolumn{3}{c|}{$\beta = 0.8$} & \multicolumn{3}{c}{$\beta = 0.9$} \\
& \textbf{PSNR$\uparrow$} & \textbf{LPIPS$\downarrow$} & \textbf{SSIM$\uparrow$} & \textbf{PSNR$\uparrow$} & \textbf{LPIPS$\downarrow$} & \textbf{SSIM$\uparrow$} & \textbf{PSNR$\uparrow$} & \textbf{LPIPS$\downarrow$} & \textbf{SSIM$\uparrow$} & \textbf{PSNR$\uparrow$} & \textbf{LPIPS$\downarrow$} & \textbf{SSIM$\uparrow$} & \textbf{PSNR$\uparrow$} & \textbf{LPIPS$\downarrow$} & \textbf{SSIM$\uparrow$} & \textbf{PSNR$\uparrow$} & \textbf{LPIPS$\downarrow$} & \textbf{SSIM$\uparrow$} \\
\midrule
pixelSplat + EAGLES   & 12.61 & 0.688 & 0.069 & 12.46 & 0.665 & 0.108 & 12.99 & 0.637 & 0.166 & 13.57 & 0.618 & 0.214 & 13.99 & 0.607 & 0.247 & 13.83 & 0.615 & 0.246 \\
pixelSplat + PUP-3DGS & 14.31 & 0.610 & 0.282 & 14.31 & 0.607 & 0.287 & 14.15 & 0.607 & 0.290 & 13.62 & 0.612 & 0.286 & 12.48 & 0.623 & 0.271 & 10.67 & 0.646 & 0.234 \\
\midrule
MVSplat + LightGaus.  & 11.99 & 0.688 & 0.136 & 11.41 & 0.695 & 0.131 & 10.80 & 0.703 & 0.119 & 9.98  & 0.717 & 0.101 & 9.28  & 0.729 & 0.085 & 8.38  & 0.744 & 0.065 \\
MVSplat + EAGLES      & 12.50 & 0.677 & 0.122 & 12.42 & 0.677 & 0.135 & 12.15 & 0.679 & 0.144 & 11.60 & 0.687 & 0.142 & 10.62 & 0.705 & 0.118 & 9.03  & 0.739 & 0.069 \\
MVSplat + PUP-3DGS    & 9.26  & 0.680 & 0.046 & 8.73  & 0.683 & 0.041 & 8.27  & 0.687 & 0.037 & 7.89  & 0.692 & 0.036 & 7.52  & 0.697 & 0.037 & 7.06  & 0.706 & 0.035 \\
\midrule
\textbf{Ours} & 22.55 & 0.301 & 0.642 & \textbf{22.47} & \textbf{0.304} & 0.639 & \textbf{22.31} & \textbf{0.312} & \textbf{0.631} & \textbf{21.96} & \textbf{0.321} & \textbf{0.619} & \textbf{21.06} & \textbf{0.342} & \textbf{0.592} & \textbf{18.48} & \textbf{0.381} & \textbf{0.552} \\
\bottomrule
\end{tabular}%
}

\label{tab:results_acid_comparison_pixel_mv_nopo_random_withfinetuning}
\end{table}

\begin{table}[!tb]
\centering
\setlength{\tabcolsep}{4pt}
\caption{Detailed Results on the RE10K dataset (With Finetuning) after pruning comparing different backbone methods across varying pruning techniques.}
\resizebox{\linewidth}{!}{
\begin{tabular}{l|ccc|ccc|ccc|ccc|ccc|ccc}
\toprule
\textbf{Method} &
\multicolumn{3}{c|}{$\beta = 0.4$} &
\multicolumn{3}{c|}{$\beta = 0.5$} &
\multicolumn{3}{c|}{$\beta = 0.6$} &
\multicolumn{3}{c|}{$\beta = 0.7$} &
\multicolumn{3}{c|}{$\beta = 0.8$} &
\multicolumn{3}{c}{$\beta = 0.9$} \\
& \textbf{PSNR$\uparrow$} & \textbf{LPIPS$\downarrow$} & \textbf{SSIM$\uparrow$}
& \textbf{PSNR$\uparrow$} & \textbf{LPIPS$\downarrow$} & \textbf{SSIM$\uparrow$}
& \textbf{PSNR$\uparrow$} & \textbf{LPIPS$\downarrow$} & \textbf{SSIM$\uparrow$}
& \textbf{PSNR$\uparrow$} & \textbf{LPIPS$\downarrow$} & \textbf{SSIM$\uparrow$}
& \textbf{PSNR$\uparrow$} & \textbf{LPIPS$\downarrow$} & \textbf{SSIM$\uparrow$}
& \textbf{PSNR$\uparrow$} & \textbf{LPIPS$\downarrow$} & \textbf{SSIM$\uparrow$} \\
\midrule
pixelSplat + EAGLES   & 12.16 & 0.703 & 0.104 & 11.91 & 0.694 & 0.104 & 11.85 & 0.670 & 0.131 & 12.40 & 0.634 & 0.188 & 13.03 & 0.612 & 0.240 & 13.21 & 0.605 & 0.265 \\
pixelSplat + PUP-3DGS & 14.31 & 0.610 & 0.282 & 14.31 & 0.607 & 0.287 & 14.15 & 0.607 & 0.290 & 13.62 & 0.612 & 0.286 & 12.48 & 0.623 & 0.271 & 10.67 & 0.646 & 0.234 \\
\midrule
MVSplat + LightGaus.  & 10.94 & 0.702 & 0.113 & 10.80 & 0.701 & 0.123 & 10.40 & 0.702 & 0.126 & 9.73  & 0.706 & 0.119 & 9.05  & 0.710 & 0.109 & 8.04  & 0.720 & 0.090 \\
MVSplat + EAGLES      & 11.44 & 0.686 & 0.116 & 11.41 & 0.684 & 0.130 & 11.20 & 0.683 & 0.140 & 10.71 & 0.689 & 0.141 & 9.79  & 0.706 & 0.122 & 8.25  & 0.737 & 0.075 \\
MVSplat + PUP-3DGS    & 9.20  & 0.683 & 0.055 & 8.73  & 0.683 & 0.054 & 8.25  & 0.683 & 0.053 & 7.72  & 0.684 & 0.053 & 7.12  & 0.685 & 0.052 & 6.37  & 0.688 & 0.046 \\
\midrule
\textbf{Ours}         & \textbf{22.29} & \textbf{0.230} & \textbf{0.740} & \textbf{22.24} & \textbf{0.240} & \textbf{0.730} & \textbf{22.11} & \textbf{0.240} & \textbf{0.730} & \textbf{21.77} & \textbf{0.250} & \textbf{0.720} & \textbf{20.74} & \textbf{0.270} & \textbf{0.690} & \textbf{17.85} & \textbf{0.310} & \textbf{0.650} \\
\bottomrule
\end{tabular}%
}

\label{tab:results_re10k_comparison_pixel_mv_nopo_random_withfinetuning}
\end{table}

\begin{figure}[htbp]
    \centering
    \includegraphics[width=0.7\linewidth]{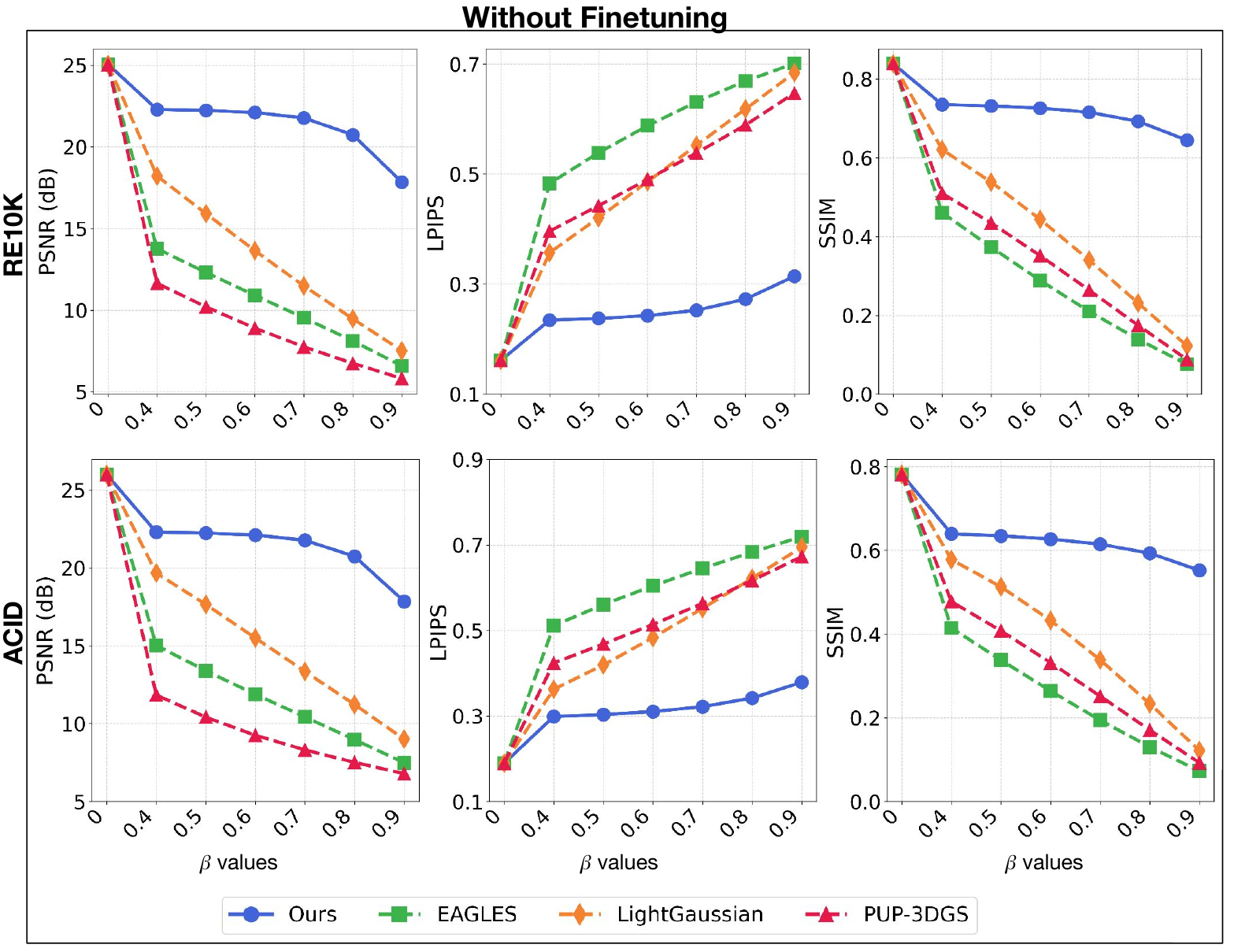}
    \caption{Plot of RE10K and ACID dataset where we compare our results with the relevant baselines without finetuning the remaining 3D Gaussians}
    \label{fig:plots_comparison_baselines_wofinetune}
\end{figure}

\begin{figure}[htbp]
    \centering
    \includegraphics[width=0.7\linewidth]{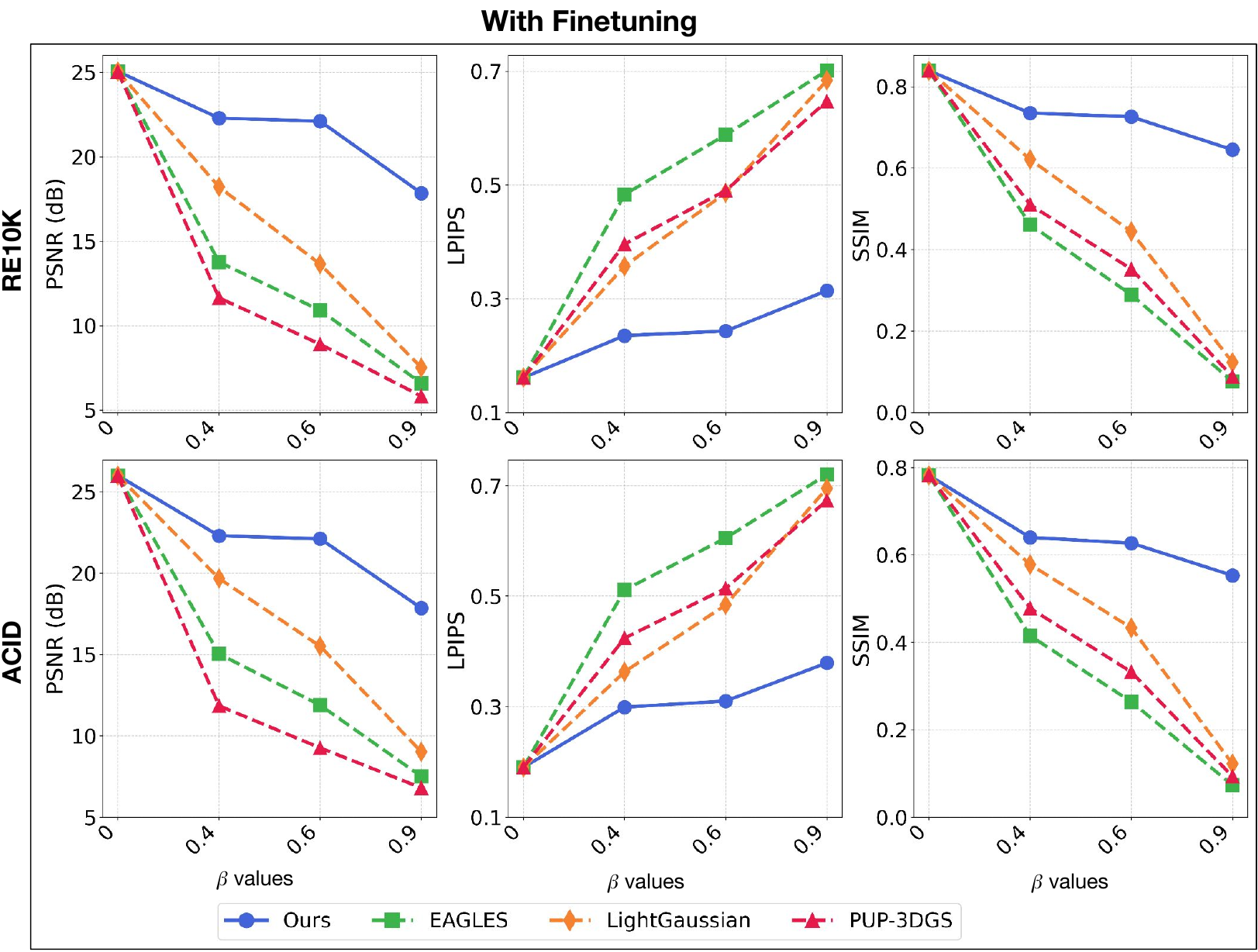}
    \caption{Plot of RE10K and ACID dataset where we compare our results with the relevant baselines with finetuning the remaining 3D Gaussians}
    \label{fig:plots_comparison_baselines_withfinetune}
\end{figure}

We've also included a detailed comparison of each scene with all baselines for a clearer understanding of our method's performance. \cref{fig:plots_comparison_baselines_wofinetune} ,\cref{fig:plots_comparison_baselines_withfinetune} shows the trends of PSNR, LPIPS and SSIM over different pruning percentages with and without finetuning on RE10K and ACID datasets with and without finetuning 3D Gaussians respectively with NoPoSplat backbone. 

\begin{figure}[htbp]
    \centering
    \includegraphics[width=0.8\textwidth]{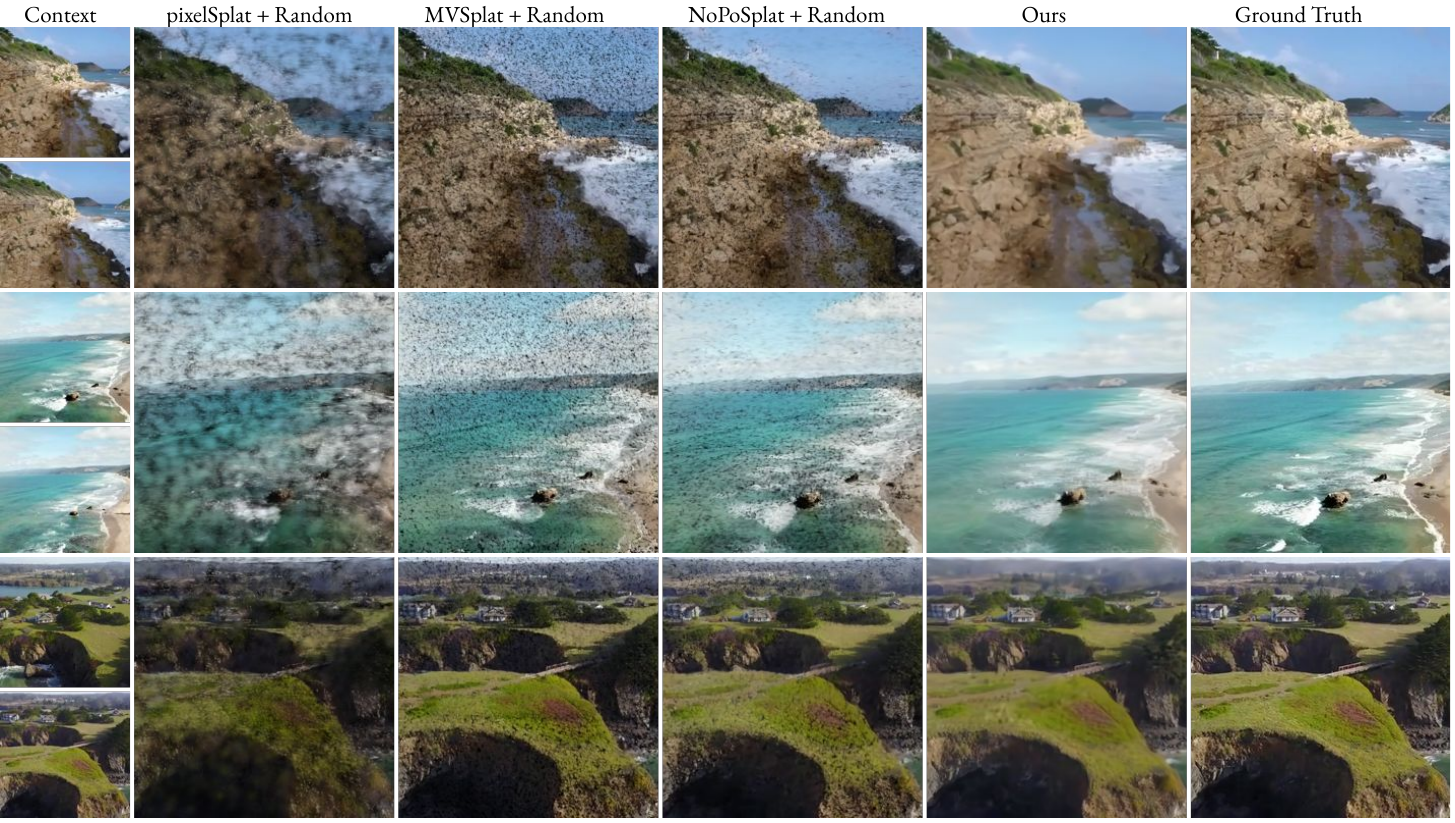}
    \caption{Detailed Results comparison with $\beta=0.4$ on ACID dataset}
    \label{fig:acid_78k_splats}
\end{figure}

\begin{figure}[htbp]
    \centering
    \includegraphics[width=0.8\textwidth]{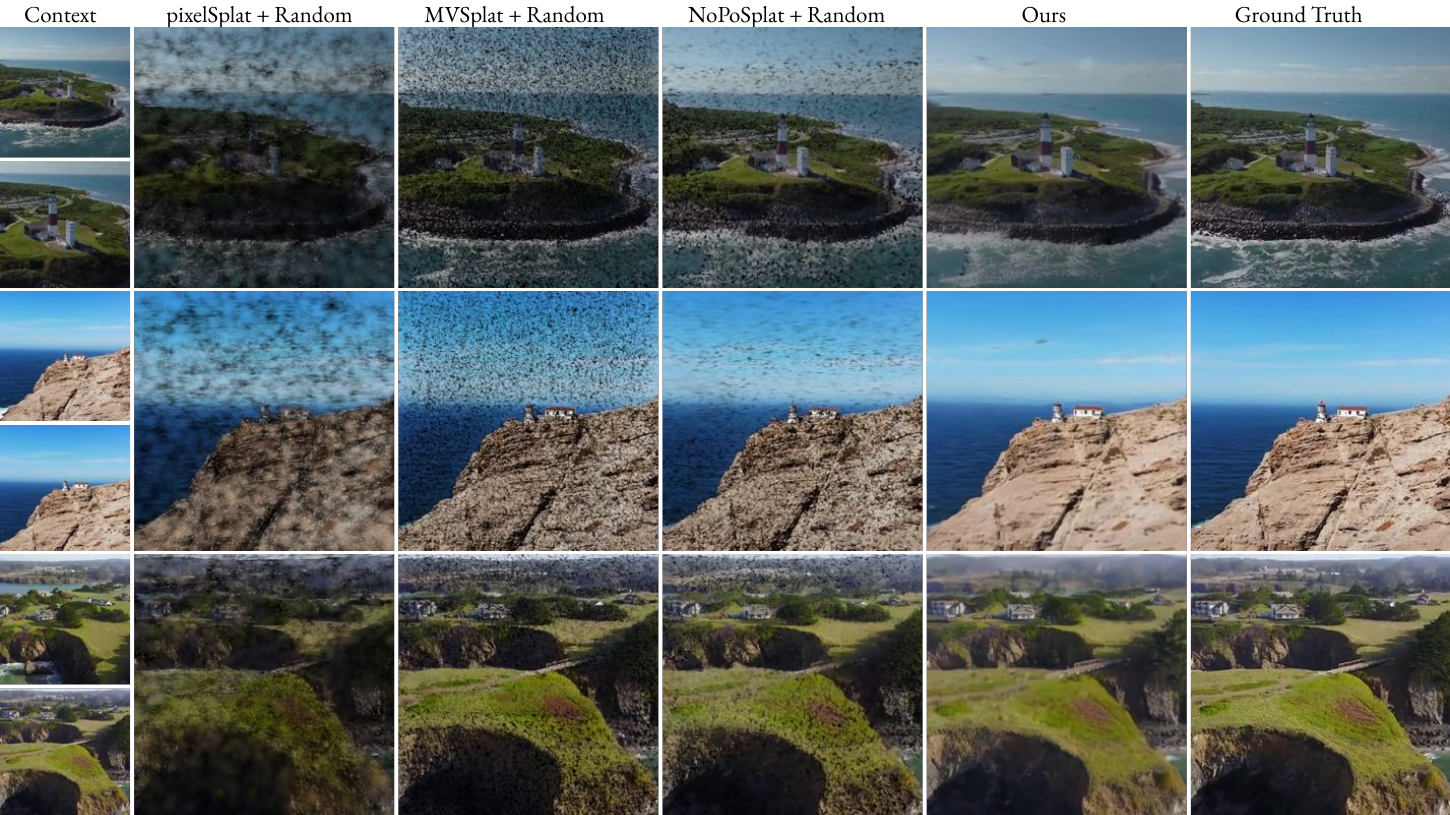}
    \caption{Detailed Results comparison with $\beta=0.5$ on ACID dataset}
    \label{fig:acid_65k_splats}
\end{figure}

\begin{figure}[htbp]
    \vspace{-2mm}
    \centering
    \includegraphics[width=0.8\textwidth]{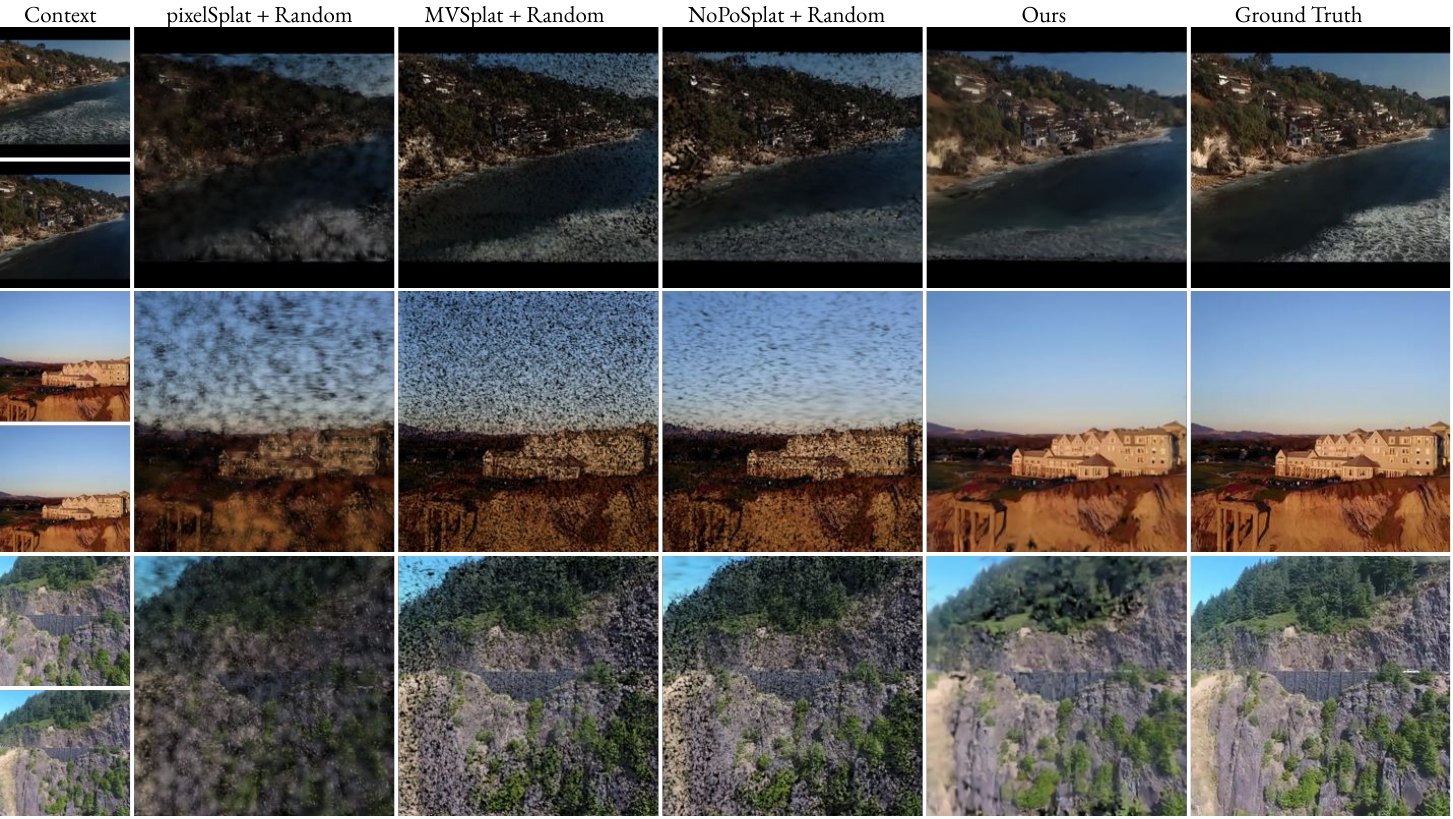}
    \caption{ Detailed Results comparison with $\beta=0.6$ on ACID dataset}
    \label{fig:acid_52k_splats}
\end{figure}

\begin{figure}[htbp]
    \vspace{-4mm}
    \centering
    \includegraphics[width=0.8\textwidth]{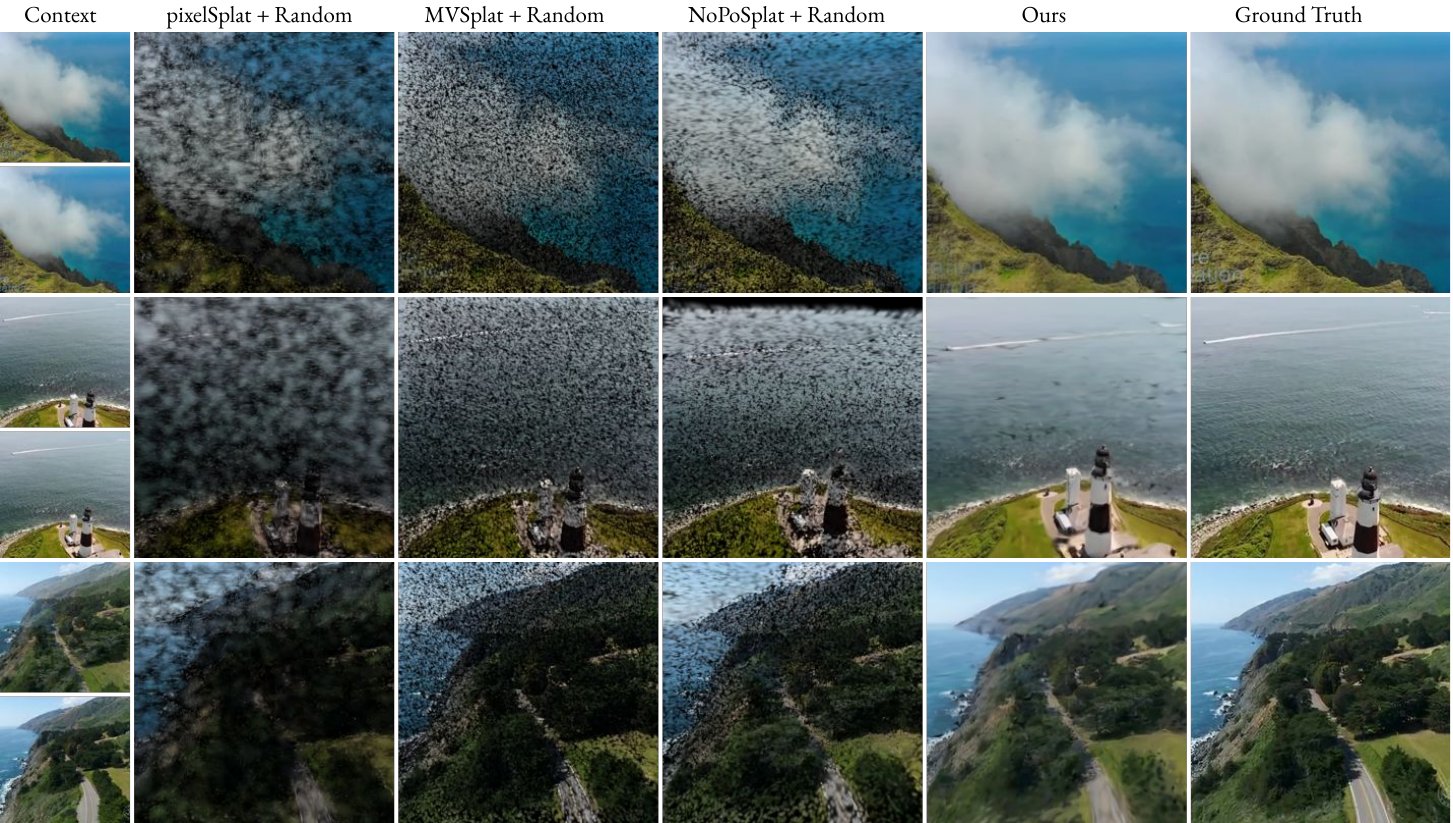}
    \caption{ Detailed Results comparison with $\beta=0.7$ on ACID dataset}
    \label{fig:acid_39k_splats}
\end{figure}

\begin{figure}[htbp]
    \centering
    \includegraphics[width=0.8\textwidth]{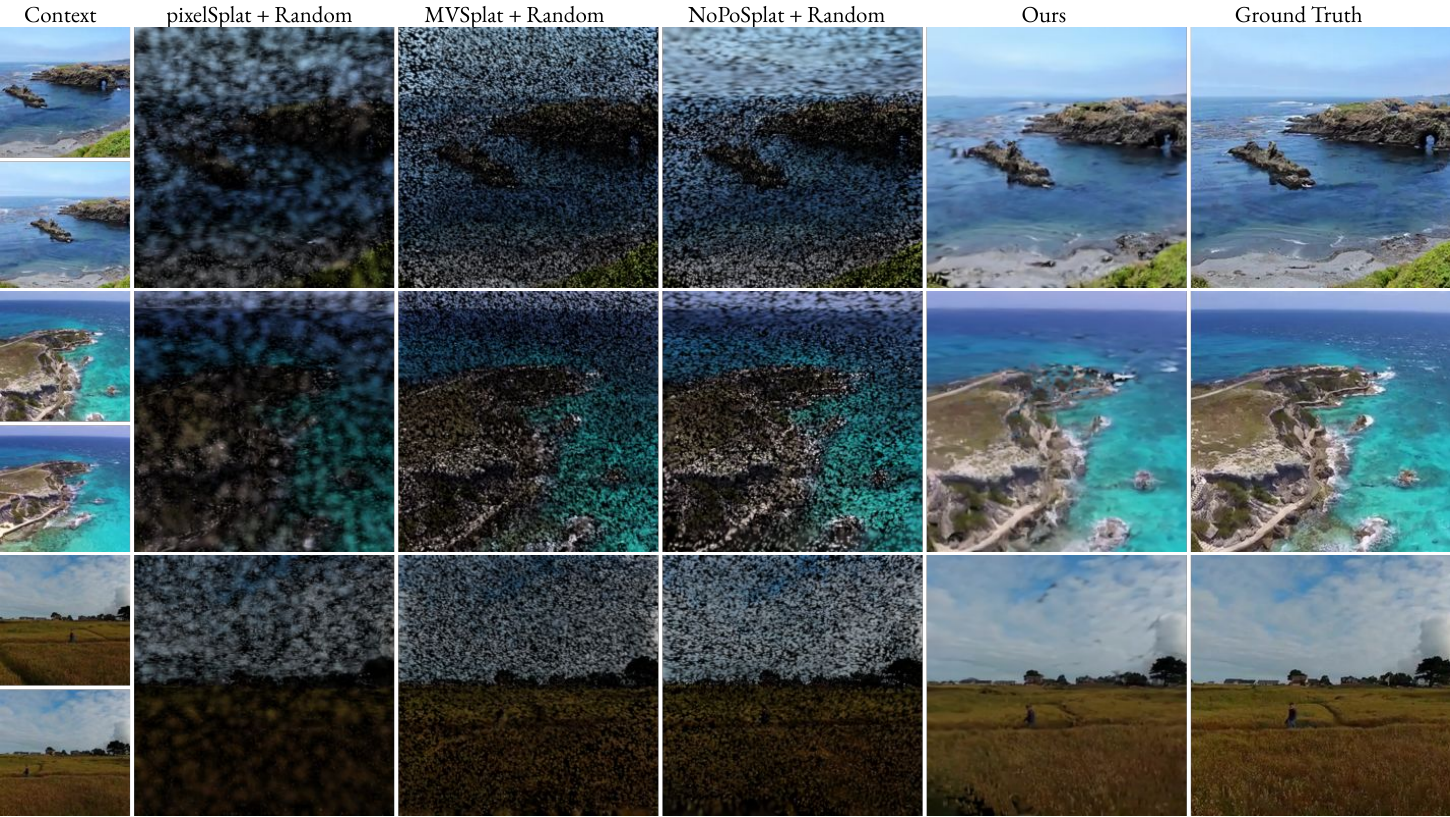}
    \caption{Detailed  Results comparison with $\beta=0.8$ on ACID dataset}
    \label{fig:acid_26k_splats}
\end{figure}

\begin{figure}[htbp]
    \centering
    \includegraphics[width=0.8\textwidth]{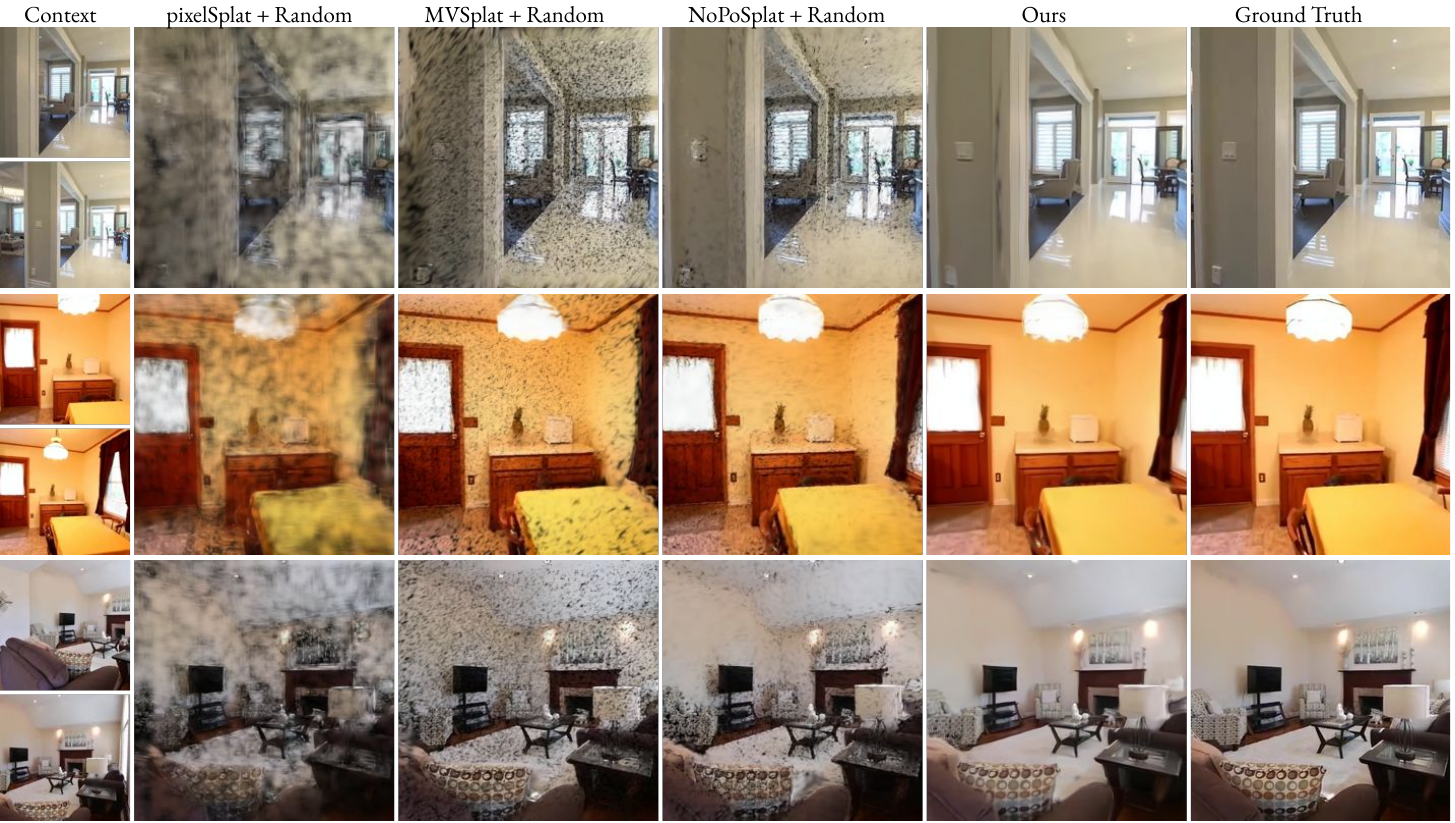}
    \caption{Detailed  Results comparison with $\beta=0.4$ on RE10K dataset}
    \label{fig:re10k_78k_splats}
\end{figure}

\begin{figure}[htbp]
    \centering
    \includegraphics[width=0.8\textwidth]{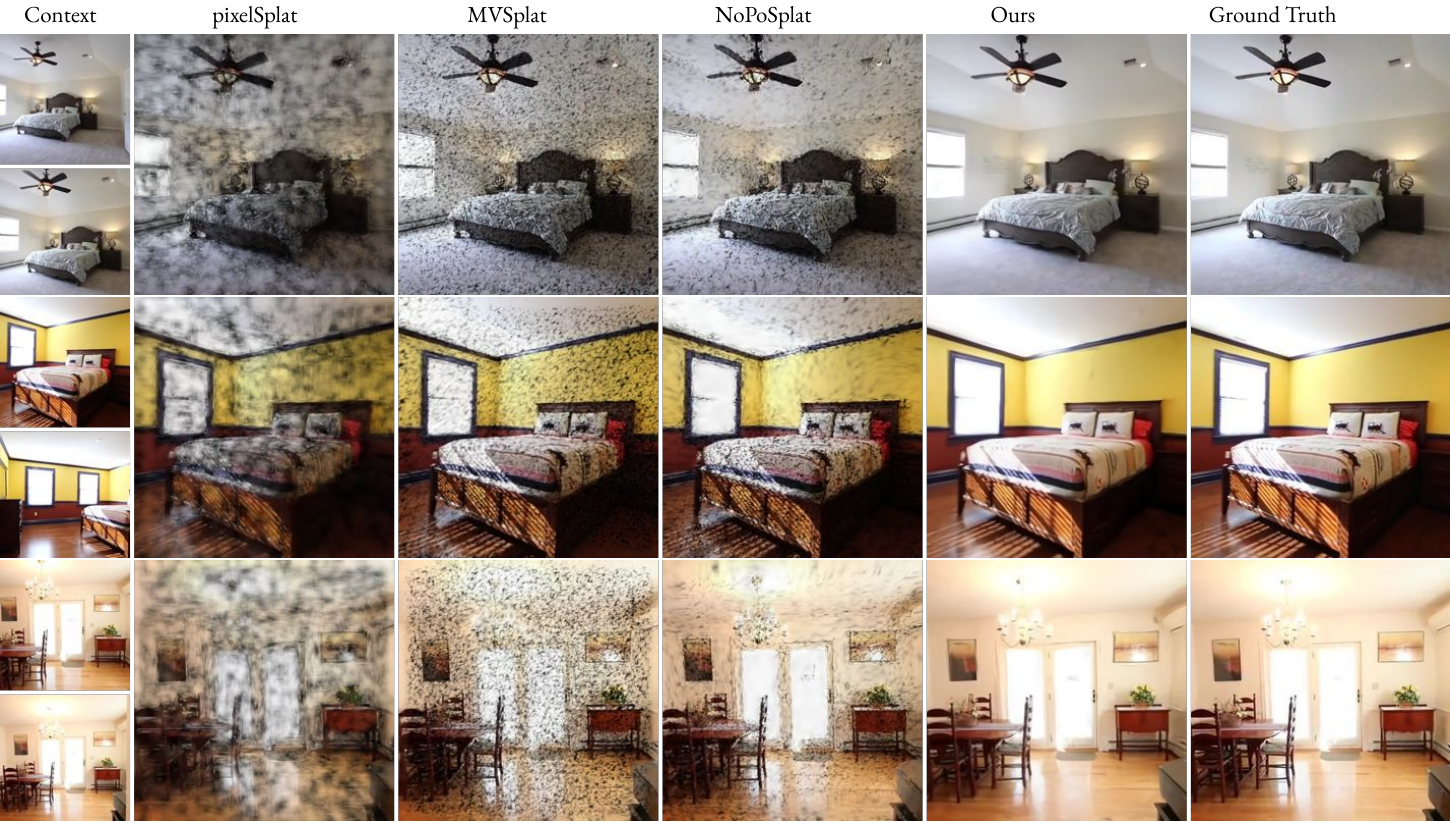}
    \caption{Detailed  Results comparison with $\beta=0.5$ on RE10K dataset}
    \label{fig:re10k_65k_splats}
\end{figure}

\begin{figure}[htbp]
    \centering
    \includegraphics[width=0.8\textwidth]{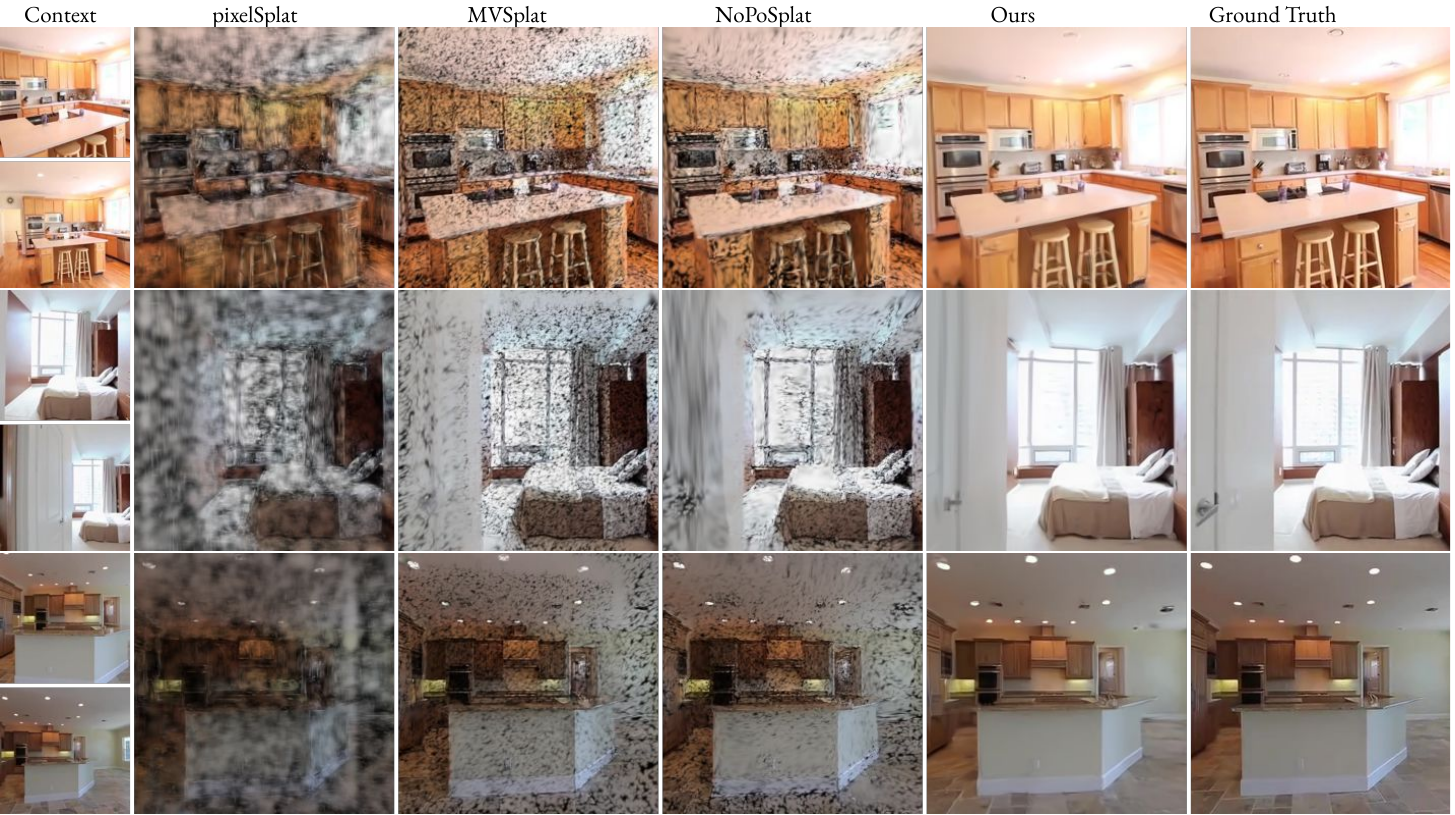}
    \caption{Detailed Results comparison with $\beta=0.6$ on RE10K dataset}
    \label{fig:re10k_52k_splats}
\end{figure}

\begin{figure}[htbp]
    \centering
    \includegraphics[width=0.8\textwidth]{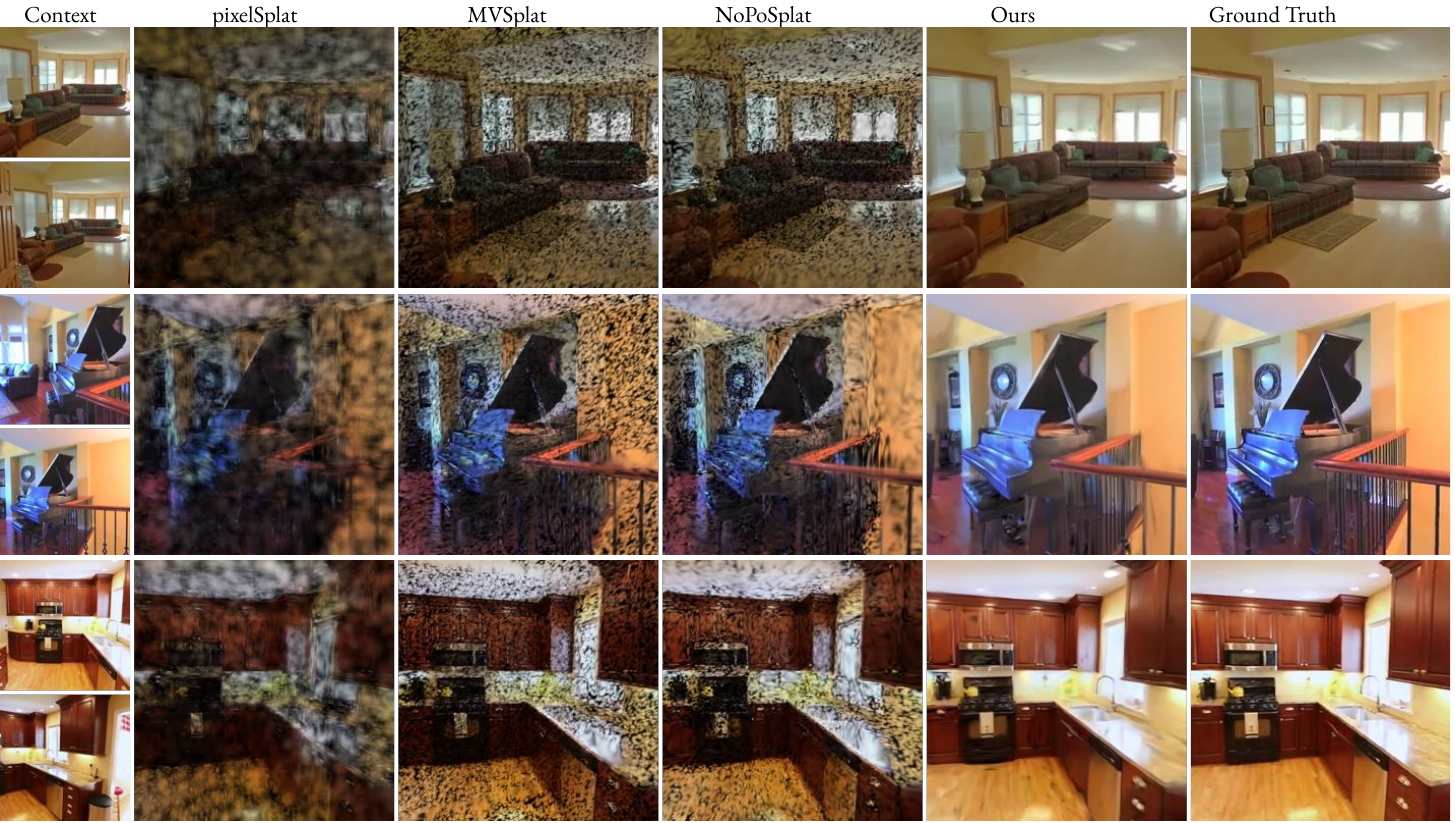}
    \caption{Detailed Results comparison with $\beta=0.7$ on RE10K dataset}
    \label{fig:re10k_39k_splats}
\end{figure}

\begin{figure}[htbp]
    \centering
    \includegraphics[width=0.8\textwidth]{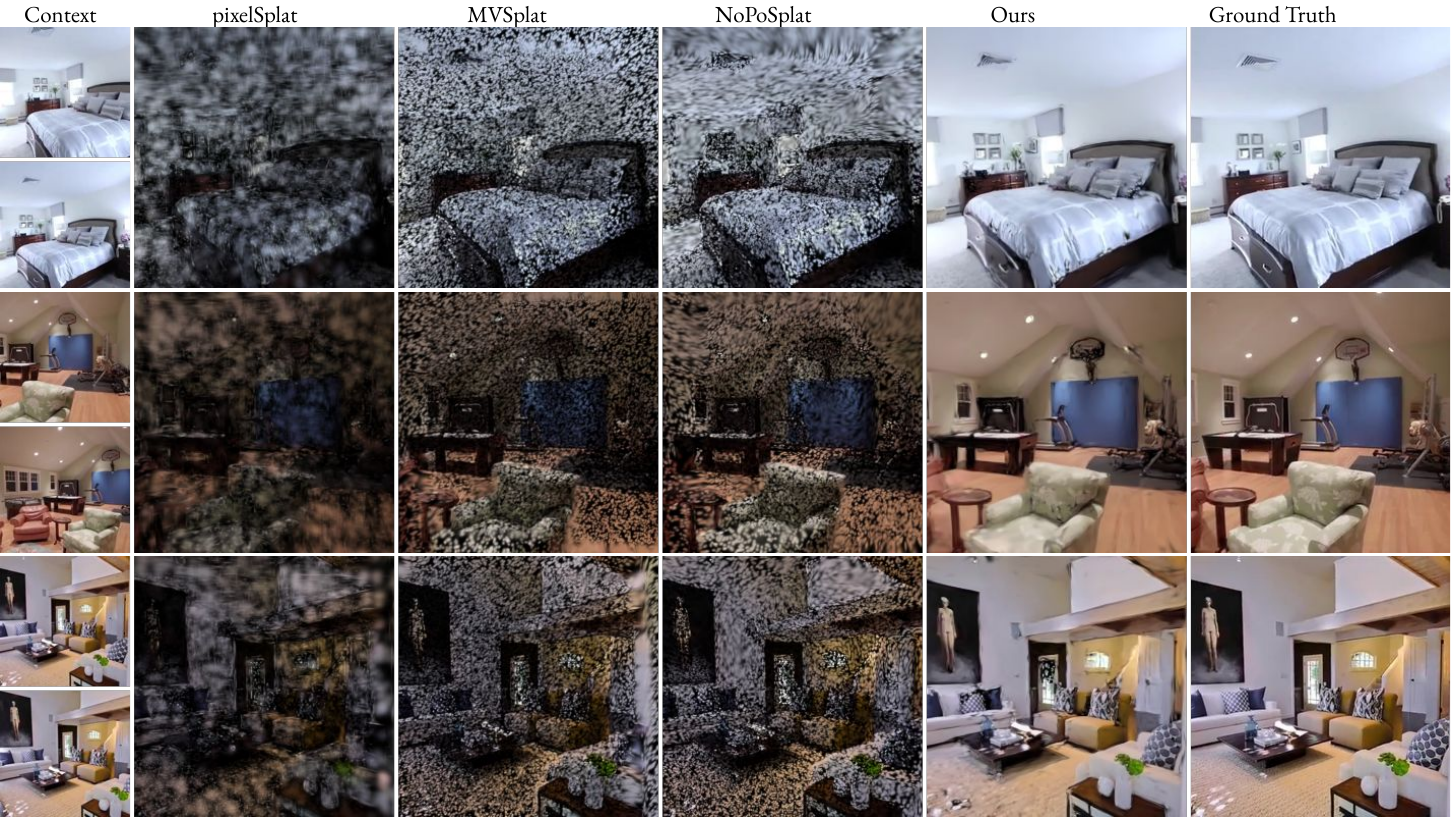}
    \caption{Detailed Results comparison with $\beta=0.8$ on RE10K dataset}
    \label{fig:re10k_26k_splats}
\end{figure}

We can see from the plot that we consistently outperform all the baselines in all pruning scenarios \cref{fig:acid_78k_splats}, \cref{fig:acid_65k_splats}, \cref{fig:acid_52k_splats}, \cref{fig:acid_39k_splats}, \cref{fig:acid_26k_splats} shows visual results on ACID dataset for pruning stages of $\beta=0.4$ ($40\%$), $\beta=0.5$ ($50\%$), $\beta=0.6$ ($60\%$), $\beta=0.7$ ($70\%$) and $\beta=0.8$ ($80\%$) of 3D Gaussians where we take random pruning strategy on various feed-forward backbone. Similarly \cref{fig:re10k_78k_splats} \cref{fig:re10k_65k_splats} , \cref{fig:re10k_52k_splats}, \cref{fig:re10k_39k_splats}, \cref{fig:re10k_26k_splats} shows results on RE10K dataset for different backbone with random pruning of 3D Gaussians.  

\begin{figure}[htbp]
    \centering
    \includegraphics[width=\linewidth]{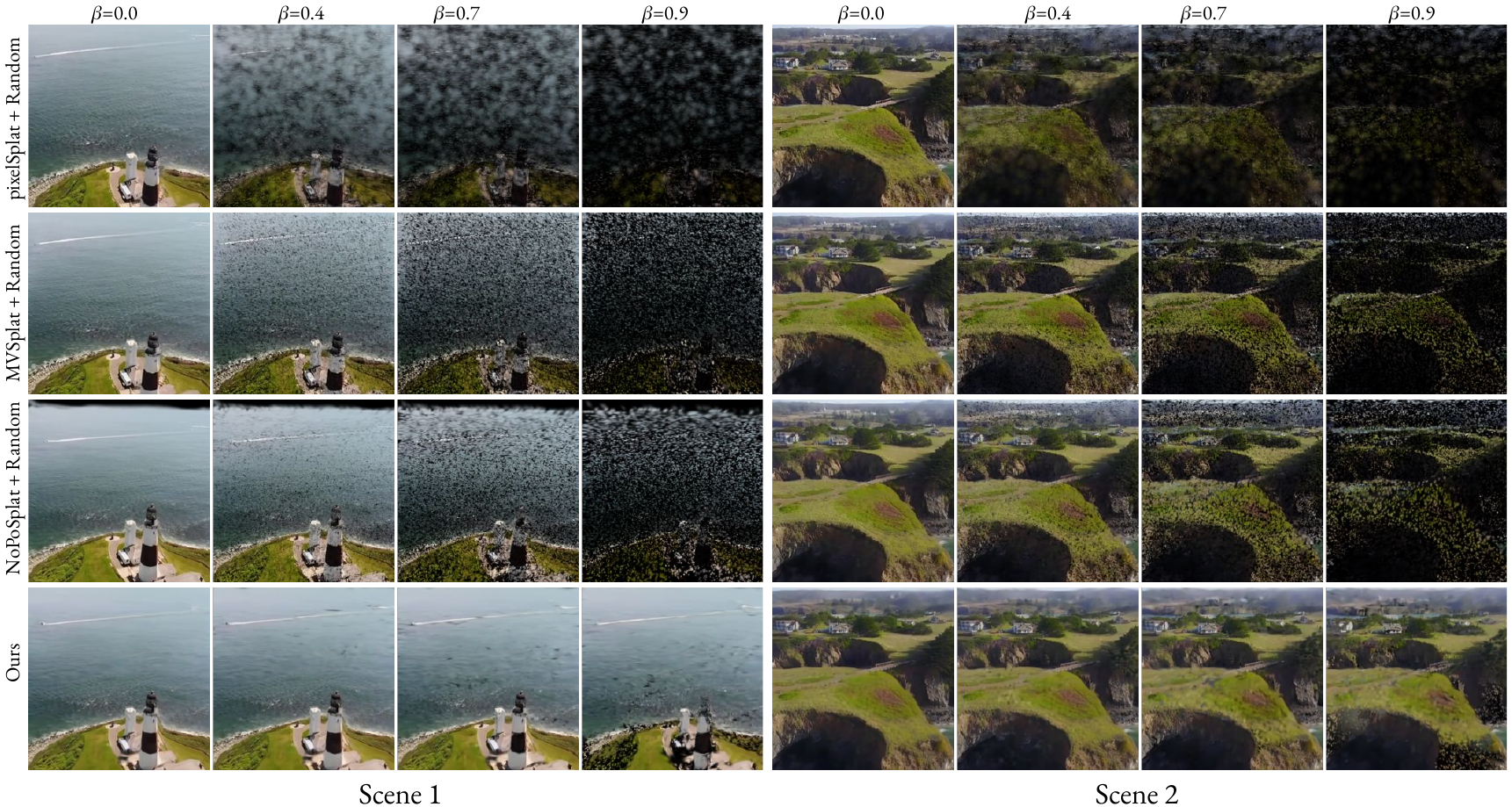}
    
    \vspace{0.5cm} % Adds some space between the images

    \includegraphics[width=\linewidth]{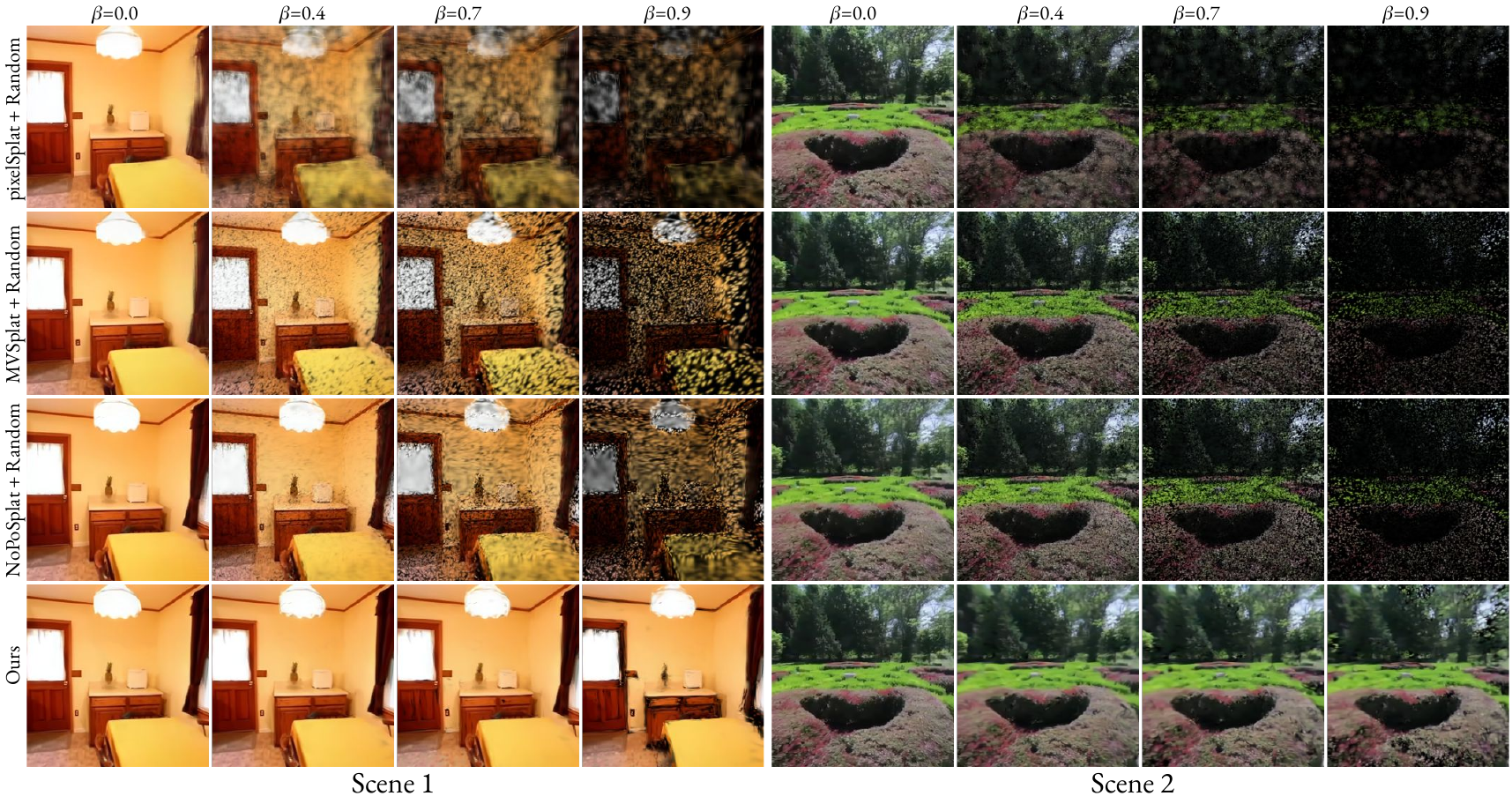}
    
    \caption{Detailed Comparison of baselines results for ACID (top) and RE10K (bottom) datasets. We perform random pruning on top of the 3D Gaussians obtained from other baselines.}
    \label{fig:detailed_pruning_comparison_random}
\end{figure}

\begin{figure}[htbp]
    \centering
    \includegraphics[width=\linewidth]{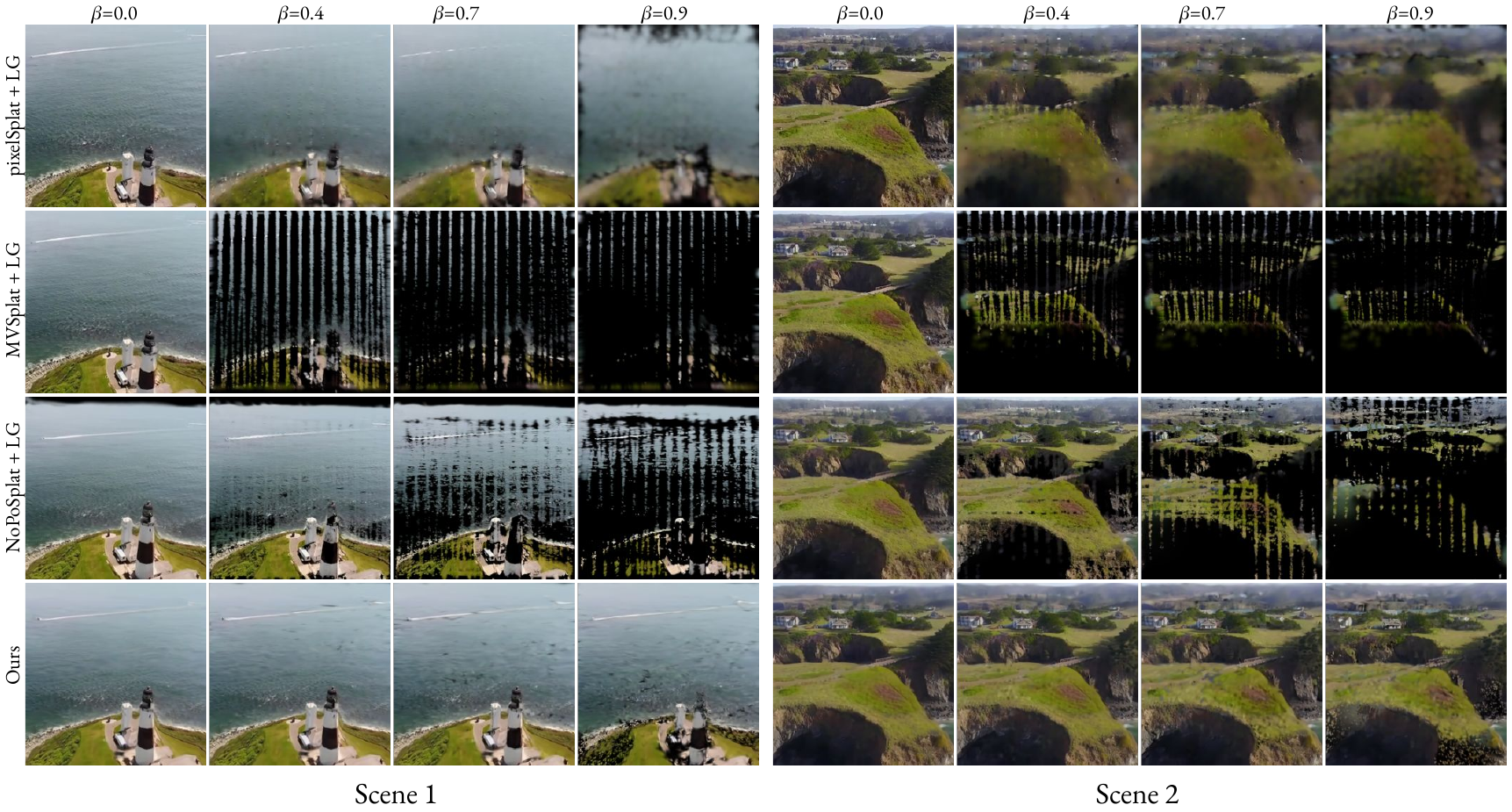}
    
    \vspace{0.5cm} % Adds some space between the images

    \includegraphics[width=\linewidth]{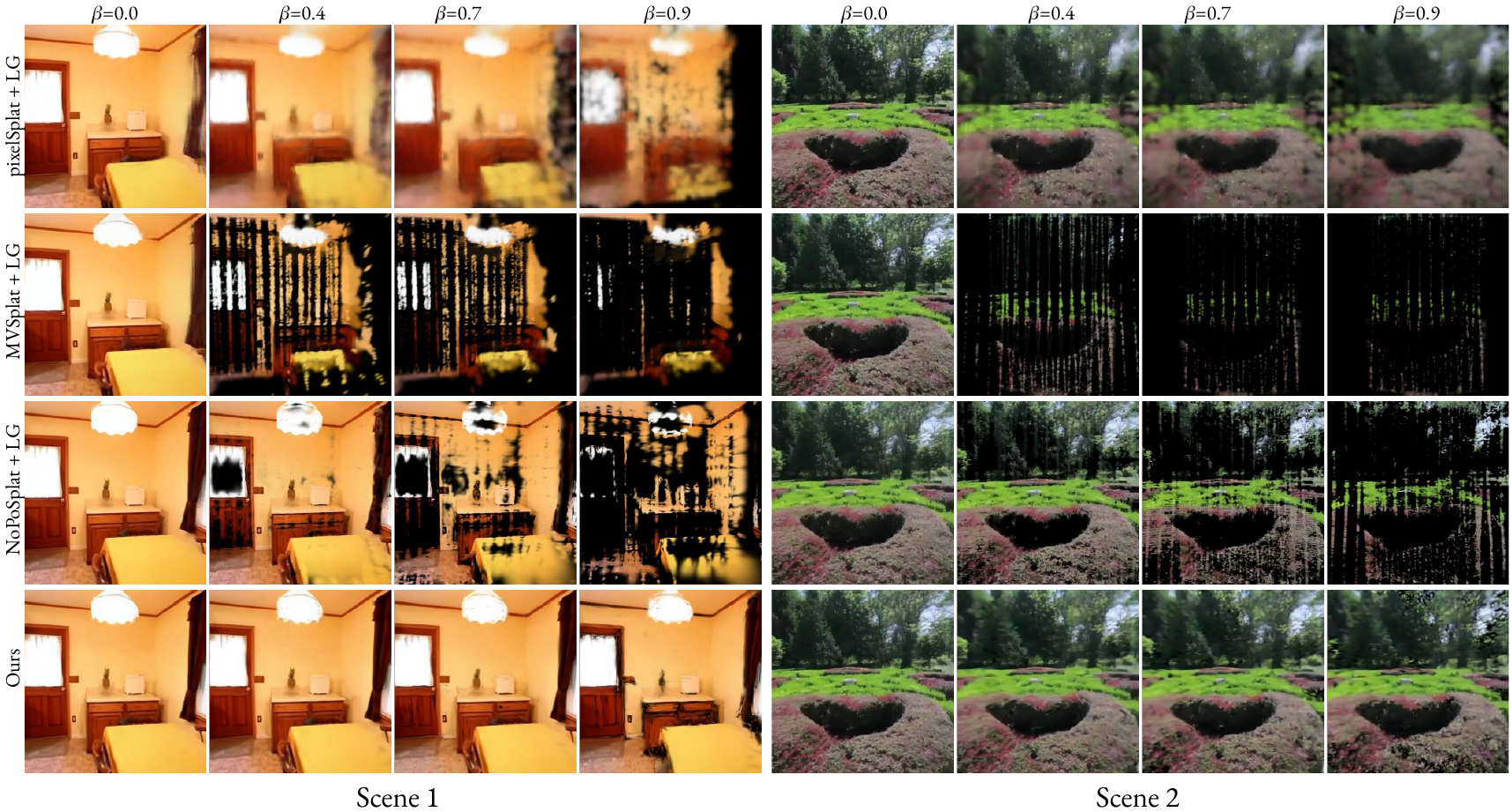}
    
    \caption{Detailed Comparison of baselines results for ACID (top) and RE10K (bottom) datasets. We perform LightGaussian pruning on top of the 3D Gaussians obtained from other baselines.}
    \label{fig:detailed_pruning_comparison_light}
\end{figure}

\cref{fig:detailed_pruning_comparison_random} presents an example scene evaluated at different 3D Gaussian counts on ACID and RE10K, where we use random pruning on different feed-forward backbone. Similarly, \cref{fig:detailed_pruning_comparison_light} shows the same scene compared against baselines incorporating the LightGaussian pruning method. 

\clearpage

\section{Cross domain evaluation}
\label{sec:cross-doman-eval}
\begin{figure}[htbp]
    \centering
    \includegraphics[width=0.7\linewidth]{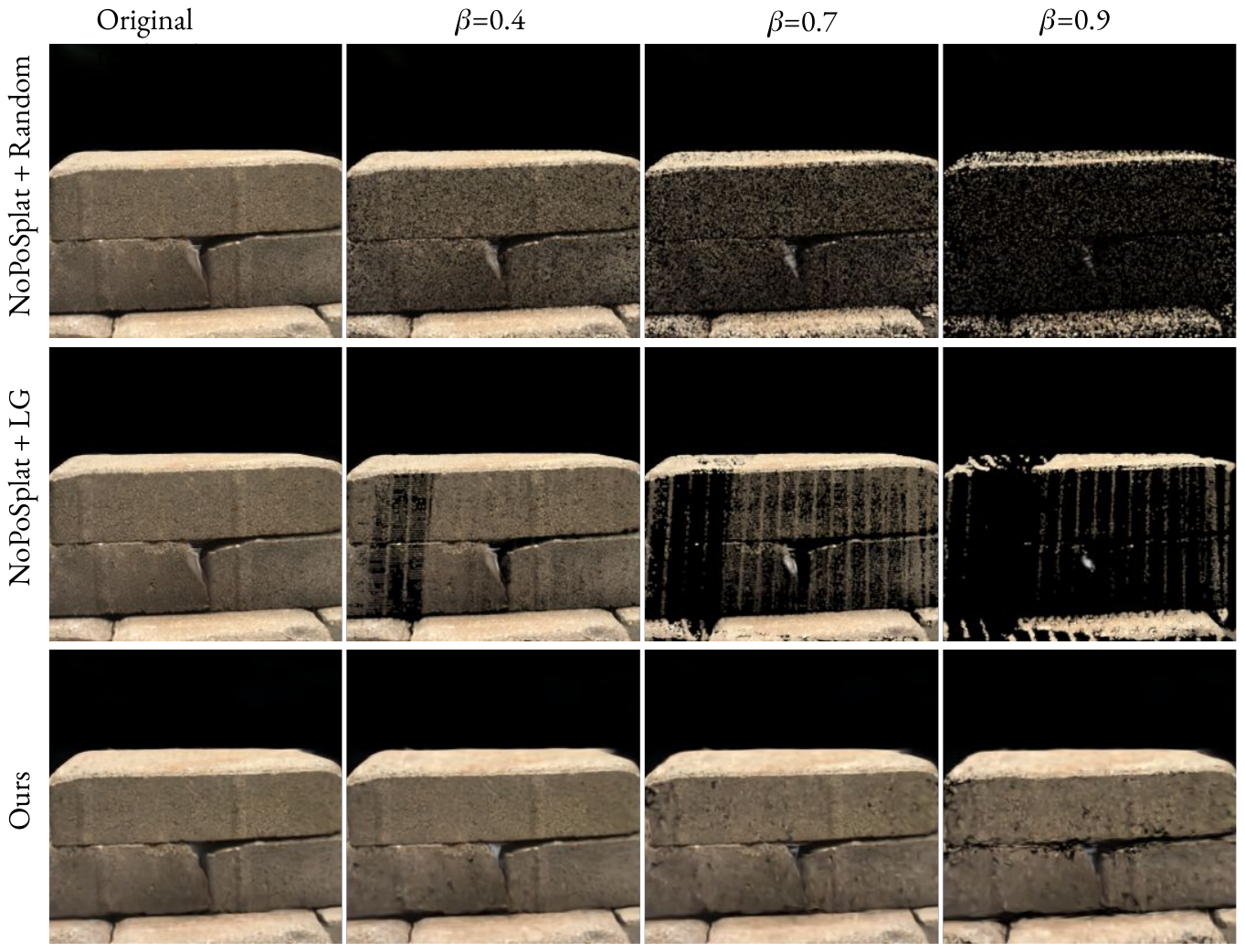}
    \caption{\textbf{Cross-Domain Generalization on DTU}: We train our method on the ACID dataset and evaluate its performance on the DTU dataset to test its out-of-distribution generalization capabilities. Our method outperforms baseline approaches in zero-shot pruning, demonstrating its ability to generalize across different datasets without prior training on the target domain.}
    \label{fig:DTU_re10k}
\end{figure}

We also evaluate the cross domain performance of our model by training the model on one particular dataset and evaluating on the other. pecifically, we trained our model on the ACID dataset and evaluated on the DTU dataset; similarly, we also trained on the RE10K dataset and evaluated on the DTU dataset.
\cref{tab:cross_domain_ablation_results_dtu_acid} shows results where we train the baseline on ACID dataset and evaluate on DTU dataset whereas \cref{tab:cross_domain_ablation_results_dtu_re10k} is trained on RE10K dataset and evaluated on DTU dataset. The baseline that we use is specifically built on NoPoSplat where we use LightGaussian pruning as one baseline and random pruning as another. We use only NoPoSplat as pixelSplat and MVSplat are not trained on large scale diverse dataset as MASt3R used in NoPoSplat. The \cref{fig:DTU_re10k} shows that our method achieves superior results compared to baselines for the same reason as we discussed in main paper. 

\begin{table}[!tb]
\centering
\setlength{\tabcolsep}{3pt}
\caption{\textbf{Out-of-distribution performance comparison: Model trained on ACID and evaluated on DTU}}
\resizebox{\linewidth}{!}{
\begin{tabular}{l|ccc|ccc|ccc|ccc|ccc|ccc}
\toprule
\textbf{Method} & \multicolumn{3}{c|}{$\beta = 0.4$} & \multicolumn{3}{c|}{$\beta = 0.5$} & \multicolumn{3}{c|}{$\beta = 0.6$} & \multicolumn{3}{c|}{$\beta = 0.7$} & \multicolumn{3}{c|}{$\beta = 0.8$} & \multicolumn{3}{c}{$\beta = 0.9$} \\
& \textbf{PSNR$\uparrow$} & \textbf{LPIPS$\downarrow$} & \textbf{SSIM$\uparrow$} & \textbf{PSNR$\uparrow$} & \textbf{LPIPS$\downarrow$} & \textbf{SSIM$\uparrow$} & \textbf{PSNR$\uparrow$} & \textbf{LPIPS$\downarrow$} & \textbf{SSIM$\uparrow$} & \textbf{PSNR$\uparrow$} & \textbf{LPIPS$\downarrow$} & \textbf{SSIM$\uparrow$} & \textbf{PSNR$\uparrow$} & \textbf{LPIPS$\downarrow$} & \textbf{SSIM$\uparrow$} & \textbf{PSNR$\uparrow$} & \textbf{LPIPS$\downarrow$} & \textbf{SSIM$\uparrow$} \\
\midrule
NoPoSplat        & 14.84 & 0.444 & 0.358 & 14.09 & 0.479 & 0.324 & 13.15 & 0.514 & 0.286 & 11.98 & 0.550 & 0.242 & 10.59 & 0.588 & 0.192 & 8.86 & 0.634 & 0.129 \\
NoPo + LightGaus & 14.85 & 0.399 & 0.372 & 13.97 & 0.440 & 0.330 & 12.85 & 0.485 & 0.278 & 11.59 & 0.536 & 0.218 & 10.25 & 0.587 & 0.154 & 8.73 & 0.646 & 0.089 \\
\midrule
\textbf{Ours}    & \textbf{15.23} & 0.424 & \textbf{0.427} & \textbf{15.17} & \textbf{0.430} & \textbf{0.422} & \textbf{15.12} & \textbf{0.437} & \textbf{0.417} & \textbf{15.01} & \textbf{0.446} & \textbf{0.411} & \textbf{14.76} & \textbf{0.460} & \textbf{0.403} & \textbf{13.98} & \textbf{0.479} & \textbf{0.391} \\
\bottomrule
\end{tabular}%
}

\label{tab:cross_domain_ablation_results_dtu_acid}
\end{table}

\begin{table}[!tb]
\centering
\setlength{\tabcolsep}{4pt}
\caption{Out-of-distribution performance comparison: Model trained on RE10K and evaluated on DTU}
\resizebox{\linewidth}{!}{
\begin{tabular}{l|ccc|ccc|ccc|ccc|ccc|ccc}
\toprule
\textbf{Method} & \multicolumn{3}{c|}{$\beta = 0.4$} & \multicolumn{3}{c|}{$\beta = 0.5$} & \multicolumn{3}{c|}{$\beta = 0.6$} & \multicolumn{3}{c|}{$\beta = 0.7$} & \multicolumn{3}{c|}{$\beta = 0.8$} & \multicolumn{3}{c}{$\beta = 0.9$} \\
& \textbf{PSNR}$\uparrow$ & \textbf{LPIPS}$\downarrow$ & \textbf{SSIM}$\uparrow$ & \textbf{PSNR}$\uparrow$ & \textbf{LPIPS}$\downarrow$ & \textbf{SSIM}$\uparrow$ & \textbf{PSNR}$\uparrow$ & \textbf{LPIPS}$\downarrow$ & \textbf{SSIM}$\uparrow$ & \textbf{PSNR}$\uparrow$ & \textbf{LPIPS}$\downarrow$ & \textbf{SSIM}$\uparrow$ & \textbf{PSNR}$\uparrow$ & \textbf{LPIPS}$\downarrow$ & \textbf{SSIM}$\uparrow$ & \textbf{PSNR}$\uparrow$ & \textbf{LPIPS}$\downarrow$ & \textbf{SSIM}$\uparrow$ \\
\midrule
NoPoSplat        & 16.18 & 0.4053 & \textbf{0.5160} & 15.22 & 0.4453 & 0.4678 & 14.05 & 0.4857 & 0.4127 & 12.62 & 0.5258 & 0.3489 & 10.97 & 0.5674 & 0.2739 & 9.00 & 0.6146 & 0.1789 \\
NoPo + LightGaus & \textbf{16.26} & \textbf{0.3547} & 0.5033 & 15.10 & \textbf{0.3981} & 0.4482 & 13.50 & 0.4522 & 0.3791 & 11.93 & 0.5114 & 0.2983 & 10.40 & 0.5693 & 0.2082 & 8.80 & 0.6339 & 0.1165 \\
\midrule
\textbf{Ours}    & 16.19 & 0.4073 & 0.4746 & \textbf{16.14} & 0.4153 & \textbf{0.4693} & \textbf{16.02} & \textbf{0.4252} & \textbf{0.4606} & \textbf{15.81} & \textbf{0.4368} & \textbf{0.4513} & \textbf{15.38} & \textbf{0.4519} & \textbf{0.4389} & \textbf{14.29} & \textbf{0.4764} & \textbf{0.4232} \\
\bottomrule
\end{tabular}%
}

\label{tab:cross_domain_ablation_results_dtu_re10k}
\end{table}

\section{Failure Case}
\label{sec:failure_case}
\begin{figure}[htbp]
\centering
\includegraphics[width=\linewidth]{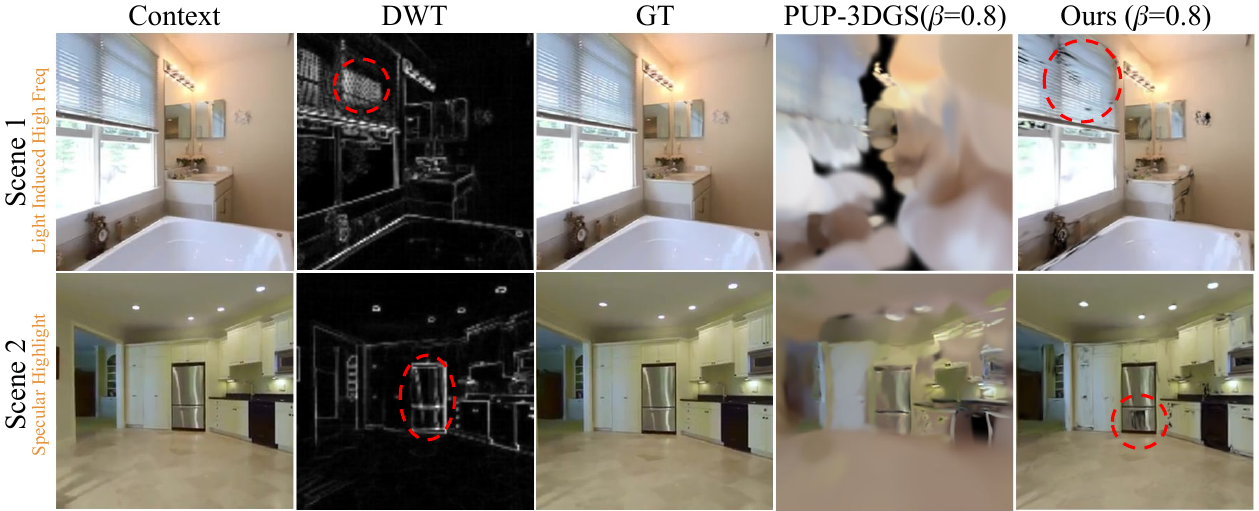}
\caption{DWT misassigns high energy to a small sub-region of a flat SuperCluster, degrading quality at high $\beta$.}
\label{fig:dwt_failure}
\end{figure}

\textbf{DWT limitation examples:} As shown in \cref{fig:dwt_failure}, due to light-induced high frequencies and specular highlights, DWT provides a misleading signal by assigning high energy to Gaussians concentrated in a small sub-region within the marked flat SuperCluster. This results in noticeable degradation at high $\beta$, whereas our method preserves quality in the remaining region. Addressing this limitation is left for future work.

\section{Additional Ablation}
\label{sec:additional_ablation_supplementary}

\subsection{Choice of k in K-Means Clustering}
\label{sec:k-means-choice}
Our scene partitioning relies on k-means clustering to form SuperClusters. An ablation study on the ACID dataset (\cref{tab:k-means-ablation_results_mainpaper}) reveals that increasing k generally improves reconstruction quality. This is because a higher k yields finer-grained SuperClusters, enabling more precise identification of low-texture regions and, consequently, more judicious placement of larger 3D Gaussians for an optimized trade-off between sparsity and visual fidelity. However, increasing $k$ beyond $300$ yields only marginal improvements in quality while significantly increasing training time. \cref{tab:withoutmask-ablation_results_supplementary} further reports the contribution of each component (GS Head training, contextual mask, structured pruning) on ACID at $\beta=0.4$ and $\beta=0.9$.

\subsection{Choice of $\gamma$ values and percentage of retained 3D Gaussians}
\label{sec:gamma-value-ablation-supplementary}
\cref{tab:different_retained_percentage_mainpaper} presents the results of our method when retaining various percentages of 3D Gaussians within low-texture SuperClusters. We observe that retaining $\gamma=0.10$ ($10\%$) of 3D Gaussians yields the best performance, with this optimal balance becoming particularly evident at high pruning levels (e.g., $13k$ ($\beta=0.9$) 3D Gaussians). This is because aggressive pruning necessitates a precise balance of sparse, representative Gaussians that can effectively fill vacated regions without creating excessive overlap or redundancy. We notice that with increasing $\gamma$ of primitives in a low-texture SuperCluster, we see a drop in the novel view metric. This is because increasing $\gamma$ leads to redundant Gaussians in the same region, which leads to suboptimal quality. This insight is also exploited in previous works such as LightGaussian, MiniSplatting, GenerativeDensification, EAGLES, etc.

\subsection{Pruning by choosing SuperClusters randomly:} 
\label{sec:prune-supercluster-randomly}
\begin{figure}[htbp]
    \centering
    \includegraphics[width=0.8\linewidth]{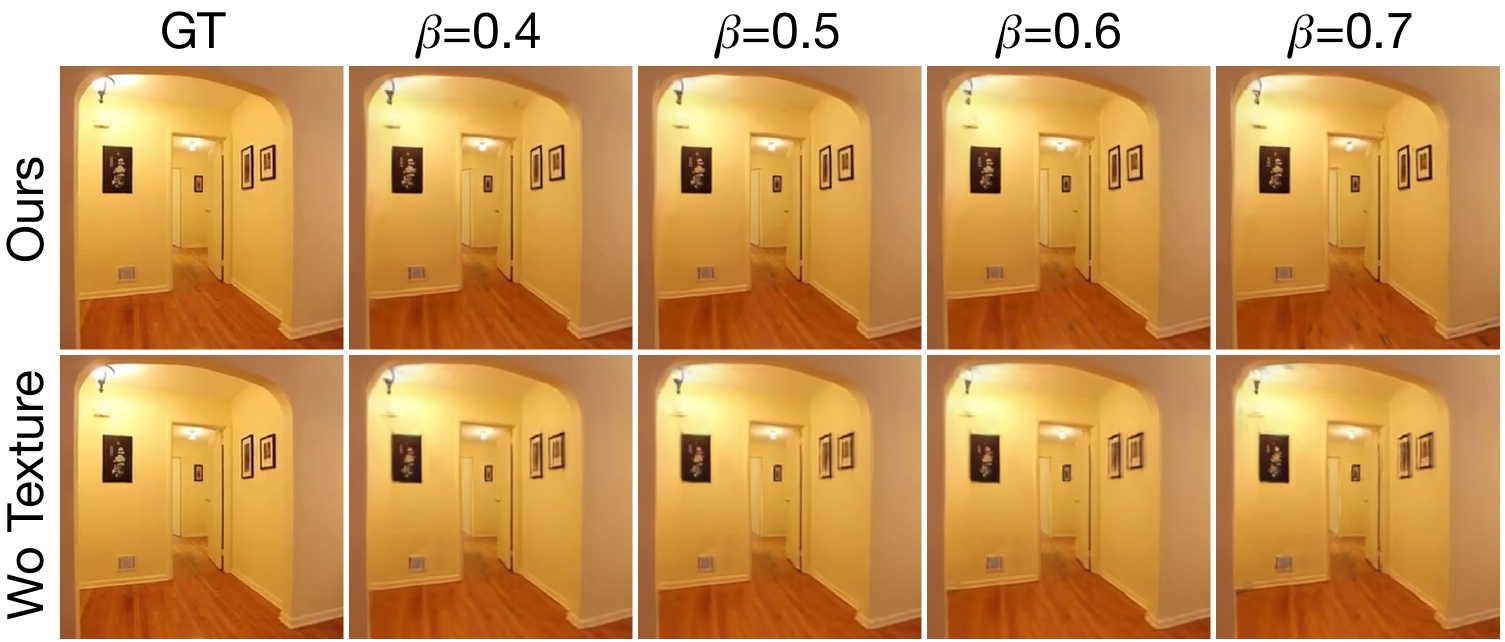}
    \caption{Ablation on RE10K when SuperClusters are selected randomly.}
    \label{fig:ablation_re10k_withouttexture}
\end{figure}

\begin{figure}[htbp]
    \centering
    \includegraphics[width=0.8\linewidth]{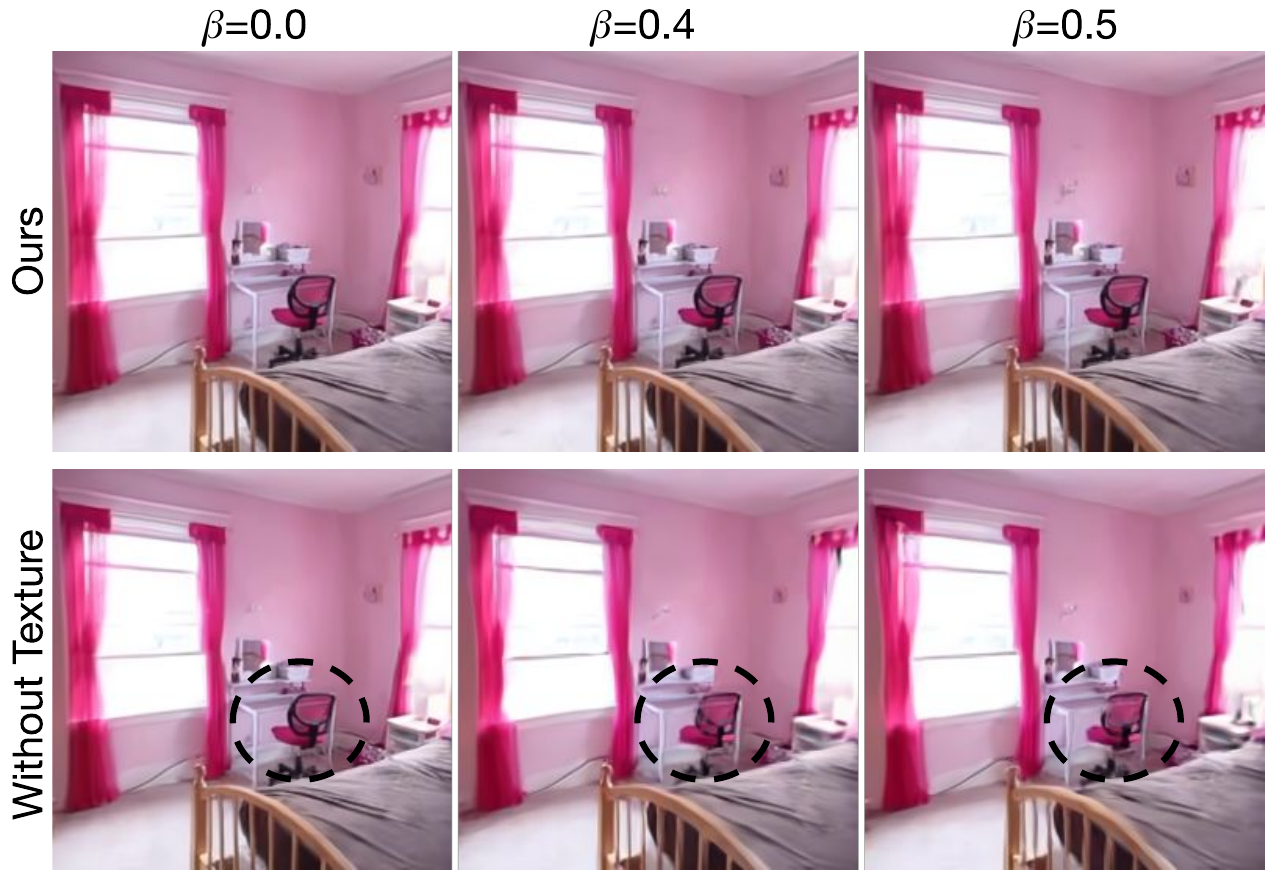}
    \caption{\textbf{Results of Ablations on RE10K:} The ablation method begins to lose information around the edges of the chair, particularly in the circular area, while our method retains the details more effectively.}
    \label{fig:ablation_results_image}
\end{figure}

Another ablation we perform is where we remove the texture aware pruning and choose SuperClusters randomly instead of taking texture into account. This way we are pruning the 3D Gaussians within a SuperCluster but the SuperCluster is chosen randomly. The results for the ablation are presented in \cref{tab:ablation_comparison_re10k_all} and \cref{fig:ablation_re10k_withouttexture} for RE10K dataset and in \cref{tab:ablation_comparison_acid_all} for ACID dataset. The table shows that texture aware selection of SuperCluster is providing better results in most of the cases compared to random selection. \cref{fig:ablation_results_image} shows results for artifacts observed in case we use random SuperCluster for pruning rather than texture based method.

\begin{table}[!tb]
\centering
\setlength{\tabcolsep}{8pt}
\caption{\textbf{Trade-off Between $k$ and Resource Usage in Pruning.} Comparison of results at different pruning stages for various $k$ values used in scene partitioning. The second column shows the memory consumed during $k-means$ clustering for each $k$. A higher $k$ generally allows for finer partitioning but also increases runtime.}
\resizebox{\linewidth}{!}{
\begin{tabular}{@{}c|ccc|ccc@{}}
\toprule
\textbf{ACID} & \multicolumn{3}{c|}{\textbf{$\beta=0.4$}}                           & \multicolumn{3}{c}{\textbf{$\beta=0.9$}}                            \\ \midrule
\textbf{k value}      & \textbf{PSNR$\uparrow$} & \textbf{LPIPS$\downarrow$} & \textbf{SSIM$\uparrow$} & \textbf{PSNR$\uparrow$} & \textbf{LPIPS$\downarrow$} & \textbf{SSIM$\uparrow$} \\ \midrule
k = 5                 & 22.359                  & 0.298             & 0.643                   & 18.054                  & 0.386             & 0.547                   \\
k=10                  & 22.414                  & 0.298             & 0.642                   & 18.057                  & 0.385             & 0.547                   \\
k=50                  & 22.416                  & 0.303             & 0.636                   & 18.063                  & 0.385             & 0.547                   \\
k=100                 & 22.488                  & 0.301             & 0.638                   & 18.066                  & 0.385             & 0.547                   \\
k=300                 & 22.549                  & 0.299             & 0.640                   & 18.476                  & 0.379             & 0.553                   \\
k=400                 & 22.559                  & 0.298             & 0.641                   & 18.146                  & 0.383             & 0.548                   \\ \bottomrule
\end{tabular}
}
    \label{tab:k-means-ablation_results_mainpaper}
\end{table}

\begin{table}[!tb]
\centering
\small
\setlength{\tabcolsep}{3pt}
\caption{\textbf{Ablation on ACID dataset.} We ablate various components of our method, demonstrating their importance.}
\resizebox{0.8\linewidth}{!}{
\begin{tabular}{l|ccc|ccc}
\toprule
\textbf{Ablation} & \multicolumn{3}{c|}{\textbf{$\beta=0.4$}} & \multicolumn{3}{c}{\textbf{$\beta=0.9$}} \\
\textbf{Component} & \textbf{PSNR$\uparrow$} & \textbf{LPIPS$\downarrow$} & \textbf{SSIM$\uparrow$} & \textbf{PSNR$\uparrow$} & \textbf{LPIPS$\downarrow$} & \textbf{SSIM$\uparrow$} \\
\midrule
w/o GS Head Training & 17.44 & 0.445 & 0.490 & 9.20 & 0.688 & 0.118 \\
Random Pruning Baseline & 22.53 & 0.305 & \textbf{0.653} & 17.50 & 0.392 & 0.544 \\
w/o Contextual Mask & 22.52 & 0.318 & 0.642 & 17.62 & 0.424 & 0.529 \\
\midrule
\textbf{Ours (Full)} & \textbf{22.55} & \textbf{0.299} & 0.639 & \textbf{18.48} & \textbf{0.379} & \textbf{0.552} \\
\bottomrule
\end{tabular}
}

\label{tab:withoutmask-ablation_results_supplementary}
\end{table}

\begin{table}[!tb]
\centering
\setlength{\tabcolsep}{14pt}
\caption{Empirical results on retaining different percentages of 3D Gaussians in low texture SuperCluster, from the results we conclude that retaining $\gamma=0.10$ 3D Gaussians yields the best results}
\resizebox{0.9\linewidth}{!}{
\begin{tabular}{@{}ccccccc@{}}
\toprule
\multicolumn{7}{c}{\textbf{ACID Dataset}}                                                                                                                                                                                                                           \\ \midrule
\multicolumn{1}{c|}{\textbf{}}   & \multicolumn{3}{c|}{\textbf{$\beta=0.4$}}                                                                                     & \multicolumn{3}{c}{\textbf{$\beta=0.9$}}                                                                 \\ \midrule
\multicolumn{1}{c|}{$\gamma$} & \textbf{\textbf{PSNR$\uparrow$}} & \textbf{LPIPS$\downarrow$} & \multicolumn{1}{c|}{\textbf{SSIM$\uparrow$}} & \textbf{PSNR$\uparrow$} & \textbf{LPIPS$\downarrow$} & \textbf{SSIM$\uparrow$} \\ \midrule
\multicolumn{1}{c|}{$\gamma=0.05$}         & 21.796                        & 0.347                            & \multicolumn{1}{c|}{0.608}                         & 17.327                        & 0.430                            & 0.526                         \\
\multicolumn{1}{c|}{$\gamma=0.10$}        & \textbf{22.549}               & \textbf{0.299}                   & \multicolumn{1}{c|}{\textbf{0.640}}                & \textbf{18.106}               & \textbf{0.384}                   & \textbf{0.548}                \\
\multicolumn{1}{c|}{$\gamma=0.15$}        & 21.587                        & 0.352                            & \multicolumn{1}{c|}{0.589}                         & 17.053                        & 0.494                            & 0.429                         \\
\multicolumn{1}{c|}{$\gamma=0.2$}        & 22.262                        & 0.304                            & \multicolumn{1}{c|}{0.629}                         & 17.452                        & 0.451                            & 0.496                         \\ \bottomrule
\end{tabular}
}

\label{tab:different_retained_percentage_mainpaper}
\end{table}

% ALL Gaussians RE10k ablations
\begin{table}[!tb]
\centering
\setlength{\tabcolsep}{8pt}
\caption{Ablation results on RE10K dataset to support texture based SuperCluster Selection. Here we chose SuperClusters randomly instead of using texture. }
\resizebox{0.8\linewidth}{!}{
    \begin{tabular}{llccc}
\toprule
\textbf{$\beta$ Value} & \textbf{Texture Info} & \textbf{PSNR$\uparrow$} & \textbf{LPIPS$\downarrow$} & \textbf{SSIM$\uparrow$} \\ 
\midrule
\multirow{2}{*}{$0.4$} & No Texture & 20.73 & 0.260 & 0.714 \\
                             & \textbf{Ours} & \textbf{22.29} & \textbf{0.235} & \textbf{0.735} \\ 
\midrule
\multirow{2}{*}{$0.5$} & No Texture & 20.22 & 0.269 & 0.705 \\
                             & \textbf{Ours} & \textbf{22.24} & \textbf{0.238} & \textbf{0.732} \\ 
\midrule
\multirow{2}{*}{$0.6$} & No Texture & 19.69 & 0.279 & 0.695 \\
                             & \textbf{Ours} & \textbf{22.11} & \textbf{0.243} & \textbf{0.726} \\ 
\midrule
\multirow{2}{*}{$0.7$} & No Texture & 19.13 & 0.288 & 0.684 \\
                             & \textbf{Ours} & \textbf{21.77} & \textbf{0.252} & \textbf{0.716} \\ 
\midrule
\multirow{2}{*}{$0.8$} & No Texture & 18.54 & 0.298 & 0.671 \\
                             & \textbf{Ours} & \textbf{20.74} & \textbf{0.273} & \textbf{0.693} \\ 
\bottomrule
\end{tabular}
    }
    
    \label{tab:ablation_comparison_re10k_all}
\end{table}

% Ablation ACID random
\begin{table}[!tb]
\centering
\setlength{\tabcolsep}{8pt}
\caption{Ablation results on ACID dataset to support wavelet based SuperCluster Selection}
\resizebox{0.8\linewidth}{!}{
\begin{tabular}{llccc}
\toprule
\textbf{$\beta$ Value} & \textbf{Texture Info} & \textbf{PSNR$\uparrow$} & \textbf{LPIPS$\downarrow$} & \textbf{SSIM$\uparrow$} \\ 
\midrule
\multirow{2}{*}{$0.4$} & No Texture & 21.31 & 0.326 & 0.623 \\
                             & \textbf{Ours} & \textbf{22.55} & \textbf{0.299} & \textbf{0.640} \\ 
\midrule
\multirow{2}{*}{$0.5$} & No Texture & 20.89 & 0.335 & 0.615 \\
                             & \textbf{Ours} & \textbf{22.47} & \textbf{0.303} & \textbf{0.635} \\ 
\midrule
\multirow{2}{*}{$0.6$} & No Texture & 20.44 & 0.344 & 0.606 \\
                             & \textbf{Ours} & \textbf{22.31} & \textbf{0.311} & \textbf{0.627} \\ 
\midrule
\multirow{2}{*}{$0.7$} & No Texture & 19.96 & 0.353 & 0.596 \\
                             & \textbf{Ours} & \textbf{21.96} & \textbf{0.322} & \textbf{0.615} \\ 
\midrule
\multirow{2}{*}{$0.8$} & No Texture & 19.42 & 0.362 & 0.586 \\
                             & \textbf{Ours} & \textbf{21.06} & \textbf{0.342} & \textbf{0.594} \\ 
\bottomrule
\end{tabular}
    }
    
    \label{tab:ablation_comparison_acid_all}
\end{table}
% \clearpage

\subsection{Additional Ablation on RE10K}

\begin{figure}[htbp]
    \centering
    \includegraphics[width=0.8\linewidth]{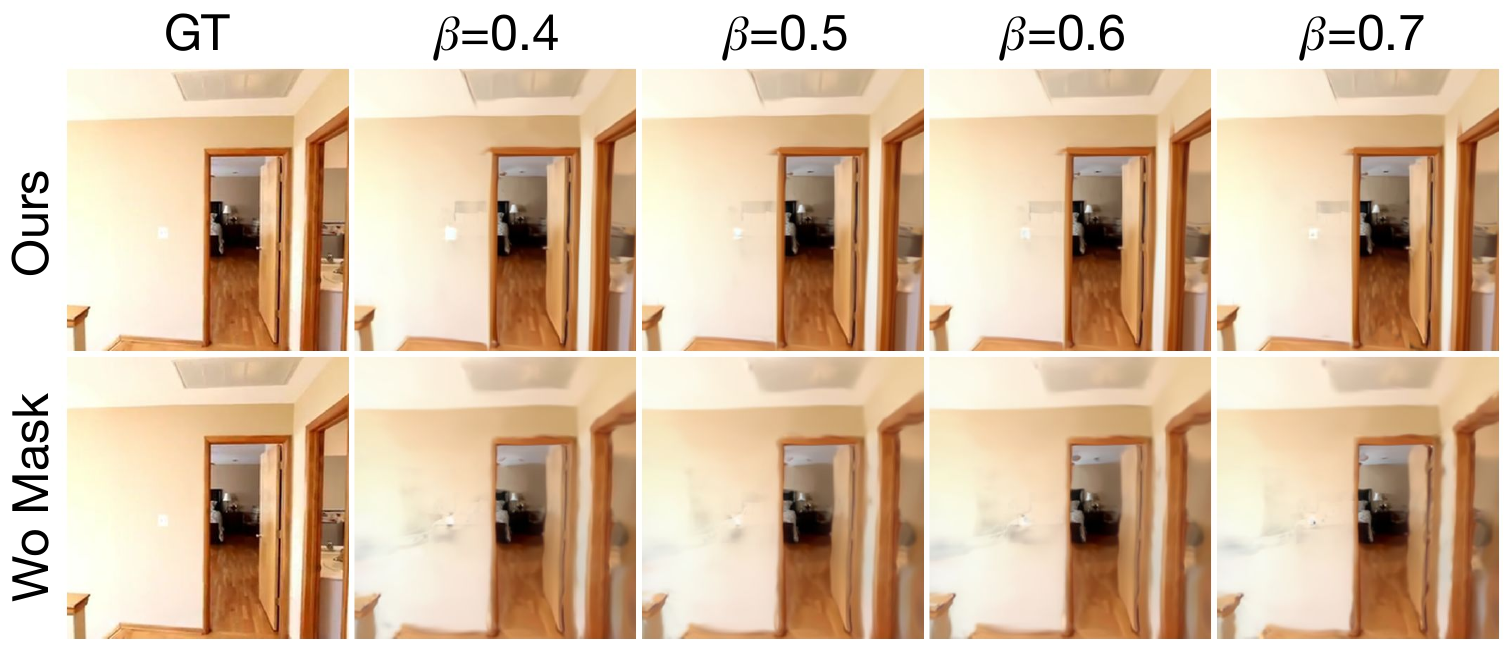}
    \caption{Ablation on RE10K when the binary mask is not provided.}
    \label{fig:ablation_re10k_withoutmask}
\end{figure}

\vspace{0.5em}
We include extended ablations on RE10K across both moderate and aggressive pruning levels in \cref{tab:additional-ablation-re10k} and \cref{fig:ablation_re10k_withoutmask}. These results confirm that each module - Gaussian Adaptation, Masking, and Structured Pruning plays a distinct and essential role.

\begin{table}[!tb]
\centering
\setlength{\tabcolsep}{8pt}
\caption{Ablation study on the RE10K dataset across different pruning regimes. The results highlight the contribution of each design component.}
\resizebox{\linewidth}{!}{
\begin{tabular}{@{}l|lll|lll@{}}
\toprule
                            & \multicolumn{3}{c|}{$\beta=0.4$}                & \multicolumn{3}{c}{$\beta=0.9$}                 \\ \midrule
\textbf{Ablation}            & \textbf{PSNR$\uparrow$}            & \textbf{LPIPS$\downarrow$}          & \textbf{SSIM$\uparrow$}           & \textbf{PSNR$\uparrow$}            & \textbf{LPIPS$\downarrow$}          & \textbf{SSIM$\uparrow$}           \\ \midrule
Without Gaussian Adaptation & 19.809          & 0.399          & 0.617          & 8.066           & 0.680          & 0.131          \\
Without Mask                & 20.055          & 0.339          & 0.656          & 18.090          & 0.389          & 0.625          \\
Random Pruning Baseline     & \textbf{22.452} & 0.239          & \textbf{0.744} & 16.580          & 0.458          & 0.486          \\
Ours                        & 22.294          & \textbf{0.235} & 0.735          & \textbf{17.848} & \textbf{0.314} & \textbf{0.645} \\ \bottomrule
\end{tabular}
}

\label{tab:additional-ablation-re10k}
\end{table}


Key insights from RE10K ablations:
\begin{itemize}
    \item \textbf{Without Gaussian Adaptation}: feed forward representation collapses under strong pruning ($13K$), showing that adapting the remaining Gaussians is critical for maintaining scene coverage.
    \item \textbf{Without Mask}: performs notably worse than the full model, demonstrating the importance of contextual, texture-aware selection rather than uniform or naïve pruning.
    \item \PaperTitle{}: consistently outperforms all ablations at high compression, validating the necessity of jointly learning masks and adaptive Gaussian updates.
\end{itemize}

\end{document}